%% file: Main.tex
\pdfoutput=1

\documentclass[10pt,twocolumn,letterpaper]{article}

\usepackage[pagenumbers]{cvpr} 

\usepackage{graphicx}
\usepackage{amsmath}
\usepackage{amssymb}
\usepackage{booktabs}
\usepackage{xcolor}
\makeatletter
\@namedef{ver@everyshi.sty}{}
\makeatother
\usepackage{tikz}
\usepackage{comment}
\usepackage{amsmath,amssymb} 
\usepackage{xspace}
\definecolor{darkyellow}{RGB}{246,190,0}

\definecolor{citecolor}{RGB}{34,139,34}
\definecolor{mygreen}{RGB}{40, 150, 60}

\usepackage{soul}
\usepackage{wrapfig}
\usepackage{subcaption}
\usepackage{algorithm2e}

\usepackage[pagebackref,breaklinks,colorlinks,citecolor=citecolor]{hyperref}

\newcommand{\ili}[1]{{\leavevmode\color{red}[ili: #1]}}

\usepackage[T1]{fontenc}
\usepackage[utf8]{inputenc}
\usepackage{babel}
\usepackage[font=small,labelfont=bf]{caption}

\usepackage[capitalize]{cleveref}
\crefname{section}{Sec.}{Secs.}
\Crefname{section}{Section}{Sections}
\Crefname{table}{Table}{Tables}
\crefname{table}{Tab.}{Tabs.}

\def\cvprPaperID{2478} 
\def\confName{CVPR}
\def\confYear{2023}

\begin{document}

\title{Simple Cues Lead to a Strong Multi-Object Tracker}

\author{Jenny Seidenschwarz$^1$\thanks{Correspondance to \tt\small j.seidenschwarz@tum.de.}
\quad
Guillem Bras\'{o} $^1$$^2$
\quad
Victor Castro Serrano$^1$
\quad
Ismail Elezi$^1$
\quad
Laura Leal-Taix\'{e}$^1$\thanks{Currently at NVIDIA.} \\
\\
$^1$Technical University of Munich \hspace{1cm} $^2$Munich Center for Machine Learning\\
}
\maketitle

\begin{abstract}
For a long time, the most common paradigm in Multi-Object Tracking was  tracking-by-detection (TbD), where objects are first detected and then associated over video frames. 
For association, most models resourced to motion and appearance cues, e.g., re-identification networks. 
Recent approaches based on attention propose to learn the cues in a data-driven manner, showing impressive results.  
%
In this paper, we ask ourselves whether simple good old TbD methods are also capable of achieving the performance of end-to-end models.
%
%
%
To this end, we propose two key ingredients that allow a standard re-identification network to excel at appearance-based tracking. 
We extensively analyse its failure cases, and show that a combination of our appearance features with a simple motion model leads to strong tracking results.  
Our tracker generalizes to four public datasets, namely MOT17, MOT20, BDD100k, and DanceTrack, achieving state-of-the-art performance. 
\hyperlink{https://github.com/dvl-tum/GHOST}{https://github.com/dvl-tum/GHOST}. 

\end{abstract}
\vspace{-0.5cm}

\input{chapters/intro_alt}
\input{chapters/Realted_Work}

\input{chapters/Methodology}
\input{chapters/Experiments}
\input{chapters/Conclusion}

\input{chapters/acknowledgement}

{\small
\bibliographystyle{ieee_fullname}
\bibliography{refs}
}
\clearpage
\input{chapters/supp_extra}
\clearpage

\end{document}

%% file: chapters/intro_alt.tex
\section{Introduction}

Multi-Object Tracking (MOT) aims at finding the trajectories of all moving objects in a video. The dominant paradigm in the field has long been tracking-by-detection, which divides tracking into two steps: (i) frame-wise object detection, (ii) data association to link the detections and form trajectories. 
One of the simplest forms of data association for online trackers is frame-by-frame matching using the Hungarian algorithm~\cite{hungarian}. 
Matching is often driven by cues such as appearance, \eg, re-identification (reID) features \cite{DBLP:conf/cvpr/0004GLL019,DBLP:journals/corr/HermansBL17,DBLP:conf/iccv/QuanDWZY19,DBLP:conf/cvpr/ChengGZWZ16,DBLP:conf/eccv/SunZYTW18,DBLP:conf/eccv/VariorHW16,DBLP:conf/cvpr/ZhangLZ019}, or motion cues \cite{DBLP:journals/ijcv/ZhangWWZL21,DBLP:conf/nips/RenHGS15,DBLP:conf/iccv/BergmannML19,DBLP:conf/cvpr/XuOBHLA20,DBLP:conf/eccv/PengWWWWTWLHF20}.
%
%
%
Even recent trackers propose data-driven motion priors~\cite{DBLP:conf/iccv/BergmannML19,DBLP:conf/cvpr/StadlerB21,DBLP:conf/cvpr/WuCS00Y21,DBLP:conf/eccv/ZhouKK20} or appearance cues, which may include external reID models~\cite{DBLP:conf/iccv/BergmannML19,DBLP:conf/cvpr/WuCS00Y21}. 

Most recent trackers based on Transformers \cite{DBLP:journals/corr/abs-2012-15460,DBLP:journals/corr/abs-2101-02702,DBLP:journals/corr/abs-2105-03247}  learn all necessary cues from data through self- and cross-attention between frames and tracked objects.
While this implicitly gets rid of any heuristic typically embedded in the handcrafted appearance and motion cues, and could be the path to more general trackers, the training strategies are highly complex and the amount of data needed to train such models is very large, to the point where MOT datasets \cite{DBLP:journals/ijcv/DendorferOMSCRR21} are not enough and methods rely on pre-training on detection datasets such as CrowdHuman~\cite{crowdhuman}.

\begin{figure}
    \centering
    \includegraphics[width=0.4\textwidth]{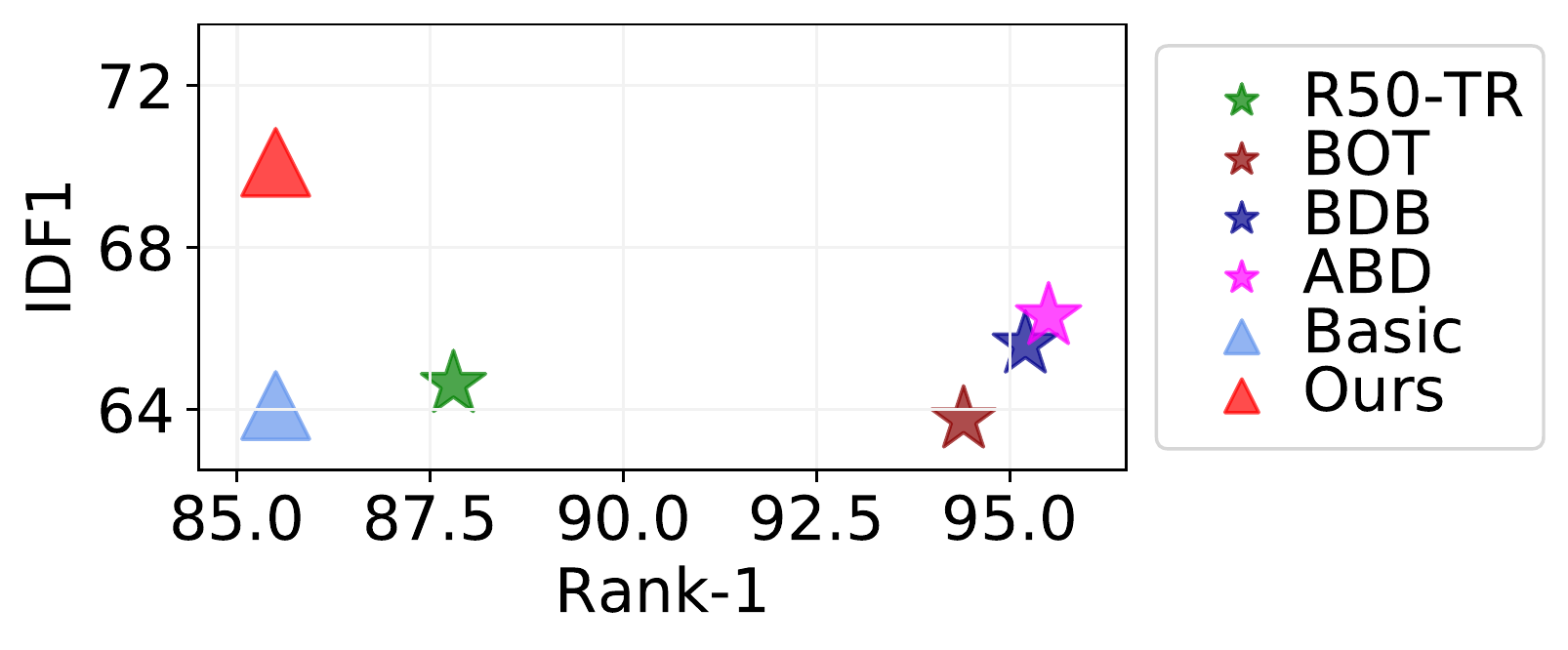}
    \vspace{-0.3cm}
    \caption{IDF1/Rank-1 of different state-of-the-art re-ID approaches. R50-TR \cite{torchreid}, BOT \cite{DBLP:conf/cvpr/0004GLL019}, BDB \cite{DBLP:conf/iccv/DaiCGZT19}, ABD \cite{DBLP:conf/iccv/ChenDXYCYRW19}. \textit{Basic} is our baseline, \textit{Ours} is our appearance model.}
    \label{fig:teaser}
    \vspace{-0.6cm}
\end{figure}

%

While interesting and challenging from a research point of view, it is questionable whether we should also follow the path of \textit{learning everything} in multi-object tracking, when there are strong priors that we know how to define and leverage, such as the good old appearance and motion cues. 
As we show in this paper, there are key observations that need to be made in order to properly leverage such cues. These observations might seem simple and obvious but have been largely overlooked by the community. If we spend as much time in properly understanding and implementing such cues as we do in training Transformers, we will be rewarded with a simple Hungarian tracker with appearance and motion cues that still dominates state-of-the-art on multiple benchmarks, and does not even need to be trained on any tracking data.


%
Our first observation is that simply using state-of-the-art re-identification (reID) networks for appearance matching is not enough for the real scenarios of MOT. In Figure~\ref{fig:teaser}, we visualize the performance of several state-of-the-art reID approaches on Market-1501 dataset \cite{DBLP:conf/iccv/ZhengSTWWT15} (x-axis), as well as the model's performance when used in a simple matching-based tracker (y-axis). It shows that the reID performance does not necessarily translate to MOT performance. 
%
%
%
We identify two problems causing the weak performance of reID models on MOT: (i) reID models need to account for the different challenges expected at different time horizons, \ie, while in nearby frames appearance of objects will vary minimally, in longer time gaps more severe changes are expected, \textit{e.g.}, caused by (partial-) occlusions and
(ii) reID performance tends to be inconsistent across MOT sequences because of their varying image statistics, which in turn differs from the relatively stable conditions of the corresponding reID training dataset.
%
%
We propose two simple but key design choices to overcome the aforementioned problems, \ie, on-the-fly domain adaptation, and different policies for \textcolor{mygreen}{active} and \textcolor{blue}{inactive} tracks.
%
%
Moreover, we conduct an extensive analysis under different conditions of visibility, occlusion time, and camera movement, to determine in which situations reID is not enough and we are in need of a motion model.
%
We combine our reID with a simple linear motion model using a weighted sum so that each cue can be given more weight when needed for different datasets. 
%

Our findings culminate in our proposed \textbf{G}ood \textbf{O}ld \textbf{H}ungarian \textbf{S}imple \textbf{T}racker 
 or \textbf{GHOST} (the order of the letters of the acronym does not change the product) that generalizes to four different datasets, remarkably outperforming the state-of-the-art while, most notably, \textit{never being trained on any tracking dataset}. 

\vspace{0.3cm}

\noindent In summary, we make the following contributions:
\begin{itemize}
    \item We provide key design choices that significantly boost the performance of reID models for the MOT task.
    \item We extensively analyze in which underlying situations appearance is not sufficient and when it can be backed up by motion.
    \item We generalize to four datasets achieving state-of-the-art performance by combining appearance and motion in our simple and general TbD online tracker GHOST.
\end{itemize}
With this paper, we hope to show the importance of domain-specific knowledge and the impact it can have, even on the simple good old models.
%
%
Our observations, \textit{i.e.}, the importance of domain adaptation, the different handling of short- and long-term associations as well as the interplay between motion and appearance are straightforward, almost embedded into the subconscious of the tracking community, and yet they have largely been overlooked by recent methods. 
Introducing our simple but strong tracker, we hope our observations will inspire future work to integrate such observations into sophisticated models further improving the state of the art, in a solution where data and priors will gladly meet.

%% file: chapters/Realted_Work.tex
\section{Related Work}
\label{sec:related_work}

\begin{figure*}[t!]
    \centering
    \begin{subfigure}[b]{0.27\textwidth}
        \centering
        \vspace{-0.8cm}
        \includegraphics[width=\textwidth]{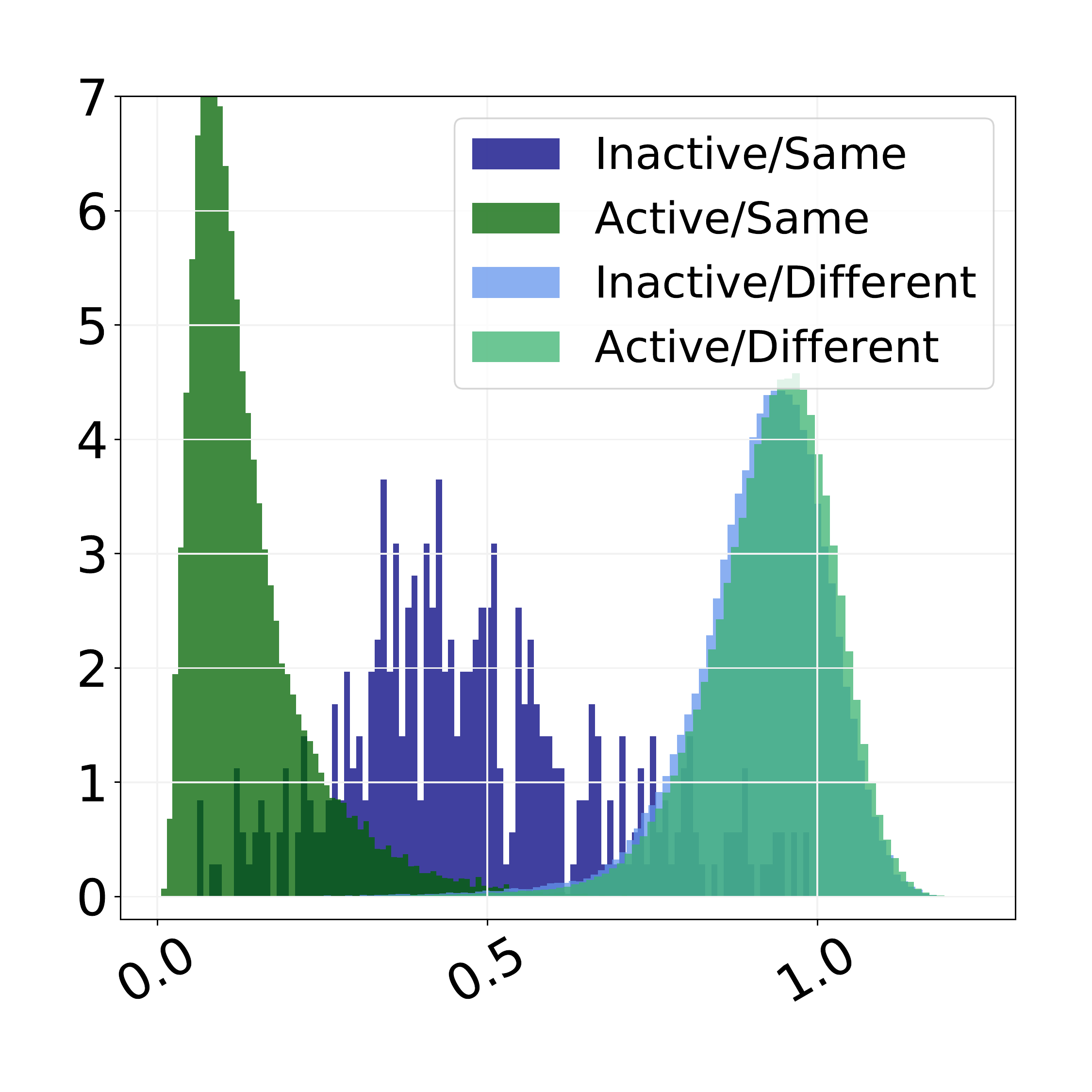}
        \vspace{-0.8cm}
        \caption{Appearance}
    \end{subfigure}
    \quad
    \begin{subfigure}[b]{0.27\textwidth}   
        \centering 
        \vspace{-0.8cm}
        \includegraphics[width=\textwidth]{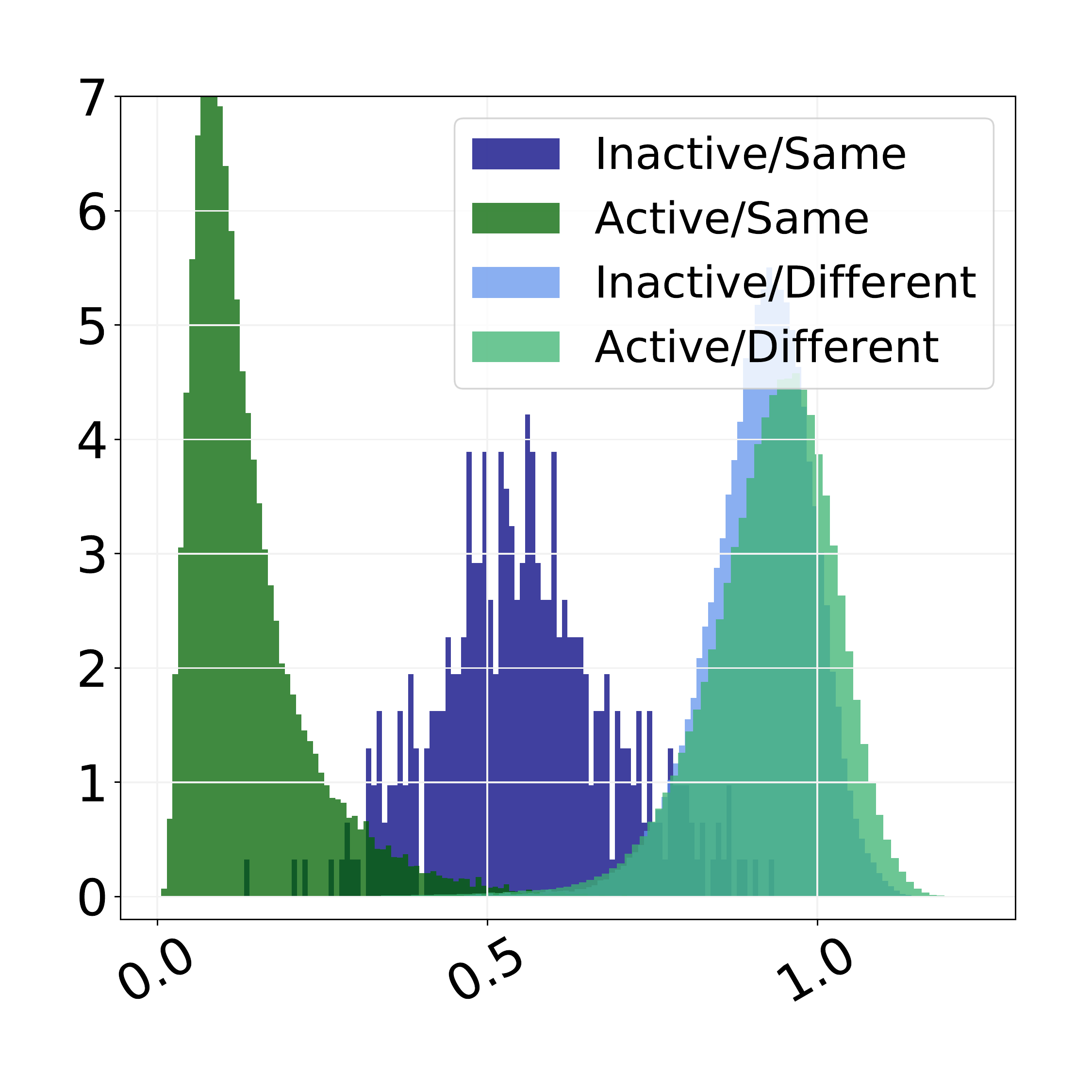}
        \vspace{-0.8cm}
        \caption{Proxy Appearance}
    \end{subfigure}
    \quad
    \begin{subfigure}[b]{0.27\textwidth}   
        \centering 
        \vspace{-0.8cm}
        \includegraphics[width=\textwidth]{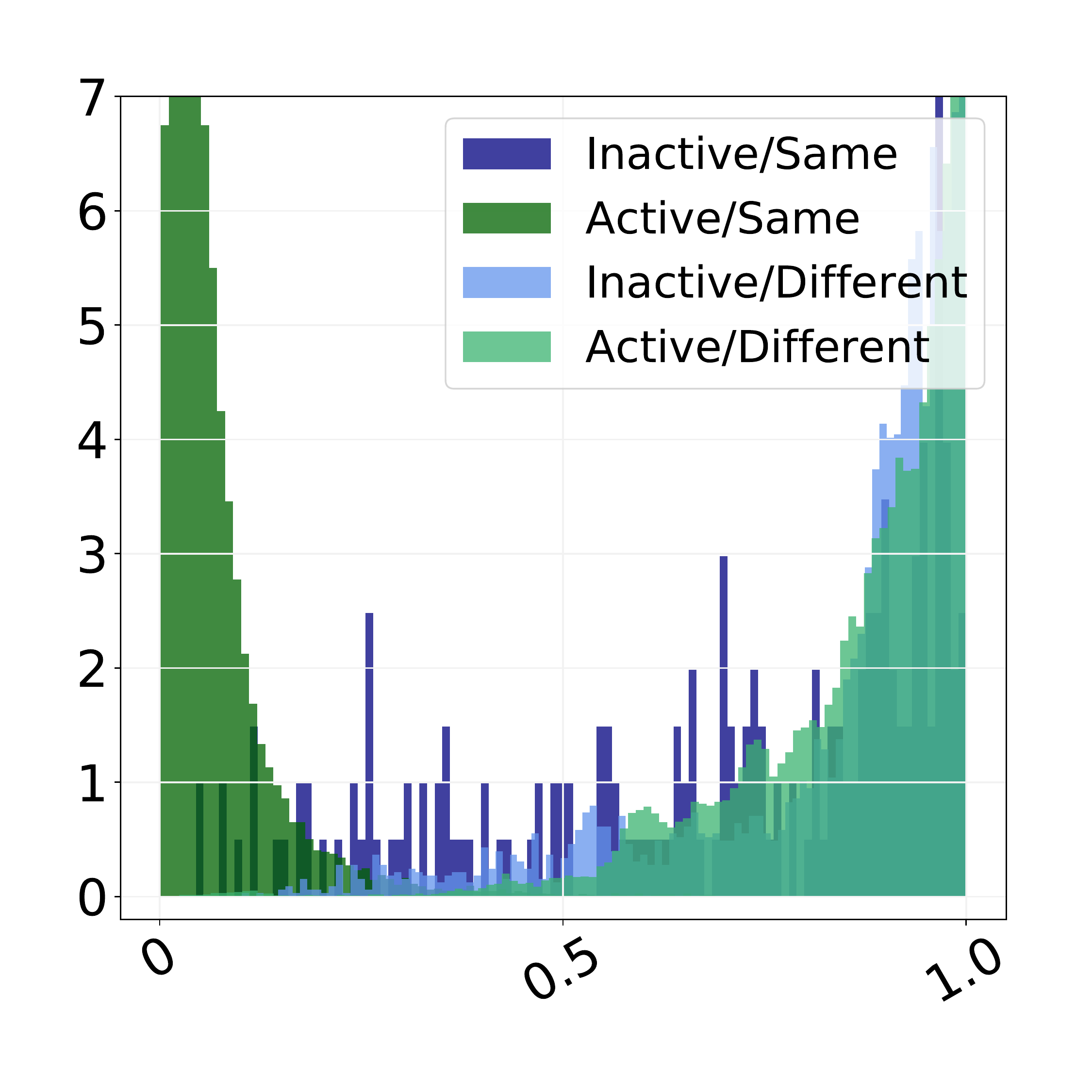}
        \vspace{-0.8cm}
        \caption{Motion}
    \end{subfigure}
    \vspace{-0.2cm}
    \caption{Distance histograms when utilizing (a) the appearance features of the last detection for active and inactive trajectories, (b) the appearance features of the last detection for active and the proxy distance for appearance features of inactive trajectories, (d) the motion distance using IoU measure for active and inactive trajectories.}
    \label{fig:distributions}
    \vspace{-0.5cm}
\end{figure*}

In the last years, TbD was the most common paradigm used in MOT 
\cite{DBLP:conf/iccv/BergmannML19,DBLP:conf/cvpr/XuOBHLA20,DBLP:conf/eccv/WangZLLW20,DBLP:journals/ijcv/ZhangWWZL21,DBLP:journals/corr/abs-2103-14258,DBLP:conf/cvpr/PangLZLL20,DBLP:conf/cvpr/PangQLCLDY21,DBLP:conf/cvpr/BrasoL20,DBLP:conf/icml/HornakovaHRS20}. 
Pedestrians are first detected using object detectors \cite{DBLP:conf/nips/RenHGS15,DBLP:conf/cvpr/RedmonF17,DBLP:conf/cvpr/YangCL16}. 
Then, detections are associated across frames to form trajectories corresponding to a certain identity utilizing motion, location, appearance cues, or a combination of them. The association can either be solved frame-by-frame for online applications or offline in a track-wise manner over the sequence.

\noindent\textbf{Graph-Based Approaches.}
One common formalism to perform data association following the TbD paradigm is viewing each detection as a node in a graph with edges linking several nodes over the temporal domain to form trajectories. Determining which nodes are connected can then be solved using maximum flow \cite{DBLP:journals/pami/BerclazFTF11} or minimum cost approaches \cite{DBLP:conf/cvpr/JiangFL07,DBLP:conf/cvpr/PirsiavashRF11,DBLP:conf/cvpr/ZhangLN08} by, \textit{e.g}, taking motion models into account \cite{DBLP:conf/iccvw/Leal-TaixePR11}. Recent advances combined track-wise graph-based models with neural networks \cite{DBLP:conf/cvpr/BrasoL20}.
%
%
We challenge those recent advances by showing that we can obtain strong TbD trackers without using a complex graph model combining our strong but simple cues. 



\noindent\textbf{Motion-Based Association.}
Different from the graph-based approaches, many TbD approaches perform frame-by-frame association directly using motion and location cues from detections and existing trajectories 
\cite{DBLP:conf/icip/BewleyGORU16,DBLP:conf/avss/BochinskiES17,DBLP:conf/eccv/ZhouKK20,DBLP:conf/eccv/PengWWWWTWLHF20,DBLP:conf/cvpr/PangLZLL20}. For short term preservation, those trackers exploit that given two nearby frames object displacements tend to be small. This allows them to utilize spatial proximity for matching by exploiting, \eg, Kalman filters \cite{DBLP:conf/icip/BewleyGORU16}. Taking this idea further, approaches following the tracking-by-regression paradigm utilize object detectors \cite{DBLP:conf/iccv/BergmannML19,DBLP:conf/eccv/ZhouKK20} to regress bounding box positions. Recent advances introduced Transformer-based approaches \cite{DBLP:journals/corr/abs-2101-02702,DBLP:journals/corr/abs-2012-15460,DBLP:journals/corr/abs-2105-03247} that perform tracking following the tracking-by-attention paradigm. 
%
%
%
%
Using sophisticated motion models, those approaches reach outstanding performance on several datasets, especially in short-term associations.  However, especially the Transformer-based approaches require sophisticated training strategies. 
Contrary to all those approaches, we show that a simple linear motion model suffices to model short-term associations in most scenarios. 
In scenes with moving cameras or scenarios requiring non-trivial long-term associations, \textit{e.g.}, scenarios with many occlusions, purely motion-based trackers struggle which calls for a combination with appearance-based cues.

\noindent\textbf{Appearance-Based Association.}
To achieve better performance in long-term association scenarios, numerous approaches use additional appearance-based re-identification networks that encode appearance cues to re-identify persons after occlusions \cite{DBLP:conf/eccv/YuLLLSY16,DBLP:conf/iccv/BergmannML19,DBLP:conf/cvpr/BrasoL20,DBLP:conf/cvpr/KimLAR21,DBLP:conf/icip/WojkeBP17,DBLP:conf/cvpr/Leal-TaixeCS16,DBLP:conf/iccv/SadeghianAS17,DBLP:conf/cvpr/SonBCH17}.
%
Further exploring this direction, a recent work \cite{DBLP:conf/cvpr/PangQLCLDY21} proposed to train a detection network solely utilizing embedding information during the training process. 
Enhancing MOT towards real-time, several works proposed to jointly compute detections and embeddings in a multi-task setting \cite{DBLP:conf/cvpr/XuOBHLA20,DBLP:conf/cvpr/GuoWWT21,DBLP:conf/eccv/WangZLLW20,DBLP:conf/cvpr/LuRVH20,DBLP:journals/ijcv/ZhangWWZL21}. Some of them introduce more balanced training schemes \cite{DBLP:conf/cvpr/GuoWWT21,DBLP:conf/cvpr/XuOBHLA20} to better leverage the synergies between both cues.
%
%
While promising, approaches using appearance additionally to motion cues require rather complex association schemes with several steps \cite{DBLP:journals/ijcv/ZhangWWZL21,DBLP:conf/icip/BewleyGORU16}.
Also, complex and highly differing training schedules or differing inference strategies make it hard to draw conclusions about what really is driving the progress in the field. 
In contrast, GHOST does not rely on complex procedures but combines lightweight motion and spiced-up appearance cues in a simple yet strong TbD tracker that only requires little training data. 

\noindent\textbf{Person Re-Identification and Domain Adaptation.}
In contrast to the tracking domain, the goal of person reID is to retrieve person bounding boxes from a large gallery set that show the same person as a given query image based on appearance cues. 
However, state-of-the-art reID models tend to significantly drop in performance when evaluated on out-of-domain samples, \textit{i.e.}, samples coming from other datasets \cite{DBLP:conf/cvpr/ChoiKJPK21,DBLP:conf/cvpr/ZhaoZYLLLS21,DBLP:conf/eccv/ZhuangWXZZWAT20}. As during application, person reID models are applied to different cameras, several approaches on cross-dataset evaluation emerged that transfer the knowledge from a given source, \textit{i.e.}, training to a given target, \textit{i.e.}, test domain utilizing domain adaptation (DA) \cite{DBLP:conf/cvpr/ChoiKJPK21,DBLP:conf/cvpr/ZhaoZYLLLS21,DBLP:conf/eccv/ZhuangWXZZWAT20}. 
%
DA often relies on adapting Batch Normalization (BN) statistics to account for distribution shifts between different domains.
The weight matrix and the BN statistics store label and domain-related knowledge, respectively \cite{DBLP:conf/iclr/LiWS0H17}. 
To update the latter, the statistics of BN layers can be updated, \textit{e.g.}, by taking the mean and variance of all target domain images \cite{DBLP:conf/iclr/LiWS0H17}, by re-training using pseudo labels \cite{DBLP:conf/cvpr/ChangYSKH19} or combining train and test datatset statistics \cite{DBLP:conf/nips/SchneiderRE0BB20}.
Apart from the statistics, the learned parameters $\beta$ and $\gamma$ can also be updated \cite{DBLP:conf/iclr/WangSLOD21}. 
An approach similar to ours but for classification \cite{DBLP:journals/corr/abs-2006-10963} updates BN statistics during test time in a batch-wise manner.
Inspired by those recent advances, we enhance our appearance model to be better suited for MOT using a simple on-the-fly domain adaptation approach. This directly adapts the model's learned training dataset statistics (source) to the sequences (targets).


%% file: chapters/Methodology.tex
\section{Methodology}
\label{sec:methodology}

Based on the good old Hungarian TbD paradigm, GHOST combines design choices that have been overlooked by the community until now. 
We start by giving the general pipeline of GHOST (see Sec. \ref{subsec:pipeline}) and then build our strong appearance model (see Sec. \ref{subsec:ReID}).
%

%

\subsection{A simple tracking-by-detection tracker}
\label{subsec:pipeline}

Our tracker takes as input a set of detections $\mathcal{O} = \{o_1, ..., o_M\}$, each represented by $o_i = (f_i, p_i)$. $f_i$ are appearance feature vectors generated from the raw detection pixels using a Convolutional Neural Network (CNN) and $p_i$ is the bounding box position in image coordinates.
A trajectory or track is defined as a set of time-ordered detections ${T}_j = \{o_{j_1}, ..., o_{j_{N_j}}\}$ where $N_j$ is the number of detections in trajectory $j$. Moreover, each trajectory has a corresponding predicted position $\hat{p}_j^t$ at time step $t$, produced by our linear motion model. During the tracking process, detections are assigned to trajectories. 
If no new detection is added to a trajectory at a given frame, we set its status to \textcolor{blue}{inactive} whereas it remains \textcolor{mygreen}{active} otherwise. We use a memory bank to keep inactive trajectories of up to 50 frames. 
The goal is to find the trajectories $\mathcal{T} = \{{T}_1, ..., {T}_M\}$ that best match the detections to the underlying ground truth trajectories.

Towards that end, we associate existing detections over consecutive frames utilizing bipartite matching via the Hungarian algorithm
as commonly done \cite{DBLP:conf/icip/WojkeBP17,DBLP:journals/ijcv/ZhangWWZL21,DBLP:conf/eccv/WangZLLW20,DBLP:conf/ijcai/LiuC0Y20,DBLP:conf/icip/BewleyGORU16}. 
The assignment is driven by a cost matrix that compares new detections with the tracks already obtained in previous frames. To populate the cost matrix, we use appearance features, motion cues, or both. Our final tracker utilizes a simple weighted sum of both. We filter detection-trajectory pairs $(i, j)$ \textit{after} the matching using matching thresholds $\tau_i$.

\subsection{Strong appearance model for MOT}
\label{subsec:ReID}

Our appearance model is based on ResNet50 \cite{DBLP:conf/cvpr/HeZRS16} with one additional fully-connected layer at the end for downsampling, and trained on a common person reID dataset \cite{DBLP:conf/iccv/ZhengSTWWT15}.
It is important to note that we \textit{do not train any part of our reID model on any MOT dataset}.  
As we will show in experiments, this basic reID model does not perform well on the MOT task. We, therefore, propose two design choices to make our appearance model stronger: (i) we handle active and inactive tracks differently; (ii) we add on-the-fly domain adaption. For this, we analyze distances between detections and tracks in given MOT sequences.

\noindent\textbf{Appearance distance histograms.} In Fig~\ref{fig:distributions} we analyze the histograms of distances between new detections and \textcolor{mygreen}{active} or \textcolor{blue}{inactive} tracks on MOT17 validation set (please refer to Sec. 8 supplementary for details) utilizing different distance measures.
In dark and light colors we show the distance of a track to a new detection of the same (positive match) and different (negative match) identity, respectively.

\noindent\textbf{Different handling of active and inactive tracks.} 
%
%
%
While the appearance embeddings of one identity barely vary between two consecutive frames, the embeddings of the same identity before and after occlusion can show larger distances due to, \eg, partial occlusion or varying poses.
This pattern can be observed in Fig~\ref{fig:distributions}(a) where we visualize the distance between new detections and the last detection of \textcolor{mygreen}{active} or \textcolor{blue}{inactive} tracks. The two dark-coloured histograms vary significantly, which suggestis that a different treatment of \textcolor{mygreen}{active} or \textcolor{blue}{inactive} tracks is necessary. Furthermore, we can see the overlap between both negative and positive matches for \textcolor{blue}{inactive}  tracks showing the inherent difficulty of matching after occlusion.

Hence, for \textcolor{mygreen}{active} tracks, we leverage the appearance features of the detection assigned to track $j$ at frame $t-1$ $f_j^{t-1}$ for the distance computation to detection $i$ at frame $t$ $d_{i,j} = d(f_i,f_j^{t-1})$.
For the \textcolor{blue}{inactive} tracks we compute the distance between the appearance feature vectors of all $N_k$ detections in the \textcolor{blue}{inactive} track $k$ and the new detection $i$ and utilize the mean of those distances as a \textit{proxy distance}:
\begin{equation}
    d_{i, k} = \frac{1}{N_k} \sum_{n=1}^{N_k} d(f_i,f_k^n)
\end{equation}
The resulting distance histograms are as visualized in Fig~\ref{fig:distributions}(b). This proxy distance leads to a more robust estimate of the true underlying distance between a detection and an \textcolor{blue}{inactive} track. Hence, in contrast to when using a single feature vector of the \textcolor{blue}{inactive} track (see  Fig~\ref{fig:distributions}(a)) utilizing the proxy distance leads to better-separated histograms (see  Fig~\ref{fig:distributions}(b)).


Moreover, the different histograms of \textcolor{mygreen}{active} and \textcolor{blue}{inactive} trajectories call for different handling during the bipartite matching. To be specific, thresholds typically determine up to which cost a matching should be allowed.
Looking at Fig~\ref{fig:distributions}(b), different thresholds $\tau_i$ divide the histograms of distances from \textcolor{mygreen}{active} and \textcolor{blue}{inactive} trajectories to detections of the same (dark colors) and different (pale colors) identities. Utilizing \textit{different matching thresholds} $\tau_{act}$ and $\tau_{inact}$ for active and inactive trajectories allows us to keep one single matching. In contrast to cascaded matching \cite{DBLP:conf/icip/BewleyGORU16}, our assignment is simpler and avoids applying bipartite matching several times at each frame.

\noindent\textbf{On-the-fly Domain Adaptation.} As introduced in Section~\ref{sec:related_work}, recent developments in the field of person reID propose to apply domain adaptation (DA) techniques as the source dataset statistics may not match the target ones \cite{DBLP:conf/cvpr/ChoiKJPK21,DBLP:conf/cvpr/ZhaoZYLLLS21,DBLP:conf/eccv/ZhuangWXZZWAT20}. 
For MOT this is even more severe since each sequence follows different statistics and represents a new target domain. 
We, therefore, propose to apply an on-the-fly DA in order to prevent performance degradation of reID models when applied to varied MOT sequences. This allows us to capitalize on a strong reID over all sequences.

Recently, several works on person reID introduced approaches utilizing ideas from DA to achieve cross-dataset generalization by adapting normalization layers to Instance-Batch, Meta Batch, or Camera-Batch Normalization layers \cite{DBLP:conf/cvpr/ChoiKJPK21,DBLP:conf/cvpr/ZhaoZYLLLS21,DBLP:conf/eccv/ZhuangWXZZWAT20}. 
%
Contrary to the above-mentioned approaches, we utilize the mean and variance of the features of the current batch, which corresponds to the detections in one frame during test time, in the BN layers of our architecture:

\begin{equation}
    \hat{x_i} = \gamma \frac{x_i - \mu_b}{\sqrt{\sigma_b + \epsilon}} + \beta
\end{equation}

where $x_i$ are features of sample $i$, $\mu_b$ and $\sigma_b$ are the mean and variance of the current batch, $\epsilon$ is a small value that ensures numerical stability, and $\gamma$ and $\beta$ are learned during training. While not requiring any sophisticated training procedure or test time adaptions nor several BN layers, this approximates the statistics of the sequences reasonably well as all images of one sequence have highly similar underlying distributions and leads to more similar distance histograms across tracking sequences. This in turn allows us to define matching thresholds $\tau_i$ that are well suited for all sequences, \textit{i.e.}, that separate all histograms well. For a more detailed analysis please refer to the supplementary material.

We empirically show that applying these design choices to our appearance model, makes it more robust towards occlusions and better suited for different sequences.

%% file: chapters/Experiments.tex
\section{Experiments}

\label{sec:experiments}

\subsection{Implementation Details}
\label{subsec:implementation_details}
Our appearance model follows common practice~\cite{DBLP:conf/iccv/BergmannML19,DBLP:conf/cvpr/BrasoL20,DBLP:conf/cvpr/DaiWCZHD21}, with a ResNet50 \cite{DBLP:conf/cvpr/HeZRS16} model 
with one additional fully-connected layer at the end to downsample the feature vectors. We train our model on the Market-1501 dataset~\cite{DBLP:conf/iccv/ZhengSTWWT15} using label-smoothed cross-entropy loss with temperature for $70$ epochs, with an initial learning rate of $0.0001$, and decay the learning rate by $10$ after $30$ and $50$ epochs. For optimization, we utilize the RAdam optimizer~\cite{DBLP:conf/iclr/LiuJHCLG020}. Moreover, we add a BN layer before the final classification layer 
during training and utilize class balanced sampling as in~\cite{DBLP:conf/cvpr/0004GLL019}.
We resize the input images to $384 \times 128$ and apply random cropping as well as horizontal flipping during training \cite{DBLP:conf/cvpr/0004GLL019}.
Evaluated on Market-1501 dataset this model achieves $85.2$ rank-1, which is far below the current state-of-the-art performance~\cite{DBLP:conf/cvpr/0004GLL019,DBLP:journals/corr/HermansBL17,DBLP:conf/iccv/QuanDWZY19,DBLP:conf/cvpr/ChengGZWZ16,DBLP:conf/eccv/SunZYTW18,DBLP:conf/eccv/VariorHW16,DBLP:conf/cvpr/ZhangLZ019}.
For tracking, we define the appearance distance between $i$ and $j$ as the cosine distance between appearance embeddings $d_a(i, j) = 1 - \frac{f_i \cdot f_j}{||f_i||\cdot||f_j||}$. As motion distance we use the intersection over union (IoU) between two bounding boxes $d_m(i, j) = IoU(p_i , p_j) = \frac{|p_i \cap p_j|}{|p_i \cup p_j|}$. 
%
\input{tables/reid_balation}

\subsection{Datasets and Metrics}
\label{subsec:datasets}
In this section we introduce the datasets we evaluate GHOST on. MOT17 and MOT20 can be evaluated in public and private detection setting. For the private detection settings, BDD and DanceTrack we use detections generated by YOLOX-X~\cite{DBLP:journals/corr/abs-2107-08430} following the training procedure of \cite{DBLP:journals/corr/abs-2110-06864}.

\noindent\textbf{MOT17.} The dataset \cite{DBLP:journals/corr/MilanL0RS16} consists of seven train and test sequences of moving and static cameras. As common practice, for public detections we utilize bounding boxes refined by CenterTrack \cite{DBLP:conf/eccv/ZhouKK20,DBLP:conf/cvpr/SalehARSG21,DBLP:journals/corr/abs-2006-02609} as well as Tracktor \cite{DBLP:conf/iccv/BergmannML19,DBLP:conf/cvpr/XuOBHLA20,DBLP:conf/cvpr/YinWMYS20,DBLP:journals/pr/PengWLWSWD20,DBLP:conf/ijcai/LiuC0Y20,DBLP:conf/cvpr/SalehARSG21,DBLP:conf/cvpr/KimLAR21} for MOT17. For our ablation studies, we split MOT17 train sequences along the temporal dimension and use the first half of each sequence as train and the second half as evaluation set ~\cite{DBLP:conf/eccv/ZhouKK20,DBLP:conf/cvpr/WuCS00Y21}.

\noindent\textbf{MOT20.} Different from MOT17, MOT20 \cite{DBLP:journals/corr/abs-2003-09003} consists of four train and test sequences being heavily crowded with over $100$ pedestrians per frame. For the public setting, we utilize bounding boxes pre-processed by Tracktor \cite{DBLP:conf/iccv/BergmannML19,DBLP:conf/cvpr/SalehARSG21,DBLP:conf/cvpr/BrasoL20}.

\noindent\textbf{DanceTrack.}
The dataset \cite{sun2022dance} significantly differs from MOT17 and MOT20 datasets in only containing videos of dancing humans having highly similar appearance, diverse motion, and extreme articulation. It contains 40, 25 and 35 videos for training, validation and testing.

\noindent\textbf{BDD.}
The MOT datasets of BDD 100k \cite{DBLP:conf/cvpr/YuCWXCLMD20} consists of 1400, 200 and 400 train, validation and test sequences with eight different classes with highly differing frequencies of the different classes. Note that our appearance model was \textit{never} trained on classes other than pedestrians.

\noindent\textbf{Metrics.} 
The benchmarks provide several evaluation metrics among which HOTA metric \cite{DBLP:journals/ijcv/LuitenODTGLL21}, IDF1 score \cite{DBLP:conf/eccv/RistaniSZCT16} and MOTA \cite{DBLP:journals/pami/KasturiGSMGBBKZ09} are the most common. While MOTA metric mainly is determined by object coverage and IDF1 mostly focus on identity preservation, HOTA balances both.

\input{tables/da_ablation}

\subsection{Appearance Ablation}
\label{subsec:reid_ablation}
In this section, we investigate the impact of the design choices of our appearance model on tracking performance. To this end, we do not utilize motion. In Table~\ref{tab:reid_blation}, we report our results on public detection bounding boxes of MOT17 as well as on BDD. 
%
%
The first row shows the performance of our basic appearance model. 

\noindent\textbf{Different Handling of Active and Inactive Tracks.} As introduced in subsection~\ref{subsec:ReID}, the distance histograms for active and inactive tracks differ significantly. 
In the second row in Table~\ref{tab:reid_blation}, we show that utilizing different thresholds for active and inactive tracks (diff $\tau$) improves our tracking performance by $0.5$ percentage points (pp) ($0.9pp$) in IDF1 and $0.2pp$ ($0.6pp$) in HOTA on MOT17 (BDD). Moreover, utilizing our proxy distance (IP) computation instead of the last detection for inactive tracks further adds $0.9pp$ ($1.8pp$) in IDF1 and $0.6pp$ ($1.0pp$) in HOTA.

\begin{figure*}
    \centering
    \begin{subfigure}[b]{0.4\textwidth}
        \centering
        \includegraphics[height=4.5cm]{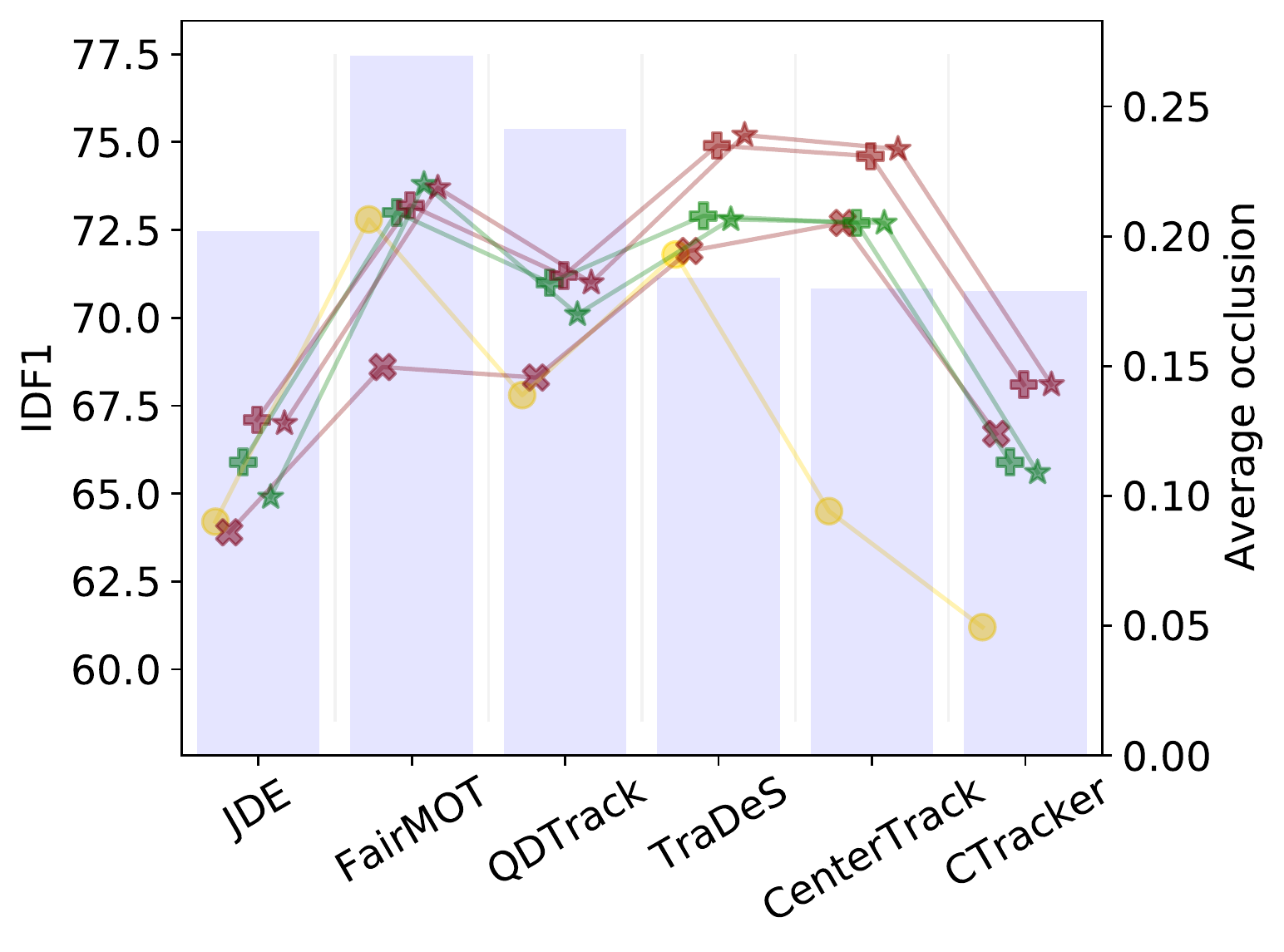}
    \end{subfigure}
    \quad
    \begin{subfigure}[b]{0.55\textwidth}   
        \centering 
        \includegraphics[height=4.5cm]{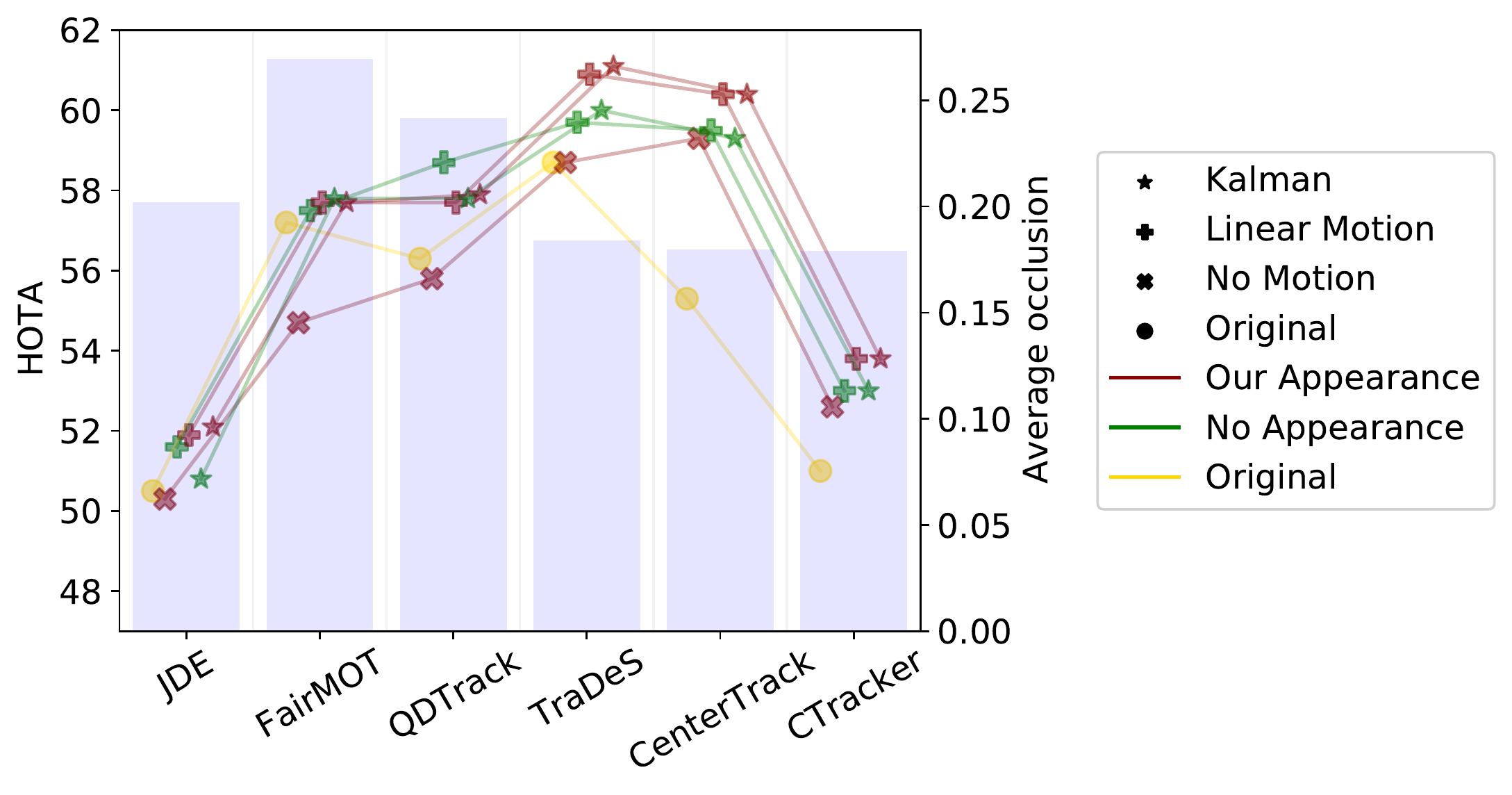}
    \end{subfigure}
    \vspace{-0.35cm}
    \caption{Analysis of contribution of appearance, motion and their combination upon private tracker bounding boxes \cite{DBLP:conf/eccv/WangZLLW20,DBLP:journals/ijcv/ZhangWWZL21,DBLP:conf/cvpr/PangQLCLDY21,DBLP:conf/cvpr/WuCS00Y21,DBLP:conf/eccv/ZhouKK20,DBLP:conf/eccv/PengWWWWTWLHF20}.}
    \label{fig:private}
    \vspace{-0.4cm}
\end{figure*}

\noindent\textbf{On-the-fly domain adaptation.} Additionally, we leverage our on-the-fly domain adaptation (DA) introduced in Subsection~\ref{subsec:ReID} that accounts for differences among the sequences, allowing us to have a well-suited threshold over all sequences. We gain another $1.7pp$ ($1.1pp$) in IDF1 and $0.9pp$ ($0.8pp$) in HOTA metrics. 
We also compare our on-the-fly DA to various different other domain adaptation approaches (see Table~\ref{tab:domain_adaptation_ablation}).
First, we ablate different versions of GHOST, \textit{i.e.}, utilizing random patches of a given frame instead of the pedestrian bounding boxes, utilizing the bounding boxes of the 10 frames before the current frame as well as feeding the whole sequence first to update the parameters. Except for the last, none of them leads to a performance improvement. We argue, that random patches do not represent the statistics of pedestrians well. Also, we fine-tune our reID network on MOT17. For this, we split the train sequences into three cross-validation splits and fine-tune one model for each split. This is necessary as the same identities are given if a sequence is split along the temporal domain. 
While the model is fine-tuned on tracking sequences, the sequences still differ in their distributions among each other. Hence, despite fine-tuning also significantly improving the performance it does not surpass DA. The same holds for using Instance-Batch Norm (IBN) \cite{DBLP:conf/eccv/PanLST18} instead of BN layers, which combine the advantages of Instance and Batch Normalization layers. 


\subsection{Strengths of Motion and Appearance}
\label{subsec:analysis}
In this subsection, we analyze the performance of our appearance cues as introduced in Subsection~\ref{subsec:ReID} to find their strengths with respect to given tracking conditions, namely, visibility level of the detection, occlusion time of the track, and camera movement. We also analyze the complementary performance of our linear motion model that we introduce in the following. Here and in the following Subsection~\ref{subsec:potential}, we apply GHOST on the bounding boxes produced by several private trackers, treating them as raw object detections. We note that this is \textit{not} a state-of-the-art comparison, and emphasize that we solely use those experiments for analysis and to show the potential of our insights. 

\noindent\textbf{Linear Motion Model.} While many works apply more complex, motion models, \eg, Kalman Filters 
\cite{DBLP:journals/ijcv/ZhangWWZL21,DBLP:conf/eccv/WangZLLW20}, social motion models \cite{DBLP:conf/iccvw/Leal-TaixePR11}, or utilize detectors as motion model \cite{DBLP:conf/iccv/BergmannML19,DBLP:conf/eccv/ZhouKK20}, we choose on purpose a simple linear model for our experiments.
Although the world does not move with constant velocity, many short-term movements, as in the case of two consecutive frames, can be approximated with linear models and assuming constant velocity.
Given detections of a track $j$, we compute the mean velocity $v_j$ between the last $k$ consecutive detections and predict the current position of a track by
\begin{equation}
    \hat{p}_j^{t} = p_j^{t-1} + v_j \cdot \Delta t,
\end{equation}

where $\Delta t$ is the time difference from one frame to another and $p_j^{t-1}$ is the position of the previous detection for active and the last predicted position for inactive tracks.

To obtain the motion distance between the new detections and the tracks, we compute the IoU between the position of a new detection $p_i$ and $\hat{p}_j^{t}$. We also visualize the corresponding distance histogram in Fig~\ref{fig:distributions}(c), showing that distance histograms between detections and tracks of the same and different identities are well separated for active tracks. In the following, we underline this observation by showing that this simple motion model is able to solve most situations. We set $k$ to use all previous positions in Subsections~\ref{subsec:analysis} and ~\ref{subsec:potential}. 

\noindent\textbf{Analysis Setup.} For our analysis, we investigate the rate of correct associations ($RCA$) on MOT17 validation set~\cite{DBLP:journals/ijcv/DendorferOMSCRR21}, which we define as:
\begin{equation}
    RCA = \frac{\text{TP-Ass}}{\text{FP-Ass} + \text{TP-Ass}},
\end{equation}
where $\text{TP-Ass}$ and $\text{FP-Ass}$ represents true positive and false positive association, respectively. We average RCA over the sets of pre-processed private detections from several trackers (see Section~\ref{subsec:potential}) to get less noisy statistics.  

%

\begin{figure*}[ht!]
    \centering
    \begin{subfigure}[b]{0.29\textwidth}
        \centering
        \includegraphics[width=0.99\textwidth]{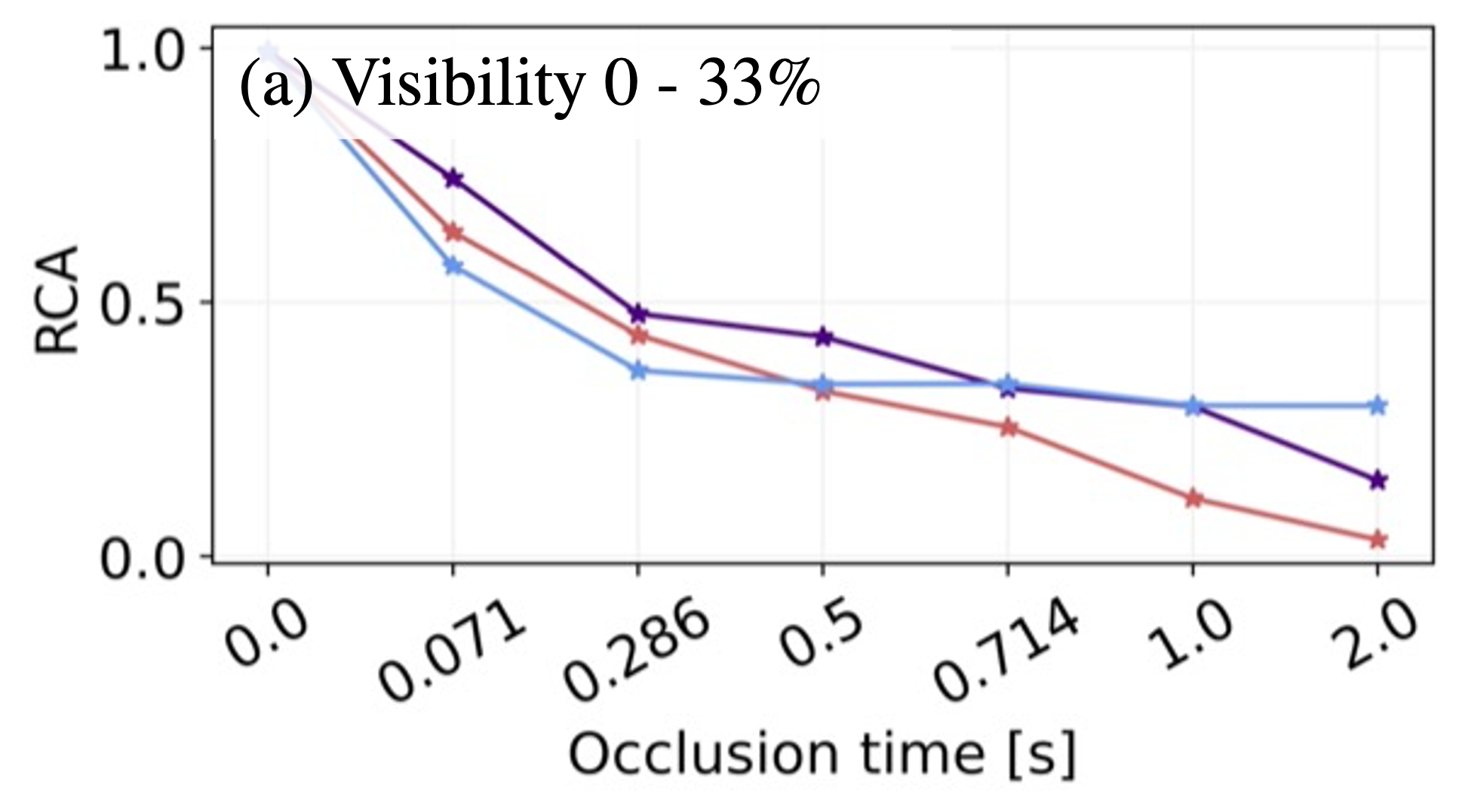}
    \end{subfigure}
    \quad
    \begin{subfigure}[b]{0.29\textwidth}   
        \centering 
        \includegraphics[width=0.99\textwidth]{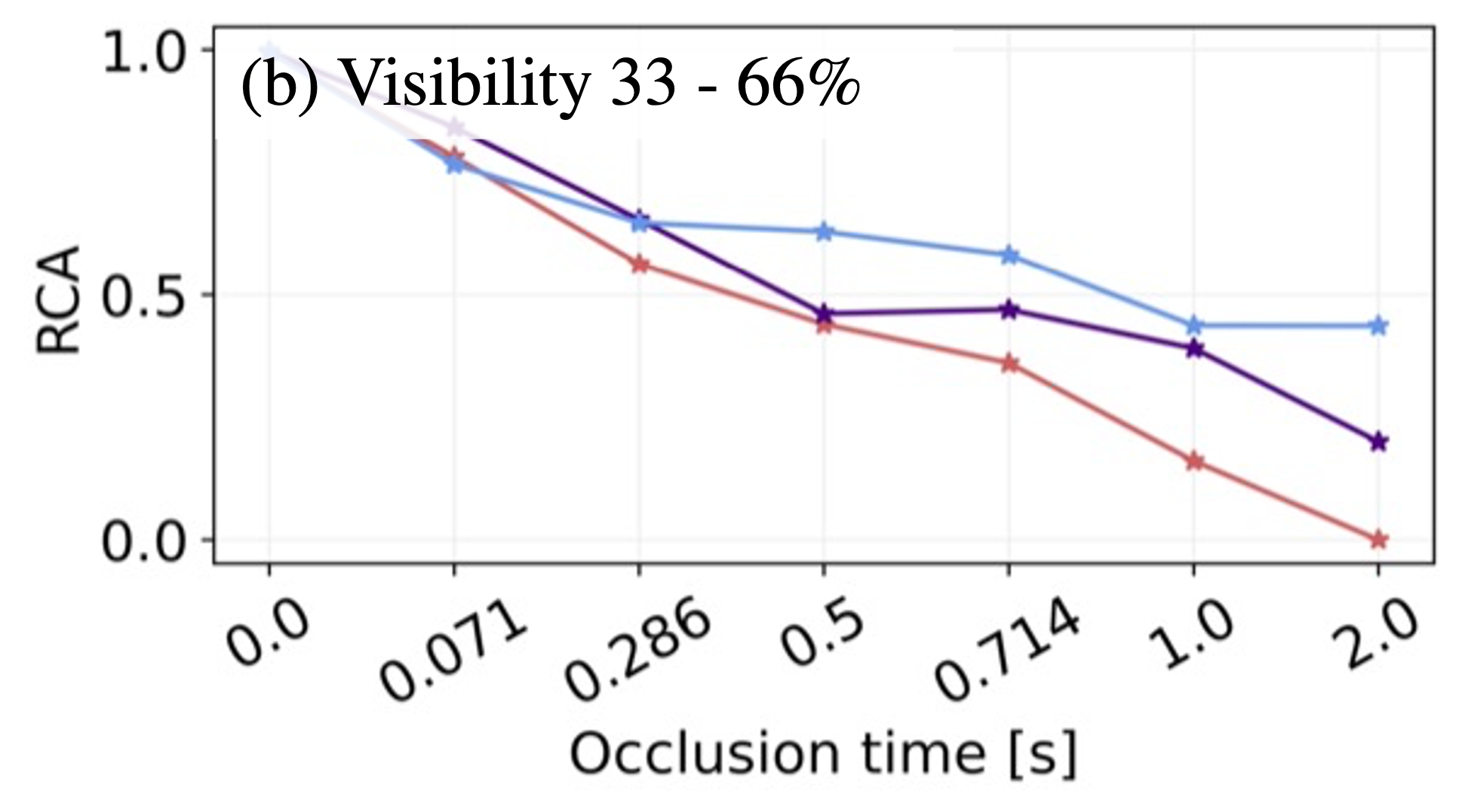}
    \end{subfigure}
    \quad
    \begin{subfigure}[b]{0.29\textwidth}   
        \centering 
        \includegraphics[width=0.99\textwidth]{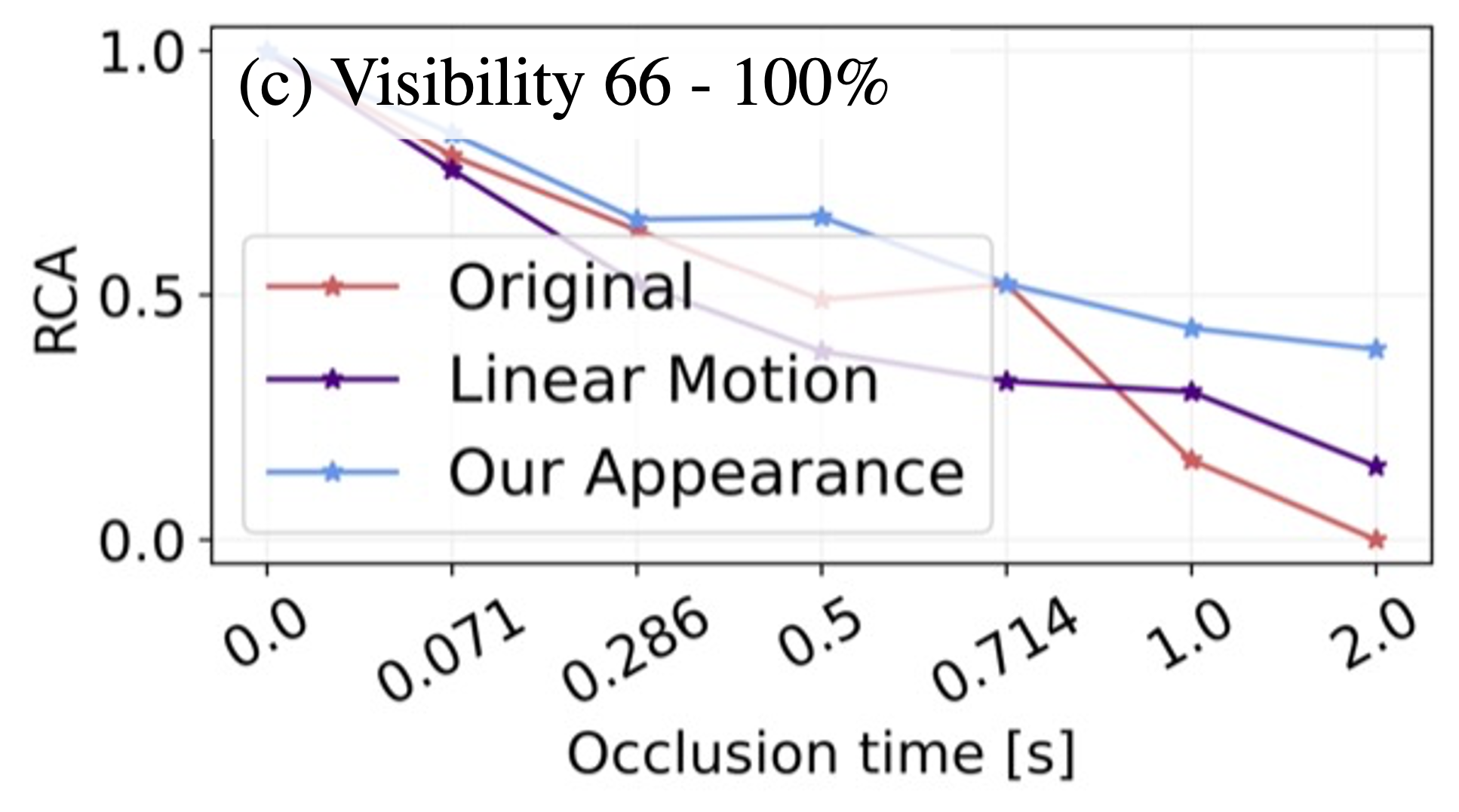}
    \end{subfigure}
    \vspace{-0.3cm}
    \caption{RCA with respect to different visibility levels for short-term vs. long-term associations.}
    \label{fig:by_visibility}
    \vspace{-0.1cm}
\end{figure*}

\begin{figure*}[ht!]
    \centering
    \begin{subfigure}[b]{0.29\textwidth}
        \centering
        \includegraphics[width=\textwidth]{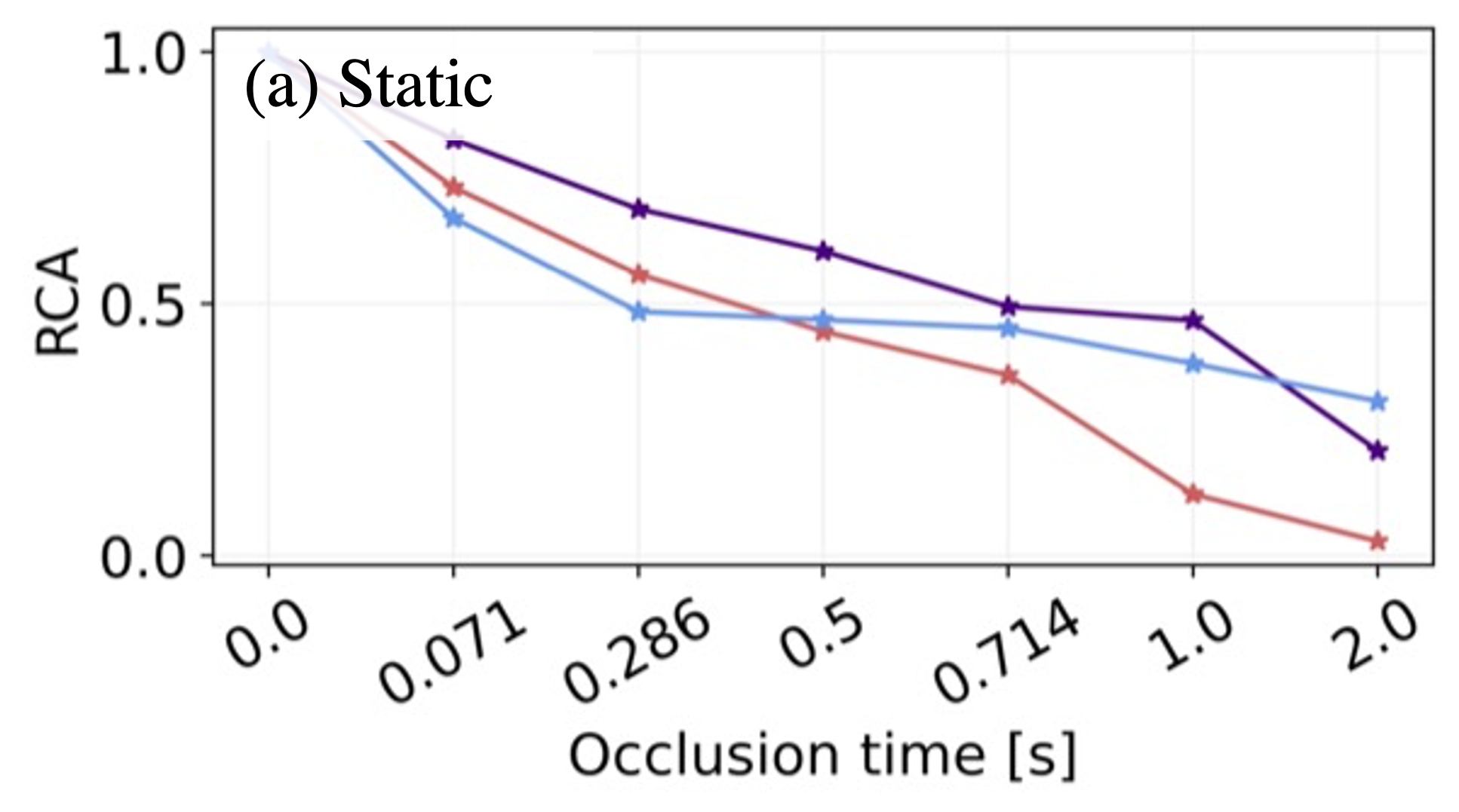}
    \end{subfigure}
    \quad
    \begin{subfigure}[b]{0.29\textwidth}   
        \centering 
        \includegraphics[width=\textwidth]{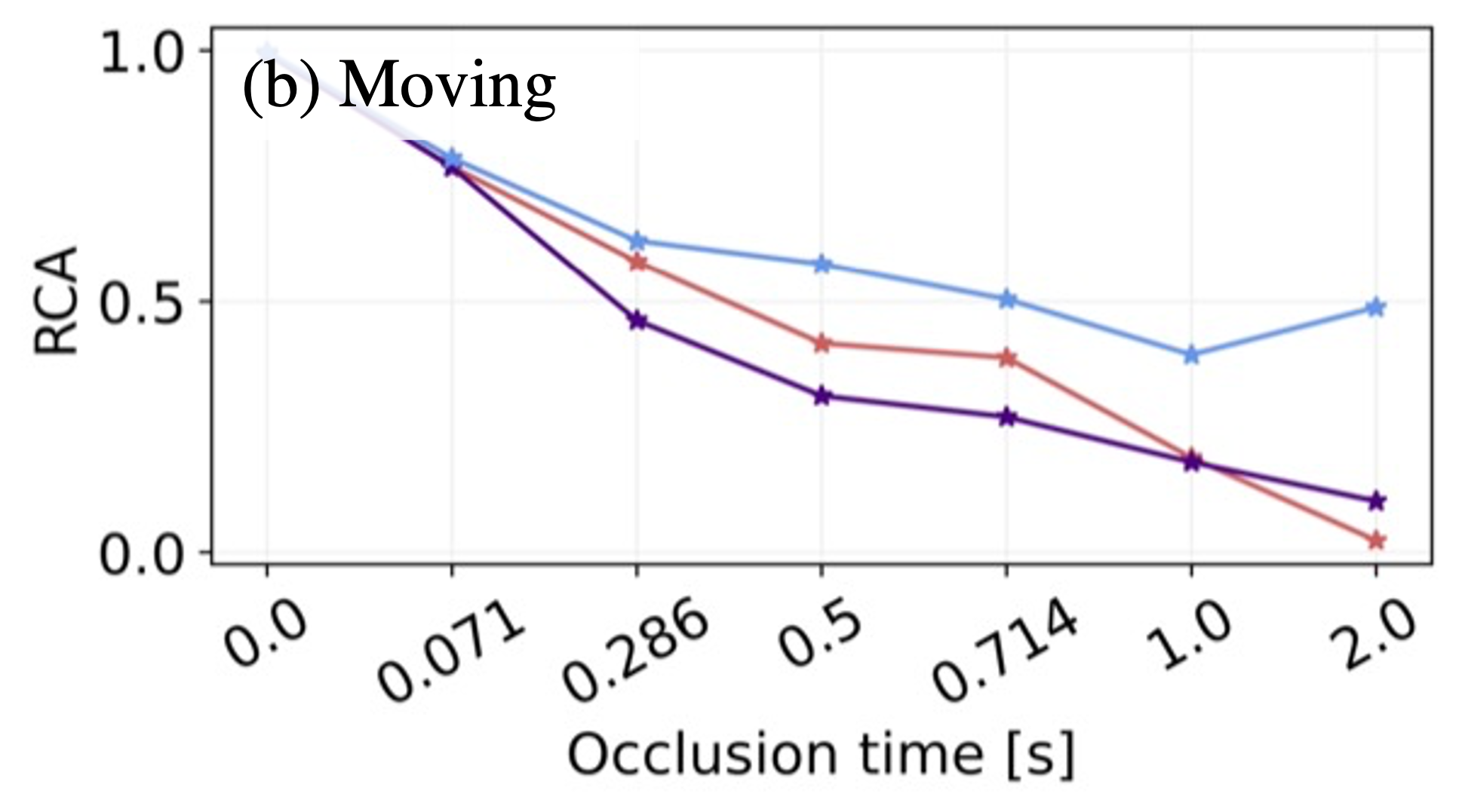}
    \end{subfigure}
    \quad
    \begin{subfigure}[b]{0.29\textwidth}   
        \centering 
        \includegraphics[width=\textwidth]{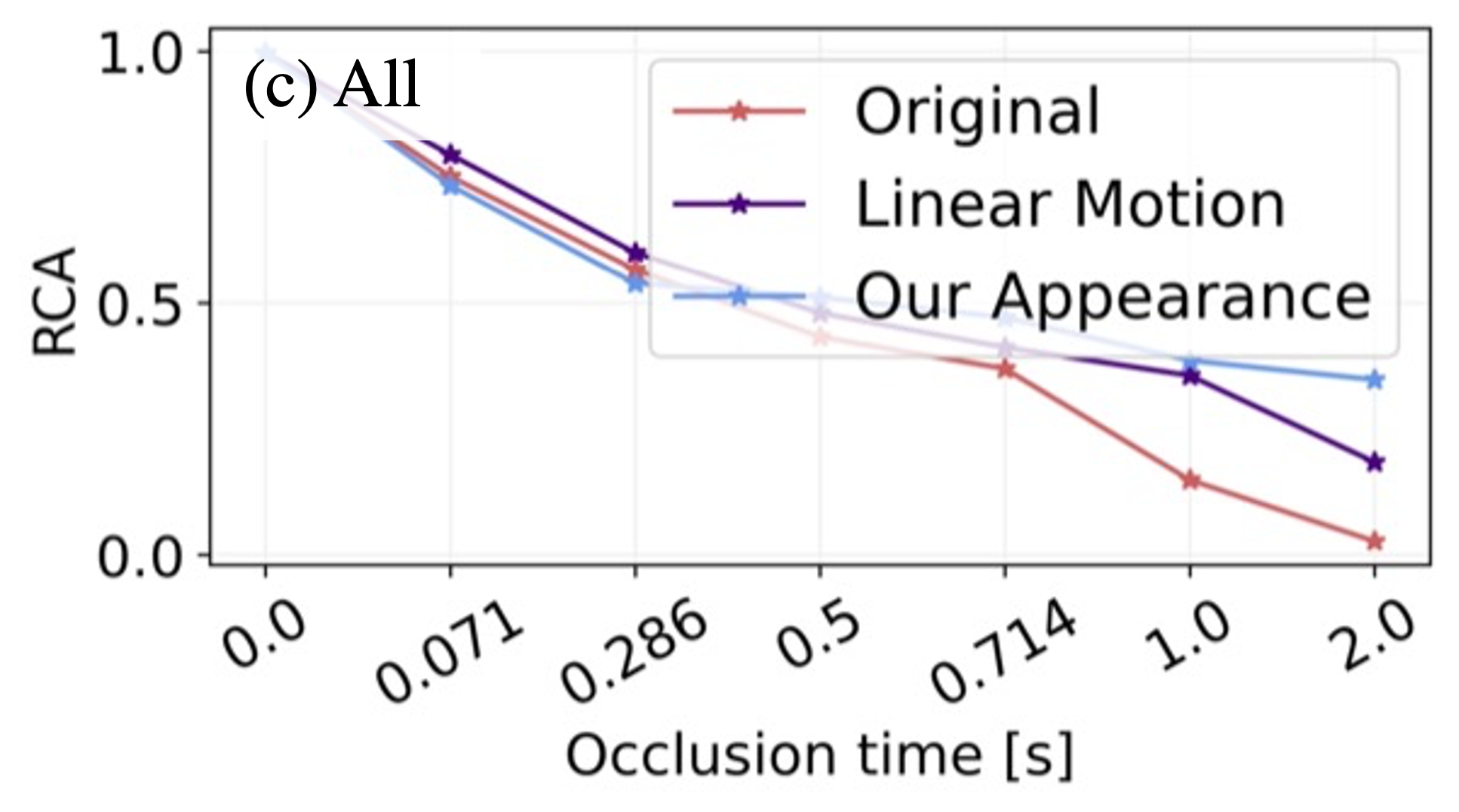}
    \end{subfigure}
    \vspace{-0.3cm}
    \caption{RCA for static, moving, and all sequences with respect to short-term vs. long-term associations.}
    \label{fig:seq_analysis_occ}
    \vspace{-0.1cm}
\end{figure*}

\begin{figure*}[ht!]
    \centering
    \begin{subfigure}[b]{0.29\textwidth}
        \centering
        \includegraphics[width=\textwidth]{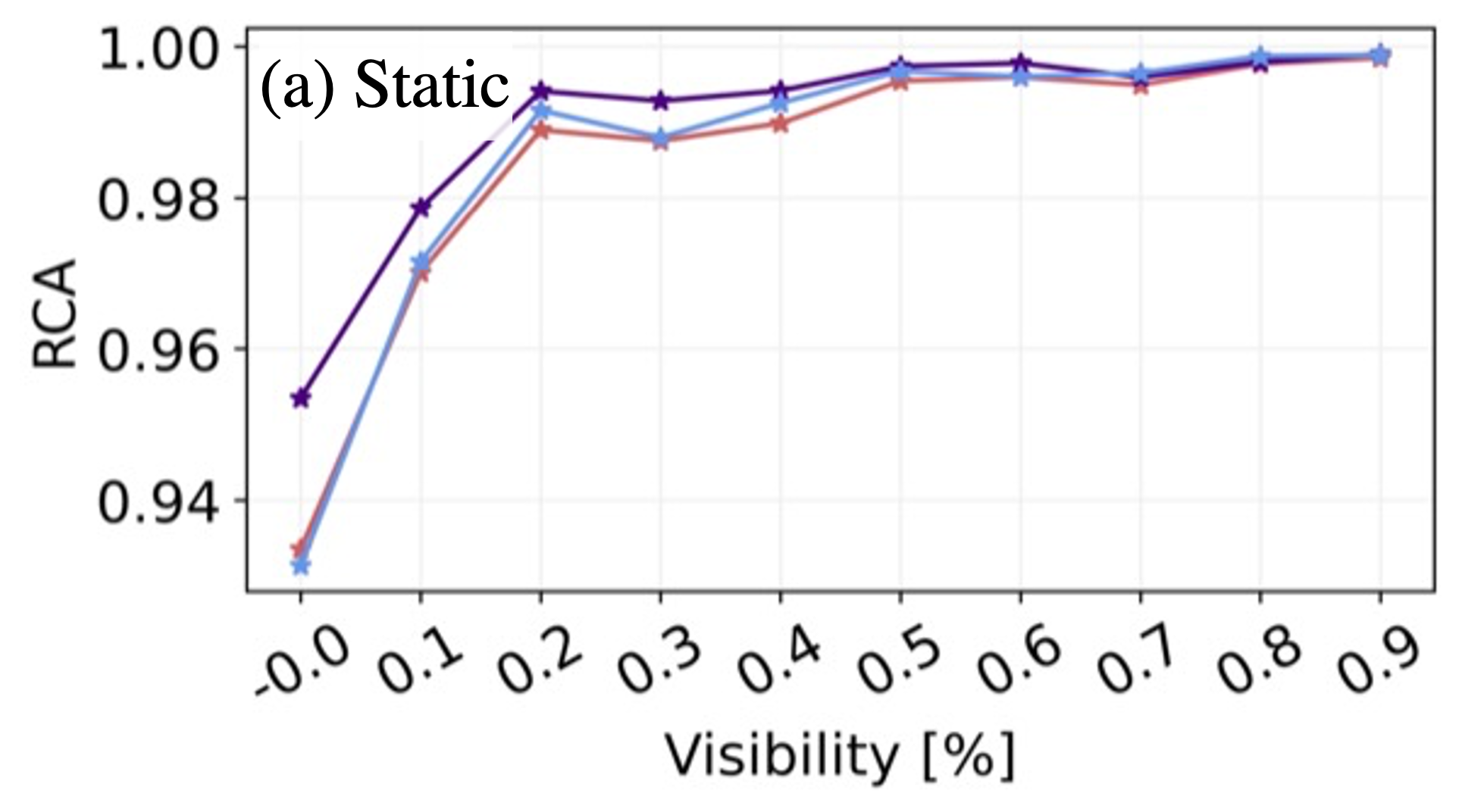}
    \end{subfigure}
    \quad
    \begin{subfigure}[b]{0.29\textwidth}   
        \centering 
        \includegraphics[width=\textwidth]{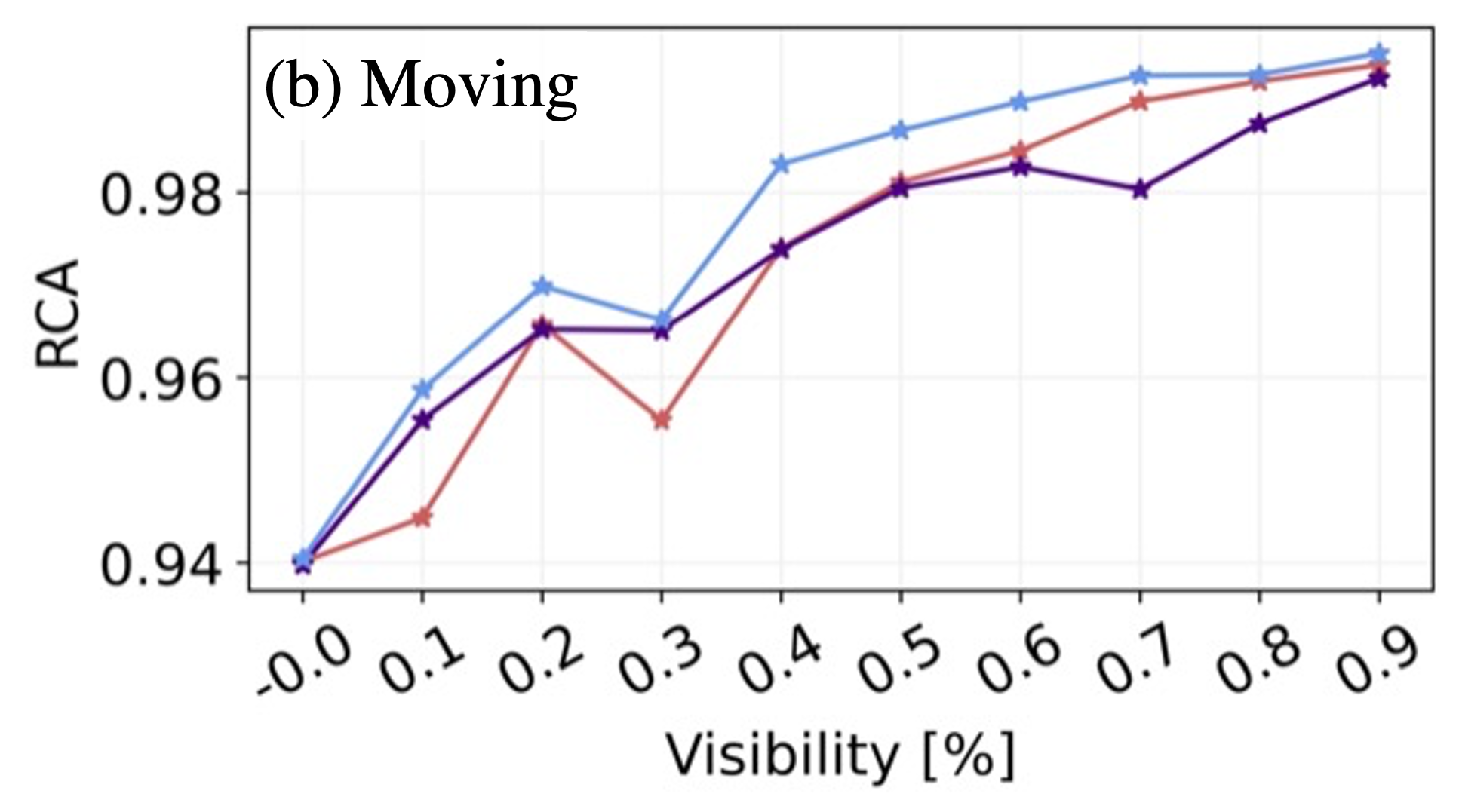}
    \end{subfigure}
    \quad
    \begin{subfigure}[b]{0.29\textwidth}   
        \centering 
        \includegraphics[width=\textwidth]{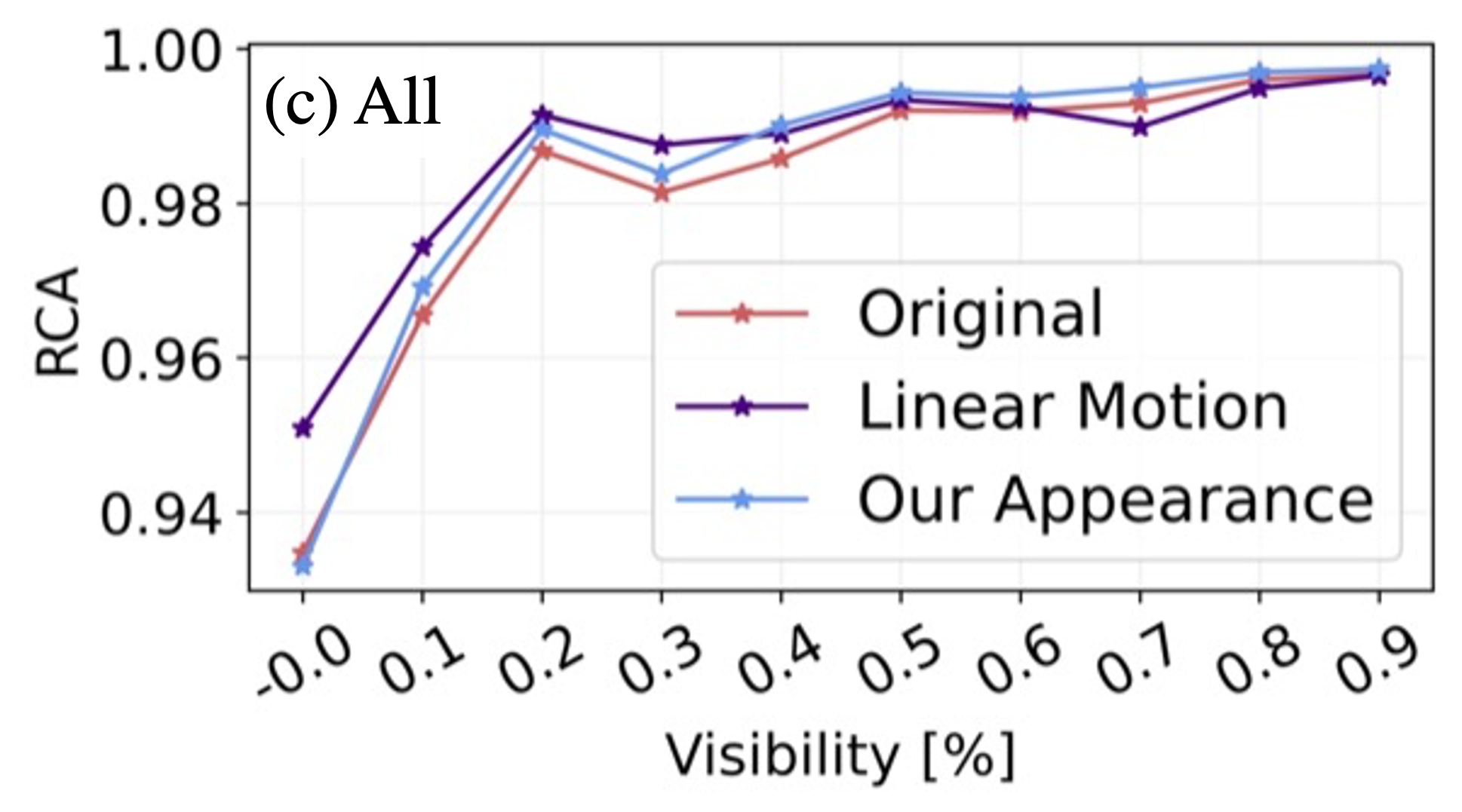}
    \end{subfigure}
    \vspace{-0.3cm}
    \caption{RCA for static, moving, and all sequences with respect to different visibility levels.}
    \label{fig:seq_analysis_vis}
    \vspace{-0.3cm}
\end{figure*}

\noindent\textbf{Observations.} 
We visualize RCA between detections and trajectories with increasing occlusion time for different visibility levels (Fig~\ref{fig:by_visibility}) as well as the performance of motion and appearance for static and moving sequences with respect to occlusion time and visibility (Fig~\ref{fig:seq_analysis_occ} and Fig~\ref{fig:seq_analysis_vis}). 
For highly visible bounding boxes (Fig~\ref{fig:by_visibility}(c)) appearance performs better than motion with respect to long- \textit{and} short-term associations.
While intuitively motion should perform especially well on short-therm associations independent of the visibility, Fig~\ref{fig:seq_analysis_vis} reveals that it struggles in moving sequences. This is due to the combination of the camera
movement and the bounding box movement which turns motion into being more non-linear.
On the other hand, in static sequences, the linear motion model performs better with respect to long-term associations than appearance (see Fig~\ref{fig:seq_analysis_occ}). 
This is caused by the fact that the lower the visibility (see Fig~\ref{fig:by_visibility}(a)), the higher the tendency of motion to perform better for long-term occlusions since motion is a strong cue in low visibility, \ie, occluded scenarios, (see Fig~\ref{fig:seq_analysis_vis}).
We show a more detailed analysis in the supplementary. 

\noindent\textbf{Conclusion.} In conclusion, the interplay of three factors mainly influences the performance of motion and appearance: visibility, occlusion time, and camera motion. However, we saw that appearance and motion complement each other well with respect to those factors.
Hence, we now move on to creating a strong tracker that combines our appearance and a simple linear motion model.

\subsection{Simple cues lead to a strong tracker}
\label{subsec:potential}
In this subsection, we ablate the combination of appearance and motion into a simple Hungarian-based online tracker.
%
%
We use the same setting as in Subsection~\ref{subsec:analysis}, \ie, we apply GHOST upon other trackers. Hence, we only report metrics related to the association performance, \ie, IDF1, and HOTA in Subsection~\ref{subsec:potential}, as our goal is not to improve on the MOTA metric, which is heavily dependent on detection performance.
We visualize the results in Fig~\ref{fig:private}, where markers and colors define which motion and appearance model is used, respectively. The blue bars represent the average occlusion level of detection bounding boxes of the different trackers.

\noindent\textbf{Appearance.} 
Compared to the performance of the original trackers, our appearance model improves the performance by up to 8.2pp in IDF1 and 4pp in HOTA for detection sets with lower average occlusion levels. In detection sets with high occlusion, pure appearance struggles, confirming that it is not suited for associations in those scenarios. 

\noindent\textbf{Motion.} Interestingly, applying only the simple linear motion model without appearance 
always improves or performs on par with the original trackers. Utilizing a Kalman filter instead of the linear motion model does not impact the performance significantly. This further highlights the strength of the simple linear motion model. 
While setting the number of previous positions to use to approximate a tracks current velocity $k$ to use all previous positions in this section, we generally found that moving camera sequences profit from a lower $k$ value due to the combination of the camera movement and the bounding box movement leading to less stable motion. The same holds for extreme movements, \textit{e.g.} as in the DanceTrack dataset.

\noindent\textbf{Combination.} We also visualize our appearance combined with linear motion or a Kalman Filter. Although we find that sequences with a moving camera profit from a lower motion weight while detections with high occlusion level profit from a higher motion weight, we fix the motion weight to $0.5$ in this experiment. Moreover, we visualize the performance of our appearance model combined with a Kalman filter. 
Fig~\ref{fig:private} shows that using a Kalman filter instead of the linear motion model does not impact the performance notably. 
However, both combinations improve significantly over using motion or our appearance alone.

\subsection{Comparison to State of the Art}
We compare GHOST to current state-of-the-art approaches. In all tables \textbf{bold} represents the best results, \textcolor{red}{red} the second best, and \textcolor{blue}{blue} the third best. 

\input{tables/public_detections}

\noindent\textbf{MOT17 Dataset.}
%
On \textit{public detections}, GHOST improves over the best previous methods, \textit{e.g.}, we improve over 
ArTIST-C \cite{DBLP:conf/cvpr/SalehARSG21} $1.8pp$ in HOTA (see Table~\ref{tab:sota}). As expected, we do not improve in MOTA, as it is mostly dependent on the detection performance. 
%
%
In the \textit{private detection setting} we perform on par with ByteTrack \cite{DBLP:journals/corr/abs-2110-06864} in HOTA and IDF1 outperforming second-best approaches by $3.7pp$ and $4.6pp$.

\noindent\textbf{MOT20 Dataset.}
Despite the complexity of the sequences being heavily crowded and exhibiting difficult lightning conditions, GHOST improves state-of-the-art \cite{DBLP:conf/iccv/BergmannML19} on \textit{public detection setting} of MOT20 by $2.6pp$ in IDF1 and $1.3pp$ in HOTA metric (see Table~\ref{tab:sota}).
This shows that the combination of our strong appearance cues and the linear motion adapts to the underlying conditions of higher occlusion levels.
%
%
In the \textit{private detection setting} of MOT20, we 
 outperform all other methods by up to $0.8pp$ on HOTA and $0.7pp$ on IDF1. 
GHOST is able to fully leverage the additional available bounding boxes in the private detection setting, substantially increasing identity preservation compared to the private setting.

\noindent\textbf{DanceTrack Dataset.}
Despite the claim of the corresponding paper \cite{sun2022dance} that reID and linear motions models are not suitable for DanceTrack, GHOST outperforms all other methods on the test set (see Table~\ref{tab:dance}). Surprisingly, Transformer-based methods like GTR \cite{DBLP:conf/cvpr/ZhouYKK22} and MOTR \cite{DBLP:journals/corr/abs-2105-03247} that are said to be general cannot transfer their performance from MOT17 and MOT20. We improve over prior state-of-the-art by \textit{2.5pp} on HOTA, \textit{3.8pp} on IDF1 and even \textit{0.7pp} on MOTA.
\input{tables/dancetrack}

\noindent\textbf{BDD100k Dataset.}
Despite our appearance model \textit{never} being trained on more than pedestrian images, it is able to generalize well to the novel classes. GHOST outperforms state-of-the-art in mHOTA and mIDF1 on the validation set by $0.3pp$ and $2pp$ and in mIDF1 on the test set by $1.2pp$ (see Table~\ref{tab:bdd_short}). In IDF1 metrics, QDTrack \cite{DBLP:conf/cvpr/PangQLCLDY21} outperforms GHOST. Since mHOTA and mIDF1 are obtained by averaging per-class IDF1 and HOTA while IDF1 and HOTA are achieved by averaging over detections, the results show that GHOST generalizes well to less frequent classes while other approaches like QDTrack \cite{DBLP:conf/cvpr/PangQLCLDY21} overfit to more frequent classes, \textit{e.g.}, car (see also per-class validation set results in the supplementary).

\input{tables/bdd_short}

%% file: tables/reid_balation.tex
\begin{table}[t!]
\centering
\resizebox{0.47\textwidth}{!}{
\begin{tabular}{@{}c|c|c|ccc|ccc@{}}
\hline

\multicolumn{3}{c}{} & \multicolumn{3}{|c|}{MOT 17} & \multicolumn{3}{|c}{BDD} \\
\hline

\textbf{diff $\tau$} & \textbf{IP} & \textbf{DA} & \textbf{HOTA} $\uparrow$ & \textbf{IDF1} $\uparrow$ & \textbf{MOTA} $\uparrow$ & \textbf{HOTA} $\uparrow$ & \textbf{IDF1} $\uparrow$ & \textbf{MOTA} $\uparrow$ \\
\hline
\hline
 & & & 61.6 & 72.2 & 69.5 & 41.3 & 47.7 & 42.0 \\
 \checkmark & & & 61.8 & 72.7 & 69.6 & 41.9 & 48.6 & 41.9  \\
 \checkmark & \checkmark & & 62.4 & 73.6 & 69.6 & 42.9 & 50.4 & 43.7 \\
 \checkmark & \checkmark & \checkmark & 63.3 & 75.3 & 69.6 & 43.7 & 51.5 & 43.9 \\
 
\hline
\end{tabular}}
\vspace{-0.1cm}
\caption{Ablation of the single parts of our method on the validation set of CenterTrack pre-processed bounding boxes. IP=inactive proxies, DA=domain adaptation.}
\vspace{-0.5cm}
\label{tab:reid_blation}
\end{table}

%% file: tables/da_ablation.tex
\begin{table}[t!]
\centering
\resizebox{0.4\textwidth}{!}{
\begin{tabular}{@{}l|ccc@{}}
\hline

& \textbf{HOTA} $\uparrow$ & \textbf{IDF1} $\uparrow$ & \textbf{IDF1} $\uparrow$ \\
\hline
\hline

BN update random patches  & 59.1 & 68.6 & 68.4 \\
BN update 10 frames before  & 62.0 & 73.0 & 68.6 \\
BN update whole sequence first & 62.4 & 73.8 & 69.1 \\
fine tuning on MOT17 & 63.0 & 74.3 & 69.6 \\
ResNet50 IBN & 63.2 & 74.6 & 69.7 \\
\hline
our domain adaptation & 63.2 & 75.4 & 69.7  \\
\hline
\end{tabular}}
\vspace{-0.1cm}
\caption{Ablation of different domain adaptation approaches.}
\vspace{-0.5cm}
\label{tab:domain_adaptation_ablation}
\end{table}

%% file: tables/public_detections.tex
\begin{table}[thb!]
\centering
\resizebox{0.385\textwidth}{!}{
\begin{tabular}{@{}lccccccc@{}}
\toprule

 & \textbf{HOTA} $\uparrow$ &  \textbf{IDF1} $\uparrow$ & \textbf{MOTA} $\uparrow$ & \textbf{IDSW} $\downarrow$ \\

\hline
\hline
\textit{Public MOT17} \\
Tracktor v2$^{\dagger}$ \cite{DBLP:conf/iccv/BergmannML19} & 44.8 & 55.1 & 56.3 & 1987  \\
UNS$^{\dagger}$ \cite{DBLP:journals/corr/abs-2111-05943} & \textcolor{black}{46.4} & \textcolor{black}{58.3} & \textcolor{black}{56.8} & 1914 \\
TrackPool$^{\dagger}$ \cite{DBLP:conf/cvpr/KimLAR21} & - & \textcolor{blue}{60.5} & 55.9 & \textcolor{red}{1188}   \\
CenterTrack$^{\ddagger}$ \cite{DBLP:conf/eccv/ZhouKK20}&\textcolor{blue}{48.2}&\textcolor{black}{59.6}& \textcolor{blue}{61.5} & 3039  \\
ArTIST-C$^{\ddagger}$ \cite{DBLP:conf/cvpr/SalehARSG21} & \textcolor{red}{48.9} & \textcolor{black}{59.7} & \textbf{62.3} & 2062  \\
\hline
GHOST$^{\dagger}$ & 47.4 & \textcolor{red}{60.6} & 56.5 & \textbf{1144} \\ 
GHOST$^{\ddagger}$ & \textbf{50.7} & \textbf{63.5} & \textcolor{red}{61.6} & \textcolor{blue}{1715} \\

\hline
\hline
\textit{Private MOT17} \\

CenterTrack \cite{DBLP:conf/eccv/ZhouKK20} & 52.2  & 64.7 & 67.8 & 3039\\
TraDeS \cite{DBLP:conf/cvpr/WuCS00Y21} & 52.7 & 63.9 & 69.1 & 3555\\
QDTrack \cite{DBLP:conf/cvpr/PangQLCLDY21} & 53.9 & 66.3 & 68.7 & 3378  \\
FairMOT \cite{DBLP:journals/ijcv/ZhangWWZL21} & \textcolor{red}{59.3} & \textcolor{blue}{72.3} & 73.7 & 3303  \\
MeMOT \cite{DBLP:conf/cvpr/CaiX0XXTS22} &  56.9 & \textcolor{red}{72.5} & 69.0 & 2724 \\
GTR \cite{DBLP:conf/cvpr/ZhouYKK22} & \textcolor{blue}{59.1} & 71.5 & \textcolor{blue}{75.3} & \textcolor{blue}{2859} \\
MOTR \cite{DBLP:journals/corr/abs-2105-03247} & 57.8 & 68.6 & 73.4 & 2439 \\
ByteTrack$^{\star}$ \cite{DBLP:journals/corr/abs-2110-06864} & \textbf{62.8} & \textbf{77.1} & \textbf{78.9}	& \textcolor{red}{2363} \\
\textcolor{gray}{ByteTrack$^{\star}$} \cite{DBLP:journals/corr/abs-2110-06864} & \textcolor{gray}{63.1} & \textcolor{gray}{77.3} & \textcolor{gray}{80.3} & \textcolor{gray}{2196} \\
\hline
GHOST & \textbf{62.8} & \textbf{77.1} & \textcolor{red}{78.7} & \textbf{2325} \\

\hline
\hline
\textit{Public MOT20} \\
GMPHD \cite{DBLP:journals/jvcir/Baisa21a} & 35.6 & 43.5 & 44.7 & 7492  \\ 
SORT \cite{DBLP:conf/icip/BewleyGORU16} &36.1  & 45.1 & 42.7 & 4470 \\
ArTIST$^{\dagger}$ \cite{DBLP:conf/cvpr/SalehARSG21} & \textcolor{blue}{41.6} & \textcolor{blue}{51.0} & \textbf{53.6} & \textcolor{red}{1531}  \\ 
Tracktor v2$^{\dagger}$ \cite{DBLP:conf/iccv/BergmannML19} & \textcolor{red}{42.1} & \textcolor{red}{52.7} & \textcolor{blue}{52.6} & \textcolor{blue}{1648} \\
\hline
GHOST$^{\dagger}$ & \textbf{43.4} & \textbf{55.3} & \textcolor{red}{52.7} & \textbf{1437}\\
\hline
\hline
\textit{Private MOT20} \\
GSDT \cite{DBLP:conf/icra/0002KW21} & 53.6 & 67.5 & 67.1 & 3230  \\
FairMOT \cite{DBLP:journals/ijcv/ZhangWWZL21} & \textcolor{blue}{54.6} & 67.3 & 61.8 & 5243  \\
MeMOT \cite{DBLP:conf/cvpr/CaiX0XXTS22} &  54.1 & 66.1 & \textcolor{blue}{63.7} & \textcolor{blue}{1938} \\
MTrack \cite{DBLP:conf/cvpr/YuLH22} & - & \textcolor{blue}{69.2} & 63.5 & 6031 \\
ByteTrack$^{\star}$ \cite{DBLP:journals/corr/abs-2110-06864} & \textcolor{red}{60.4} & \textcolor{red}{74.5} & \textbf{74.2}	& \textbf{925} \\
\textcolor{gray}{ByteTrack$^{\star}$} \cite{DBLP:journals/corr/abs-2110-06864} & \textcolor{gray}{61.3} & \textcolor{gray}{75.2} & \textcolor{gray}{77.8} & \textcolor{gray}{1223} \\
\hline
GHOST & \textbf{61.2} & \textbf{75.2} & \textcolor{red}{73.7} & \textcolor{red}{1264} \\ 
\bottomrule

\end{tabular}}
\vspace{-0.25cm}
\caption{Comparison to state-of-the-art for public and private detections on MOT17 and MOT20. $^{\dagger}$ and $^{\ddagger}$ indicate bounding boxes refined by Tracktor \cite{DBLP:conf/iccv/BergmannML19} and CenterTrack \cite{DBLP:conf/eccv/ZhouKK20}. $^{\star}$ since ByteTrack uses different thresholds for different sequences of the test set and interpolation we recomputed their results without both (recomputed black / \textcolor{gray}{original gray}).}
\vspace{-0.65cm}
\label{tab:sota}
\end{table}

%% file: tables/dancetrack.tex
\begin{table}[t!]
\centering
\resizebox{0.45\textwidth}{!}{
\begin{tabular}{@{}lccccc@{}}
\toprule

& \textbf{HOTA} $\uparrow$ & \textbf{IDF1} $\uparrow$ &  \textbf{MOTA} $\uparrow$  & \textbf{DetA} $\uparrow$ & \textbf{AssA} $\uparrow$\\
\hline
\hline
CenterTrack \cite{DBLP:conf/eccv/ZhouKK20} & 41.8 & 35.7  & 86.8 & \textcolor{blue}{78.1} & 22.6\\
FairMOT \cite{DBLP:journals/ijcv/ZhangWWZL21} & 39.7 & 40.8 & 82.2 & 66.7 & 23.8 \\
QDTrack \cite{DBLP:conf/cvpr/PangQLCLDY21} & \textcolor{red}{54.2} & 50.4  & \textcolor{blue}{87.7} & \textcolor{red}{80.1} & \textcolor{blue}{36.8}\\
TraDeS \cite{DBLP:conf/cvpr/WuCS00Y21} & 43.3 & 41.2 & 86.2 & 74.5 & 25.4 \\
MOTR \cite{DBLP:journals/corr/abs-2105-03247} & \textcolor{red}{54.2} & \textcolor{blue}{51.5} & 79.7 & 73.5 & \textbf{40.2} \\
GTR \cite{DBLP:conf/cvpr/ZhouYKK22} & \textcolor{blue}{48.0} & 50.3 & 84.7 & 72.5 & 31.9 \\
ByteTrack \cite{DBLP:journals/corr/abs-2110-06864} & 47.7  & \textcolor{red}{53.9} & \textcolor{red}{89.6} & 71.0 & 32.1 \\
\hline
GHOST & \textbf{56.7} & \textbf{57.7} & \textbf{91.3} & \textbf{81.1} & \textcolor{red}{39.8} \\
\bottomrule
\end{tabular}}
\vspace{-0.2cm}
\caption{Comparison to state-of-the-art on DanceTrack.}
\vspace{-0.3cm}
\label{tab:dance}
\end{table}

%% file: tables/bdd_short.tex
\begin{table}[t!]
\centering
\resizebox{0.45\textwidth}{!}{
\begin{tabular}{@{}lcccccc@{}}

\toprule

 &  \textbf{mHOTA} $\uparrow$  &  \textbf{mIDF1} $\uparrow$ & \textbf{mMOTA} $\uparrow$  &  \textbf{HOTA} $\uparrow$ &  \textbf{IDF1} $\uparrow$  &  \textbf{MOTA} $\uparrow$ \\
\hline\hline

\textit{validation} \\

Yu et. al. \cite{DBLP:conf/cvpr/YuCWXCLMD20} & - & 44.5 & 25.9 & - & 66.8 & 56.9 \\ 
ByteTrack \cite{DBLP:journals/corr/abs-2110-06864} & \textcolor{red}{45.4} & \textcolor{red}{54.6} & \textbf{45.2} & \textcolor{red}{61.6} & \textcolor{blue}{70.2} & \textbf{68.7} \\
QDTrack \cite{DBLP:conf/cvpr/PangQLCLDY21} & \textcolor{blue}{41.7} & 51.5 & 36.3 & \textcolor{blue}{60.9} & \textbf{71.4} & \textcolor{blue}{63.7} \\
MOTR \cite{DBLP:journals/corr/abs-2105-03247} & - & 43.5 & 32.0 & - & - & - \\
TETer \cite{DBLP:conf/eccv/LiDDHY22} & - & \textcolor{blue}{53.3} & \textcolor{blue}{39.1} & - & - & - \\

\hline

GHOST  & \textbf{45.7} & \textbf{55.6} & \textcolor{red}{44.9} &\textbf{61.7} & \textcolor{red}{70.9} & \textcolor{red}{68.1} \\
\hline
\hline
\textit{test} \\

Yu et. al. \cite{DBLP:conf/cvpr/YuCWXCLMD20} & - & 44.7 & 26.3 & - & 68.2 & 58.3 \\ 
ByteTrack \cite{DBLP:journals/corr/abs-2110-06864} & - & \textcolor{red}{55.8} & \textbf{40.1} & - & \textcolor{blue}{71.3} & \textbf{69.6} \\
QDTrack \cite{DBLP:conf/cvpr/PangQLCLDY21} & - & \textcolor{blue}{52.3} & \textcolor{blue}{35.5} & - & \textbf{72.3} & \textcolor{blue}{64.3} \\

\hline

GHOST  & \textbf{46.8} & \textbf{57.0} & \textcolor{red}{39.5} &\textbf{62.2} & \textcolor{red}{72.0} & \textcolor{red}{68.9} \\

\bottomrule

\end{tabular}}
\vspace{-0.25cm}
\caption{Comparison to state-of-the-art on BDD100k.}
\vspace{-0.65cm}
\label{tab:bdd_short}
\end{table}

%% file: chapters/Conclusion.tex
\vspace{-0.2cm}
\section{Conclusion}
\vspace{-0.15cm}
\label{sec:conclusion}
In this paper, we show that good old TbD trackers are able to generalize to various highly differing datasets incorporating domain-specific knowledge. 
For our general simple Hungarian tracker \textbf{GHOST}, we introduce a spiced-up appearance model that handles active and inactive trajectories differently.
Moreover, it adapts itself to the test sequences by applying an on-the-fly domain adaptation. 
We analyze where our appearance and simple linear motion model struggle with respect to visibility, occlusion time, and camera movement. 
Based on this analysis, we decide to use a weighted sum that gives more weight to either cue when needed, depending on the situations in the datasets.
Despite being straightforward, our insights have been largely overlooked by the tracking community. 
We hope to inspire future research to further investigate on, extend, and integrate these ideas into novel and more sophisticated trackers.

%% file: chapters/acknowledgement.tex
\noindent\textbf{Acknowledgements}
This work was partially funded by the Sofja Kovalevskaja Award of the Humboldt Foundation.

%% file: chapters/supp_extra.tex
\appendix

\textbf{{\larger[2] Supplementary Material}}
\newline

In this supplementary material, we first comment on novelty in science in Section~\ref{sec:novelty} before we show results on integrating GHOST into to tracking appeoaches in Section~\ref{sec:baseline}. The, we give the per-class performance of GHOST on BDD100k validation set in Section~\ref{sec:bdd}. In Section~\ref{sec:detailed_rca} we introduce the computation of the rate of correct associations (RCA) followed by a deeper analysis of it. We then conduct a deeper analysis of our domain adaptation in Section~\ref{sec:inter}. Afterwards, we show how we choose parameters based on our analysis in Section~\ref{sec:knowledge} and deeper investigate on the usage of different proxies in Section~\ref{sec:proxies}, the combination of appearance and motion using different weights for the sum in Section~\ref{sec:weights}, the number of frames to be used to approximate the velocity in Section~\ref{sec:num_frames} as well as on how to utilize different thresholds in Section~\ref{sec:threshs}. Also, we conduct experiments on different values of inactive patience in Section~\ref{sec:memory}. Then, we introduce how we generated the distance histograms in Section~\ref{sec:distance_dist}.
Moreover, in Section~\ref{sec:similar} we first outline the difference between our approach and trackers with similar components and compare the generality of our model to the generality of ByteTrack \cite{DBLP:journals/corr/abs-2110-06864} in Section~\ref{sec:generality}. Finally, we comment on the latency of our approach in Section~\ref{sec:latency}, and visualize several long-term occluded and low visibility bounding boxes on MOT17 public detection that GHOST successfully associates in Section~\ref{sec:vis}.

\section{"A painting can be beautiful even if it is simple and the technical complexity is low. So can a paper." \cite{OnNovelty}.}
\label{sec:novelty}
Inspired by the blog post of Michael Black on novelty in science \cite{OnNovelty}, we would like to discuss the common understanding of novelty in this paragraph.
Despite often confused, incremental changes do not necessarily mean that a paper can not introduce novelty and novel ideas. "If nobody thought to change that one term, then it is ipso facto novel. The inventive insight is to realize that a small change could have a big effect and to formulate the new loss" \cite{OnNovelty}.
Furthermore, it is of major importance to sometimes step back and formulate "a simple idea" since this "means stripping away the unnecessary to reveal the core of something. This is one of the most useful things that a scientist can do." \cite{OnNovelty}. If a simple idea improves the state of the art, "then it is most likely not trivial" \cite{OnNovelty}. 
Technical novelty is the most obvious type of novelty that reviewers look for in papers, but it is not the only one \cite{OnNovelty}.
In our understanding, if the reader takes away an idea from the paper that changes the way they do research, this can be considered a positive impact of the paper. Hence, the paper is novel (it has sparked a new idea in the reader’s mind). 
We hope readers also see it that way and we can progress with simpler, more interpretable, stronger models, and not only complex transformer-based pipelines trained on huge GPU farms. 
\input{tables_supp/bdd}
\begin{table}
\centering
\resizebox{0.45\textwidth}{!}{
\begin{tabular}{@{}l|l|ccc|ccc@{}}
\hline
& & \multicolumn{3}{c}{Original} & \multicolumn{3}{c}{Original + GHOST} \\
\hline
 & dataset & HOTA & IDF1 & MOTA & HOTA & IDF1 & MOTA\\
 \hline
ByteTrack & DanceTrack & 47.1 & 51.9 & 88.3 & 54.0 & 89.5 & 54.5 \\
Tracktor & MOT17 & 57.7 & 65.9 & 61.8 & 58.7 & 67.5 & 61.8 \\
\hline
\end{tabular}}
\caption{Applying GHOST in other Trackers.}
\label{tab:reb1}
\end{table}

\begin{figure*}[t]
    \centering
    \begin{subfigure}[b]{0.23\textwidth}
        \centering
        \includegraphics[width=\textwidth]{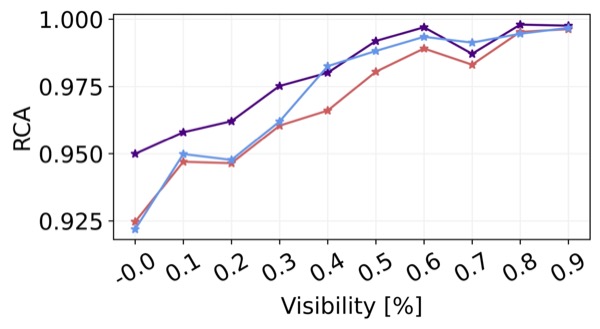}
        \caption{MOT17-02 \textit{S}}
    \end{subfigure}
    \quad
    \begin{subfigure}[b]{0.23\textwidth}   
        \centering 
        \includegraphics[width=\textwidth]{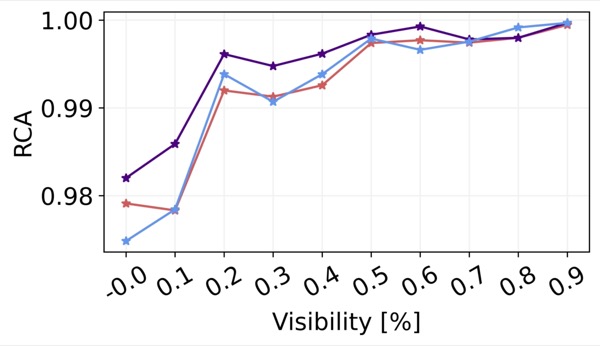}
        \caption{MOT17-04 \textit{S}}
    \end{subfigure}
    \quad
    \begin{subfigure}[b]{0.23\textwidth}   
        \centering 
        \includegraphics[width=\textwidth]{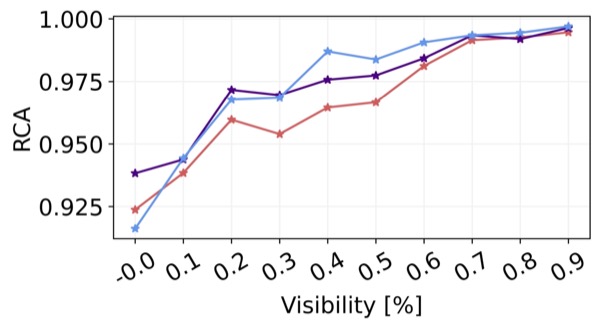}
        \caption{MOT17-05 \textit{M}}
    \end{subfigure}
    \quad
    \begin{subfigure}[b]{0.23\textwidth}   
        \centering 
        \includegraphics[width=\textwidth]{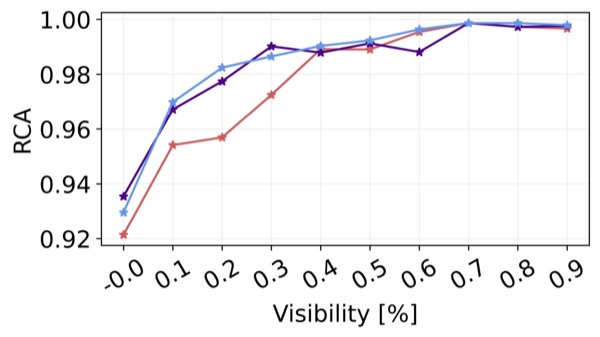}
        \caption{MOT17-09 \textit{S}}
    \end{subfigure}
    \quad
    \begin{subfigure}[b]{0.23\textwidth}   
        \centering 
        \includegraphics[width=\textwidth]{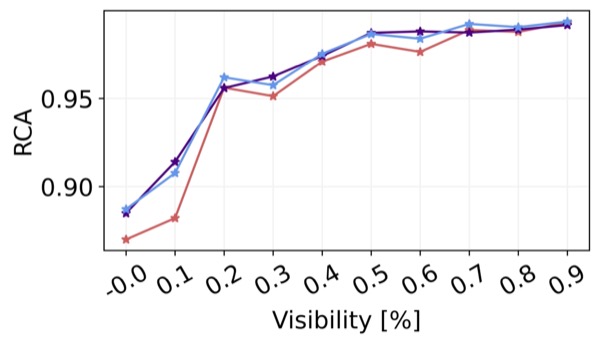}
        \caption{MOT17-10 \textit{M}}
    \end{subfigure}
    \quad
    \begin{subfigure}[b]{0.23\textwidth}   
        \centering 
        \includegraphics[width=\textwidth]{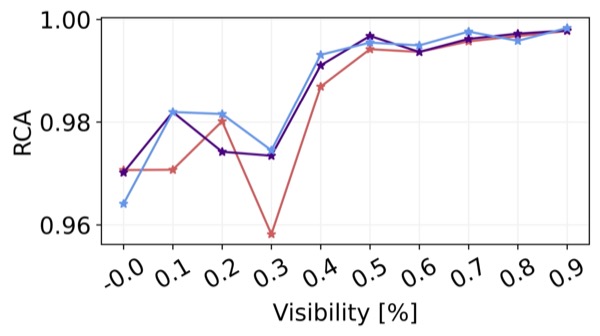}
        \caption{MOT17-11 \textit{M}}
    \end{subfigure}
    \quad
    \begin{subfigure}[b]{0.23\textwidth}   
        \centering 
        \includegraphics[width=\textwidth]{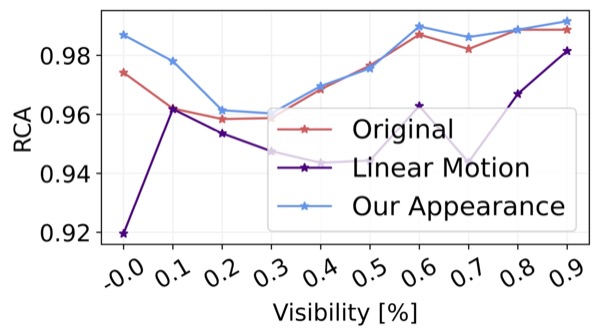}
        \caption{MOT17-13 \textit{M}}
    \end{subfigure}
    \caption{RCA for static \textit{S} and moving \textit{M} sequences with respect to visibility.}
    \label{fig:seq_analysis_vis}
\end{figure*}

\begin{figure*}[t]
    \centering
    \begin{subfigure}[b]{0.23\textwidth}
        \centering
        \includegraphics[width=\textwidth]{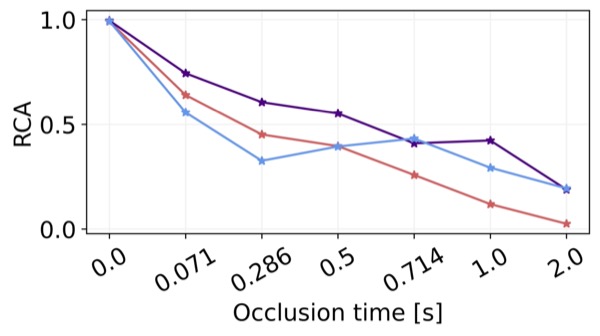}
        \caption{MOT17-02 \textit{S}}
    \end{subfigure}
    \begin{subfigure}[b]{0.23\textwidth}   
        \centering 
        \includegraphics[width=\textwidth]{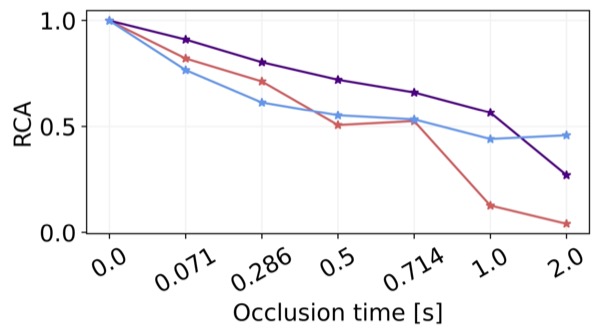}
        \caption{MOT17-04 \textit{S}}
    \end{subfigure}
    \begin{subfigure}[b]{0.23\textwidth}   
        \centering 
        \includegraphics[width=\textwidth]{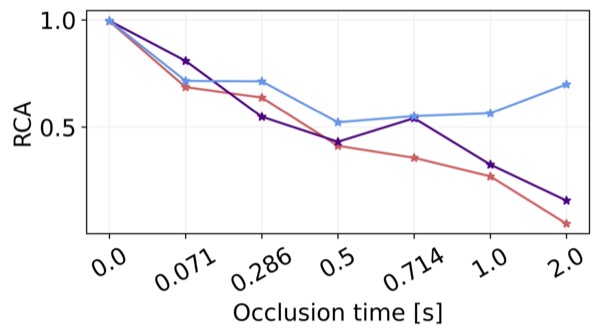}
        \caption{MOT17-05 \textit{M}}
    \end{subfigure}
    \begin{subfigure}[b]{0.23\textwidth}   
        \centering 
        \includegraphics[width=\textwidth]{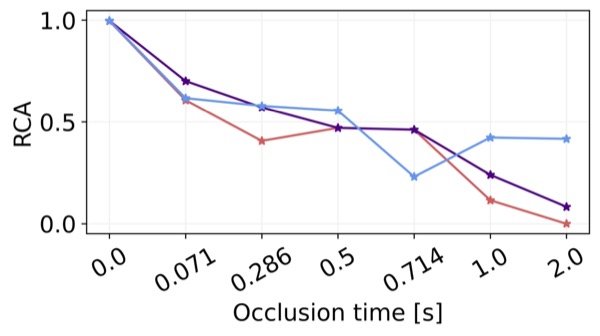}
        \caption{MOT17-09 \textit{S}}
    \end{subfigure}
    \begin{subfigure}[b]{0.23\textwidth}   
        \centering 
        \includegraphics[width=\textwidth]{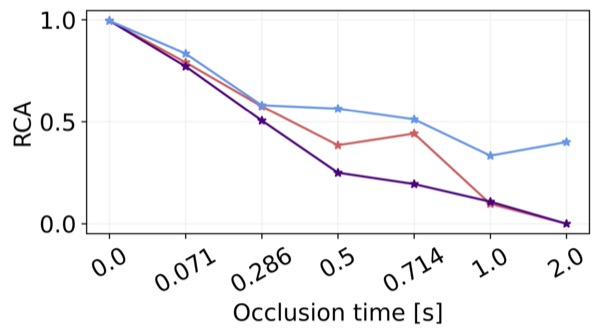}
        \caption{MOT17-10 \textit{M}}
    \end{subfigure}
    \begin{subfigure}[b]{0.23\textwidth}   
        \centering 
        \includegraphics[width=\textwidth]{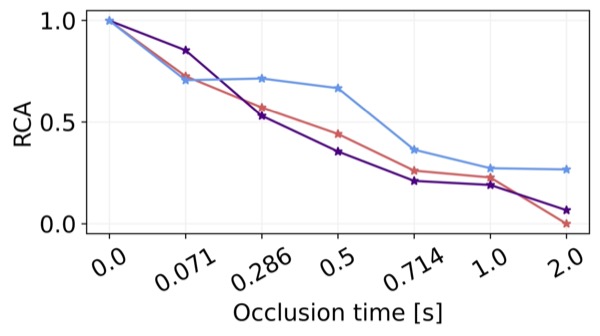}
        \caption{MOT17-11 \textit{M}}
    \end{subfigure}
    \begin{subfigure}[b]{0.23\textwidth}   
        \centering 
        \includegraphics[width=\textwidth]{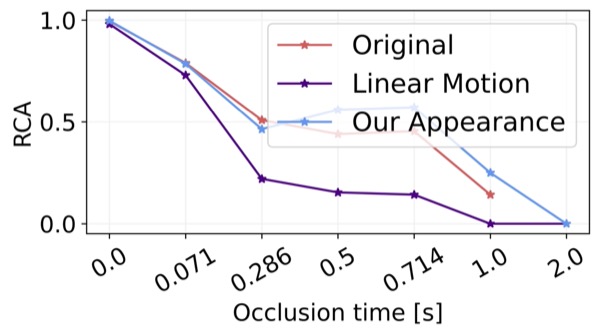}
        \caption{MOT17-13 \textit{M}}
    \end{subfigure}
    \caption{RCA for static \textit{S} and moving \textit{M} sequences with respect to short-term vs. long-term associations.}
    \label{fig:seq_analysis_occ}
\vspace{-0.3cm}
\end{figure*}

\section{Integrating GHOST within Other Trackers.} 
\label{sec:baseline}
Our baseline in the main paper is the simple Hungarian tracker in introduced in Sec 3.1. Furthermore, we apply GHOST additionally to other trackers as visualized in Fig 3 of the main paper. However, to show that it can also be integrated into existing trackers, in Tab~\ref{tab:reb1} we provide the performance of utilizing our reID instead of the baseline reID in Tracktor on MOT17. Since Tracktor on its own is a motion model we cannot apply our linear motion. Moreover, we provide results on utilizing our reID model and our linear motion instead of the Kalman Filter in ByteTrack on DanceTrack. Apart from the gain of using reID, the Kalman Filter struggles with the extreme motion while we can adapt the number of frames for velocity computation to the dataset.  

\section{Per-Class Evaluation on BDD100k Validation Set}
\label{sec:bdd}
In this section, we show the class-wise performance on the BDD100k validation set (Table~\ref{tab:bdd}). As on the test set (see main paper), we perform better than ByteTrack \cite{DBLP:journals/corr/abs-2110-06864} and QDTrack \cite{DBLP:conf/cvpr/PangQLCLDY21} in the overall IDF1 and HOTA metrics as well as in IDF1 and HOTA of less frequent classes like rider, bus, bycicle. 
On the other hand, QDTrack \cite{DBLP:conf/cvpr/PangQLCLDY21} outperforms us in the overall IDF1 metric per detection box, mainly due to their higher performance for highly frequent classes like car or pedestrian. 
This shows that QDTrack works well only for highly frequent classes, indicating a high dependency to the train set. Note, our model is only trained on the pedestrian class, which makes our performance on other classes a good demonstration of the generality of our approach. 

\section{Detailed Analysis of Rate of Correct Associations per Sequence}
\label{sec:detailed_rca}
We show a per sequence analysis of the rate of correct associations (RCA) of motion and appearance with respect to visibility in Fig~\ref{fig:seq_analysis_vis}, and short-term vs. long-term association in Fig~\ref{fig:seq_analysis_occ} where \textit{M} indicates moving sequence and \textit{S} indicates static sequence. For this we first introduce how to compute the RCA value.

\noindent\textbf{Computation of RCA.}
Given the output file of a tracker, to compute the rate of correct associations (RCA), we first match the given detections to ground truth identities following the same matching as the one used for the computation of the MOTA metric \cite{DBLP:journals/pami/KasturiGSMGBBKZ09}. 
For each detection $o_i$, we then find the last previous detection that belongs to the same ground truth ID $o_{i, prev}$. If $o_{i}$ was assigned to the same tracker ID as $o_{i, prev}$, we count it as a true positive association (TP-Ass), and if it was assigned to a different tracker ID, we count it as a false positive association (FP-Ass). This leads to the RCA value:
\begin{equation}
    RCA = \frac{\text{TP-Ass}}{\text{FP-Ass} + \text{TP-Ass}},
\end{equation}
To get the performance for different visibility levels and occlusion time, we organize $o_i$ into bins. For example, when we investigate the performance for visibility $0-33\%$, we take only detections $o_i$ into account that are $0-33\%$ visible. The same holds for occlusion time: if we investigate occlusion time $0.5-0.7s$, we only take detections $o_i$ into account whose prior detection of the same ground truth ID was $0.5-0.7s$ ago.
This procedure allows us to investigate the performance of different trackers with respect to different influencing factors solely based on the detection output files. 

\begin{figure*}[htp!]
    \centering
    \begin{subfigure}[b]{0.45\textwidth}
        \centering
        \includegraphics[width=\textwidth]{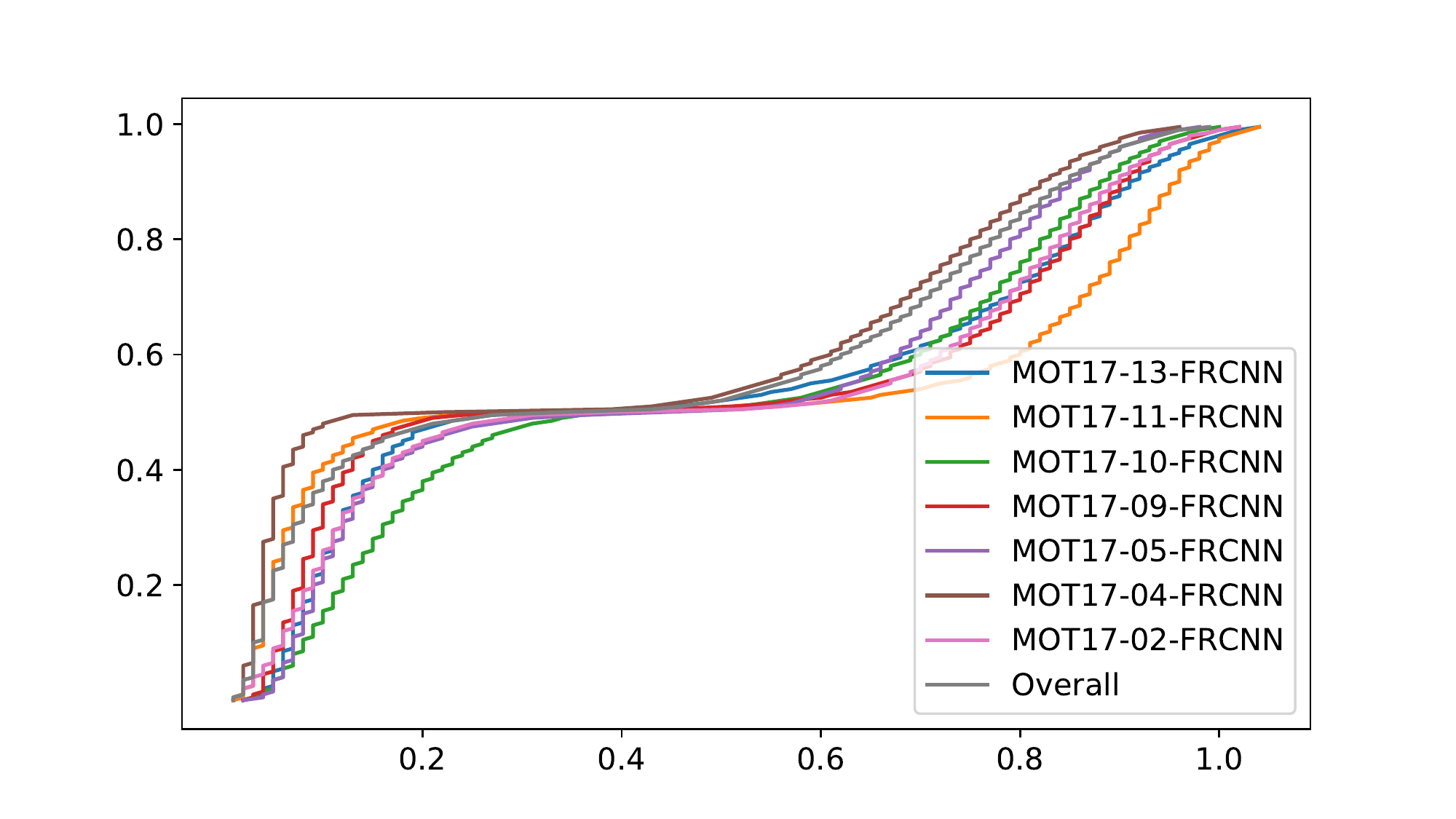}
        \caption{Without on-the-fly domain adaptation.}
        \label{fig:quantilles_act_without}
    \end{subfigure}
    \quad
    \begin{subfigure}[b]{0.45\textwidth}   
        \centering 
        \includegraphics[width=\textwidth]{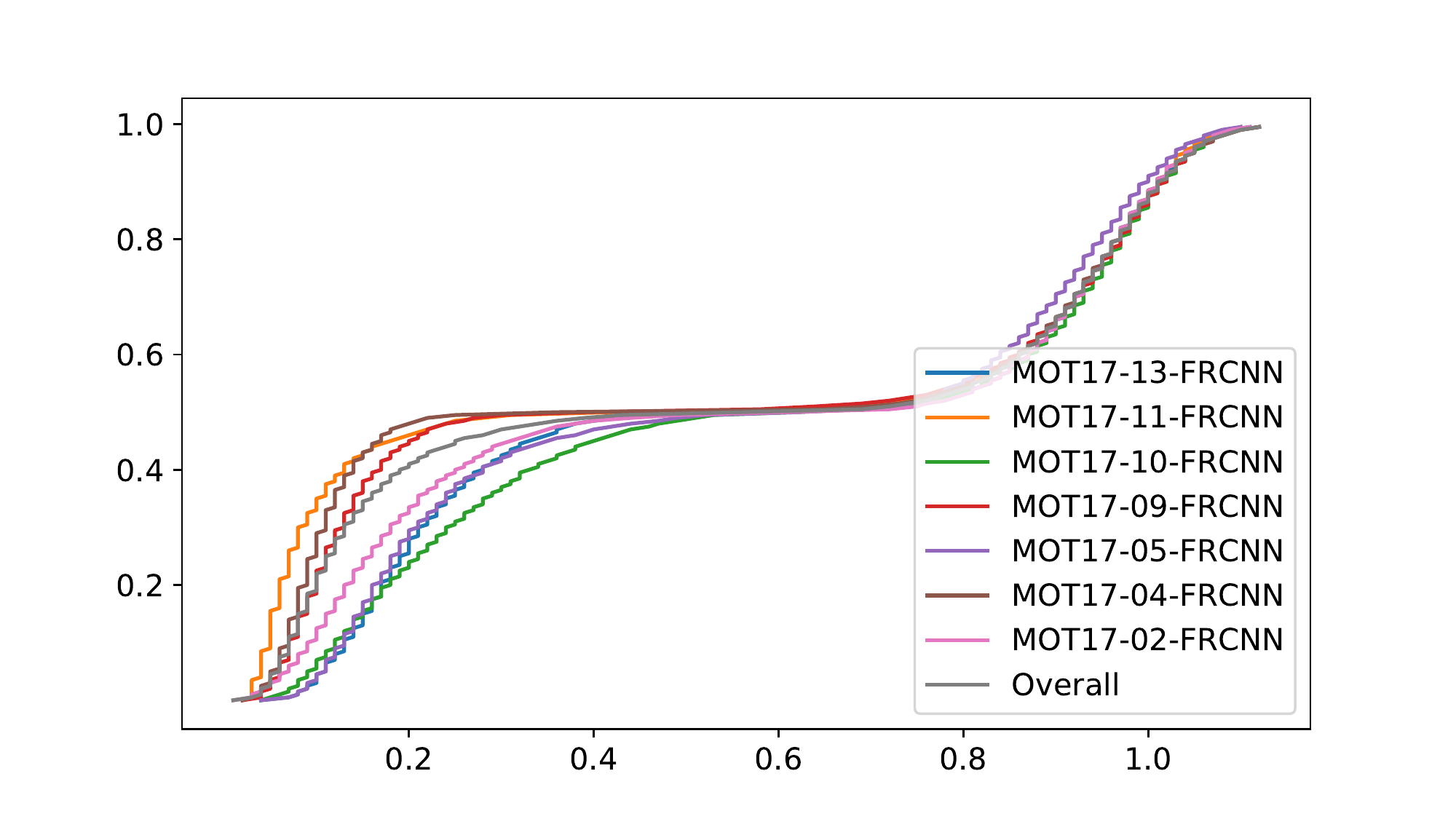}
        \caption{With on-the-fly domain adaptation.}
        \label{fig:quantilles_act_with}
    \end{subfigure}
    \caption{Cumulative sum of absolute bin difference between $f_{a,d}$ and $f_{a,s}$ on MOT17 validation set.}
    \label{fig:quantilles_act}
    \vspace{-0.3cm}
\end{figure*}

\begin{figure*}[htp!]
    \centering
    \begin{subfigure}[b]{0.45\textwidth}
        \centering
        \includegraphics[width=\textwidth]{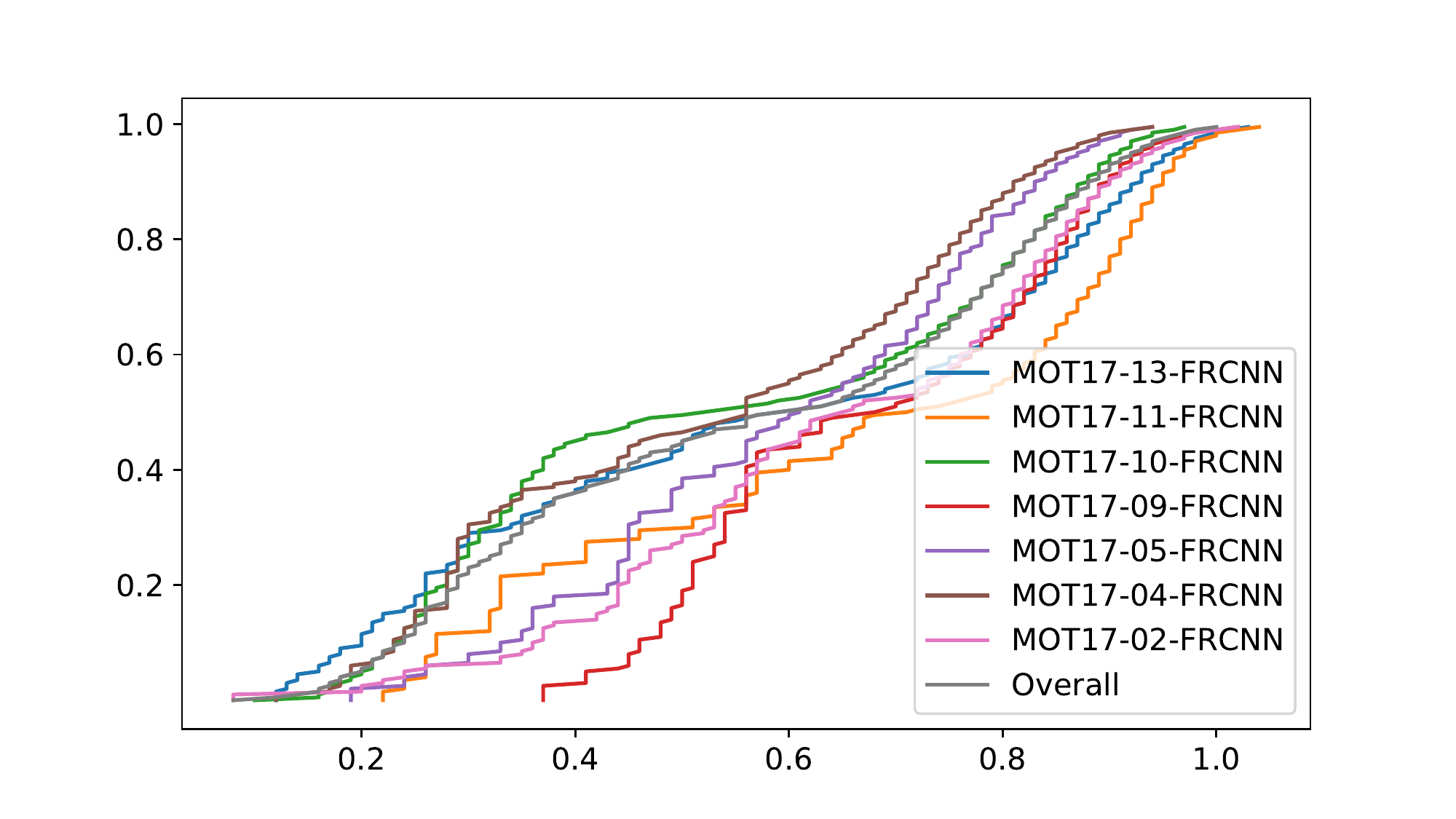}
        \caption{Without on-the-fly domain adaptation.}
        \label{fig:quantilles_inact_without}
    \end{subfigure}
    \quad
    \begin{subfigure}[b]{0.45\textwidth}   
        \centering 
        \includegraphics[width=\textwidth]{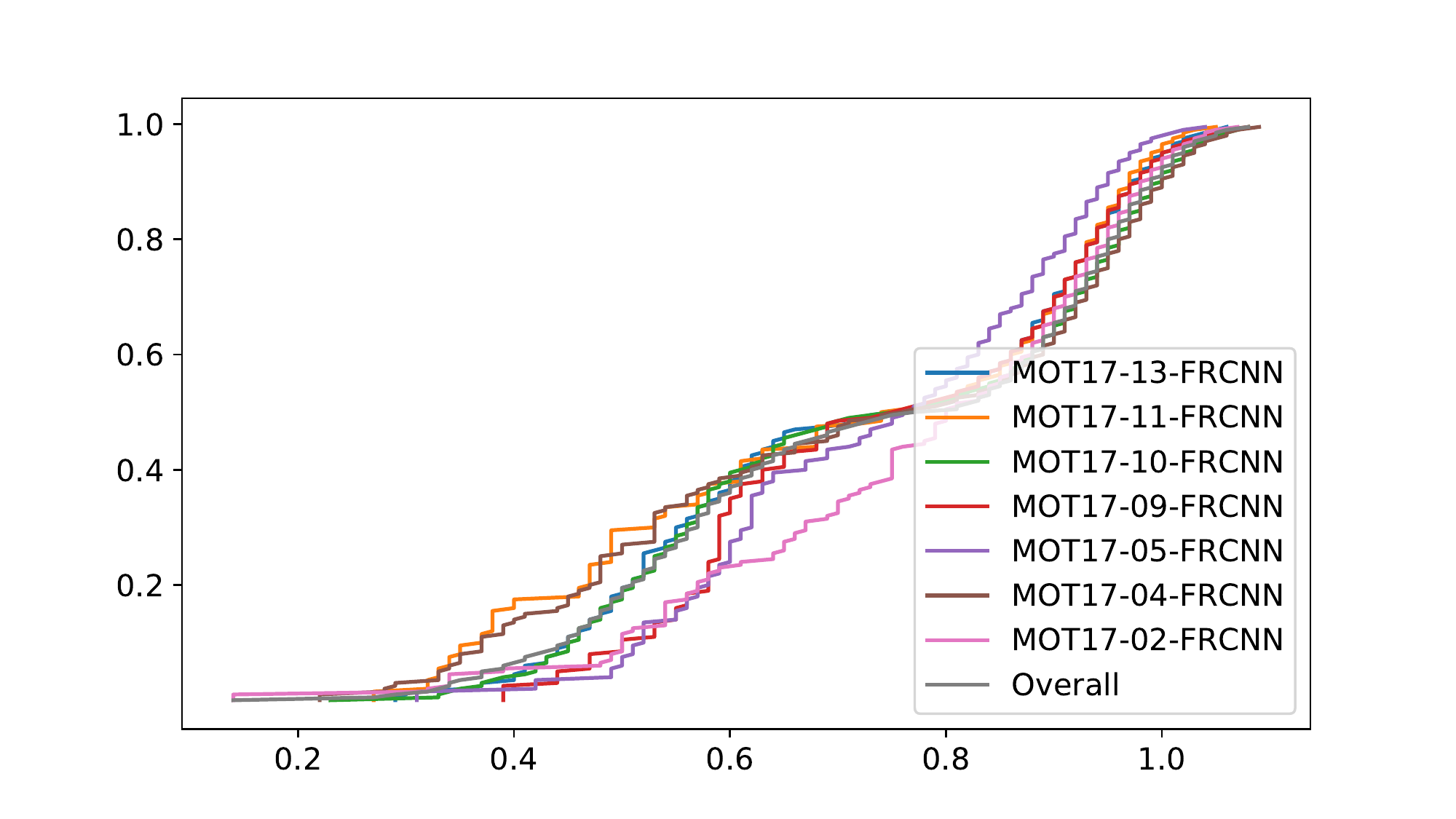}
        \caption{With on-the-fly domain adaptation.}
        \label{fig:quantilles_inact_with}
    \end{subfigure}
    \caption{Cumulative sum of absolute bin difference between $f_{a,d}$ and $f_{a,s}$ on MOT17 validation set.}
    \label{fig:quantilles_inact}
    \vspace{-0.6cm}
\end{figure*}

\begin{figure}[htp!]
    \centering
    \includegraphics[width=0.4\textwidth]{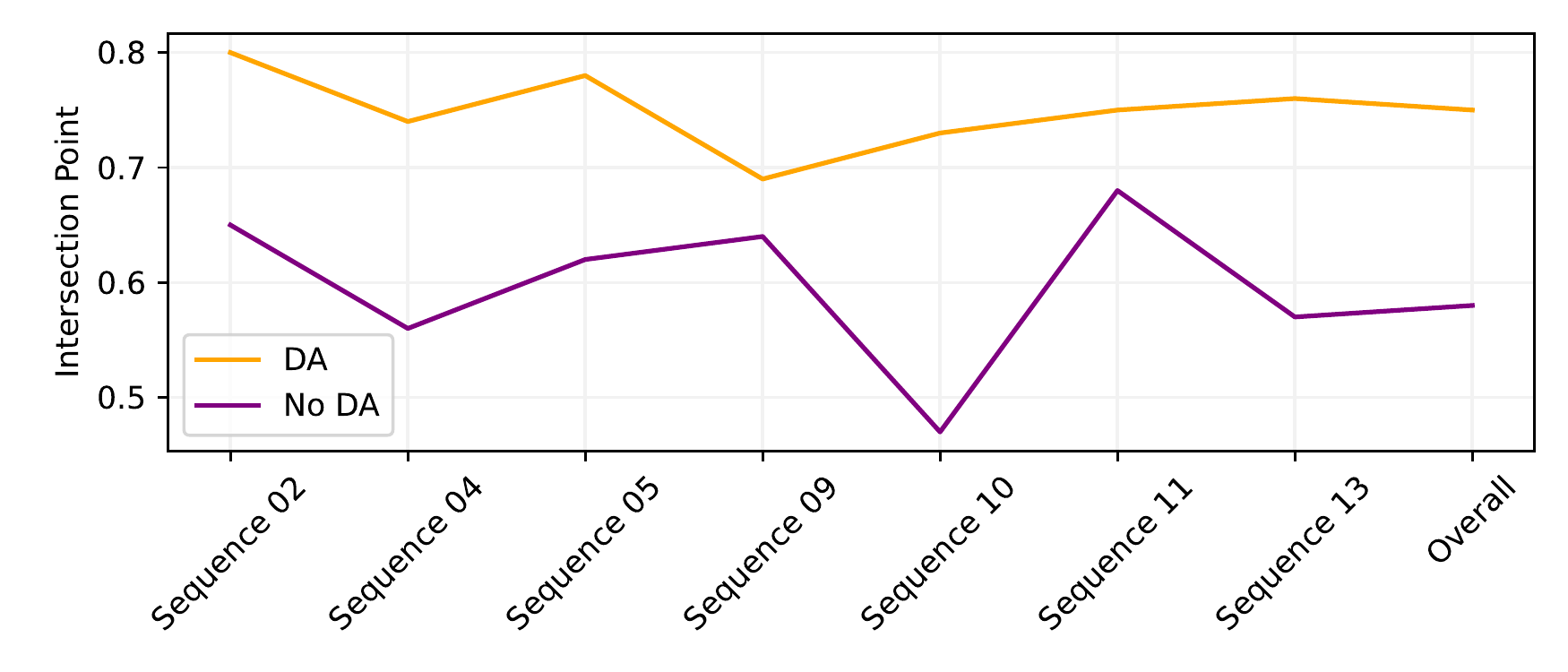}
    \caption{Visualization of the intersection points between distance histograms from detections to inactive tracks of the same and different identities when using domain adaptation (DA) compared to when not using it.}
    \label{fig:intersection_2}
\vspace{-0.6cm}
\end{figure}

\noindent\textbf{Visibility.} Motion cues perform better especially in the static sequences MOT17-02 and MOT17-04. In the static sequence MOT17-09, which is recorded from a low viewpoint, and the moving sequences MOT17-05, MOT17-10, and MOT17-11, motion and appearance perform approximately on par. 
In MOT17-13, which shows heavy camera movements, the performance of the motion model drops significantly.
Those observations show that for suitable camera angles in static sequences motion outperforms appearance independent of the visibility, while for sequences with severe camera movement or unsuitable camera angles, appearance outperforms motion even for low visibility scenarios.
For moving camera sequences, the motion of the object and the camera add up, resulting in more noisy and non-linear motion observed in pixel space, even though the underlying motion might be linear. 
Similarly, a low viewpoint leads to a distorted observation of the underlying motion from the camera perspective. 
When the camera angle comes closer to a bird's eye view perspective (MOT17-04) the observed motion is less distorted.

\noindent\textbf{Occlusion time.} Fig~\ref{fig:seq_analysis_occ} shows that all moving sequences show a higher RCA for appearance than motion cues. For static sequences, motion performs slightly better in MOT17-02 and MOT17-04. In the static sequence MOT17-09, the sequence recorded from a low viewpoint, both perform approximately on par. 
For suitable camera angles motion is a good cue even for long-term associations in static sequences, while appearance outperforms motion even for short-term associations in moving ones. This stems from the fact that motion gets more non-linear observed from camera perspective in moving camera sequences. While appearance still suffers from occlusion in static sequences even when recorded from a well-suited camera angle, those conditions allow for surprisingly well performance of motion even with respect to long-term associations.

\section{A Deeper Analysis on on-the-fly Domain Adaptation}
\label{sec:inter}
In the main paper, we visualize the distance histograms between active and inactive tracks to new detections of the same or different classes (see Fig 2 main paper). In this section, we show that the \textit{intersection point} that divides the distance histograms between active (inactive) tracks of the same and different classes well varies less over the different sequences when using our on-the-fly domain adaptation (see Fig 2(b) in main paper) compared to when not using it (see Fig 2(a) in main paper). Furthermore, we show that the distributions are generally more \textit{similar and stable} over different sequences with out on-ht-fly domain adaptation.

\noindent\textbf{Intersection Points.} Given the distribution $f_{a,d}$ of distances between active tracks ($a$) and new detections of a different ID ($d$) and the distribution $f_{a,s}$ of distances between active tracks and new detections of the same ID ($s$), we find a well suited intersection point ${x_{s,d}^a}^*$ separating both distributions by minimizing the sum of the costs of false positive and false negative matches.
Towards this end, for a given point $x_{s,d}^a$ we define the false positive costs as percentile value of $f_{a,d}$ at $x_{s,d}^a$ given by $p_{a,d}^{x_{s,d}}$, \textit{i.e.}, the percentile of $f_{a,d}$ that lies left to $x_{s,d}^a$. We define the false negative cost for $x_{s,d}^a$ as $100-p_{a,s}^{x_{s,d}}$ utilizing the percentile value of $f_{a,s}$ at $x_{s,d}^a$ since we want to punish the false negatives that lie to the right of this point.
%
Similar points $x_{s, d}^{a}$ across sequences allow to choose one single well-suited threshold $\tau_{act}$ over all sequences. The same holds for the inactive track distributions $f_{i,s}$ and $f_{i,d}$ and the corresponding $x_{s, d}^{i}$ of the different sequences. As we visualize in Fig~\ref{fig:intersection_2}, $x_{s, d}^{i}$ varies significantly less across tracking sequences when using our on-the-fly DA compared to when not using it. 

\noindent\textbf{Similarity and Stability.} 
While the variance of $x_{s, d}^{a}$ is higher for both settings, \textit{i.e.}, with and without the on-the-fly DA, the distributions are generally more separated when using DA compared to when not using it. To show this, we conduct a second experiment. 
For each sequence, we define the same bins in the range from $0-1$ and turn the distributions $f_{a,d}$ and $f_{a,s}$ into histograms $h_{a,d}$ and $h_{a,s}$. In each bin $i$ we compute the absolute difference between the two histograms $d_{a, i}$ and normalize it by the sum of all absolute distances. Finally, we plot the cumulative histogram (see Fig~\ref{fig:quantilles_act}). The more aligned the cumulative histograms over all sequences and the broader the saddle point, the more similar the sequences across each other and the better separated the distributions of the same and different IDs, respectively. Note, that the cumulative sums of the different sequences are much more aligned when utilizing on-the-fly domain adaptation (see Fig~\ref{fig:quantilles_act_with}) than when not using it (see Fig~\ref{fig:quantilles_act_without}) which makes it easier to find a common threshold $\tau_{act}$.  Moreover, the saddle point is much broader when using on-the-fly domain adaptation which makes our approach more stable with respect to different thresholds $\tau_{act}$.

Despite the difference visualization for the inactive track distributions being less unified over the sequences in general, the differences over the different sequences when using on-the-fly domain adaptation are more aligned compared to when not using it (see Fig~\ref{fig:quantilles_inact}). Combined with the less varying ${x_{s, d}^{i}}^*$, this leads not only to an overall better suited but also more stable threshold $\tau_{inact}$. 

\section{Using the Knowledge of our Analysis.}
\label{sec:knowledge}
In our work we present an in-depth analysis on appearance distance computation based on embedding features (see Fig 2 in the main paper) as well as motion vs. appearance model performance (see Fig 3-6 in the main paper). Based on those insights we introduce our simple tracker GHOST. For example, we utilize the analysis of the differences between the reID distance of active and inactive tracks to detections to adapt the thresholds and and choose a proxy distance computation method. Also, we utilize the inisghts that reID performs worse for high occlusion levels and linear motion performs worse in moving camera scenes and with extreme motion to adapt the motion weight as well as the number of frames used in the motion model. We do not only present GHOST but also a large number of analysis that reveal insights for the community. 
%
In the following we provide a deeper analysis on the hyperparameters and design choices of GHOST for the single datasets.

\begin{table*}
\centering
\resizebox{0.65\textwidth}{!}{
\begin{tabular}{@{}l|cccc@{}}
\hline
& BDD & DanceTrack & MOT17 & MOT20\\
\hline
motion & moving cam & extreme motion & partially moving cam & static cam \\
occlusion & medium & medium & medium & high \\
motion weight & 0.4 & 0.4 & 0.6 & 0.8 \\
\# frames motion model & 10 & 5 & 90 & 30 \\
\hline
\end{tabular}}
\vspace{-0.2cm}
\caption{Motion Model Parameters.}
\vspace{-0.4cm}
\label{tab:reb2}
\end{table*}

\vspace{-0.1cm}
\subsection{The Usage of Different Proxies}
\vspace{-0.1cm}
\label{sec:proxies}
\begin{figure*}[t]
    \centering
    \begin{subfigure}[b]{0.49\textwidth}
        \centering
        \includegraphics[width=\textwidth]{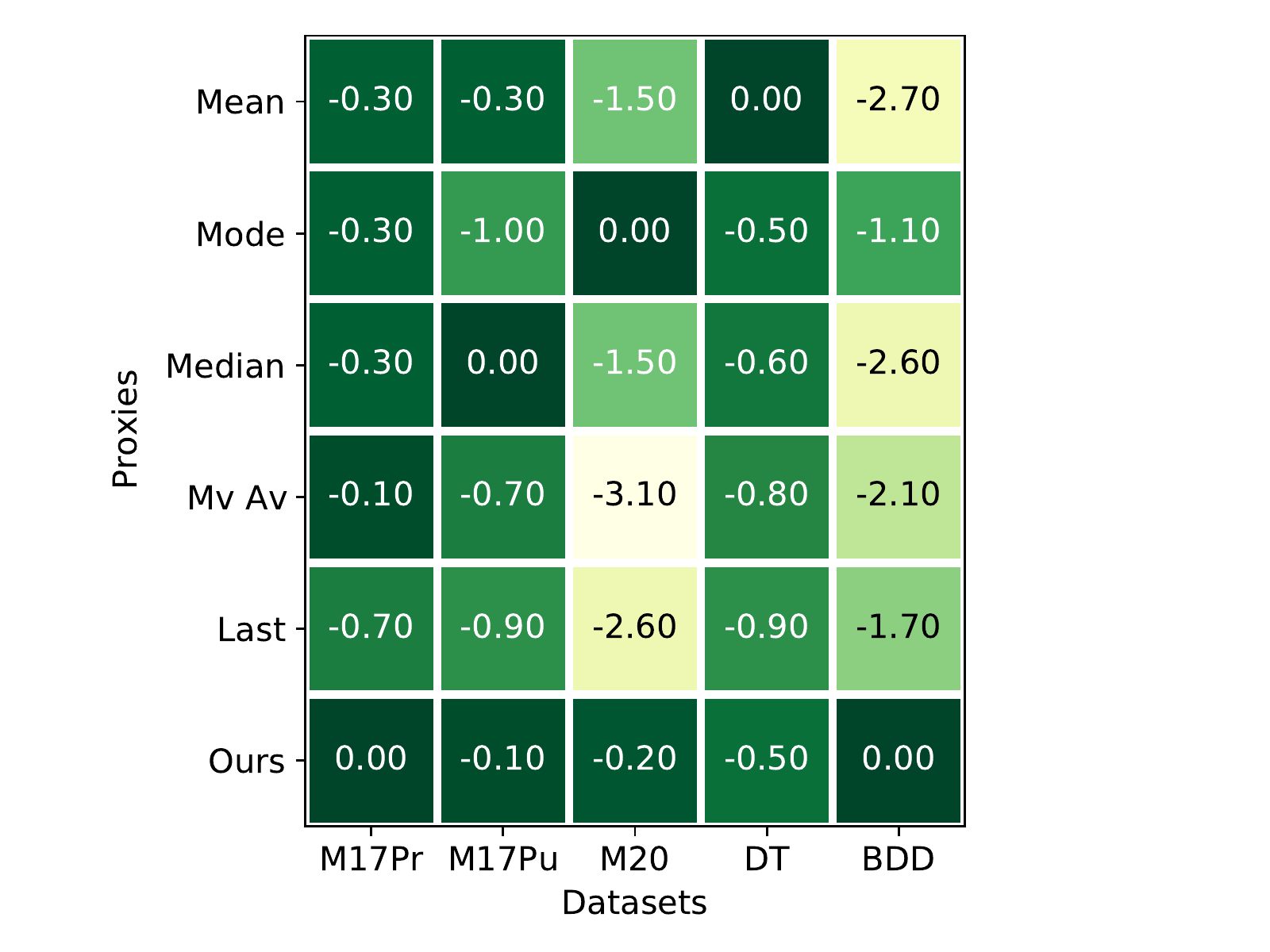}
        \vspace{-0.6cm}
        \caption{HOTA}
    \end{subfigure}
    \begin{subfigure}[b]{0.49\textwidth}   
        \centering 
        \includegraphics[width=\textwidth]{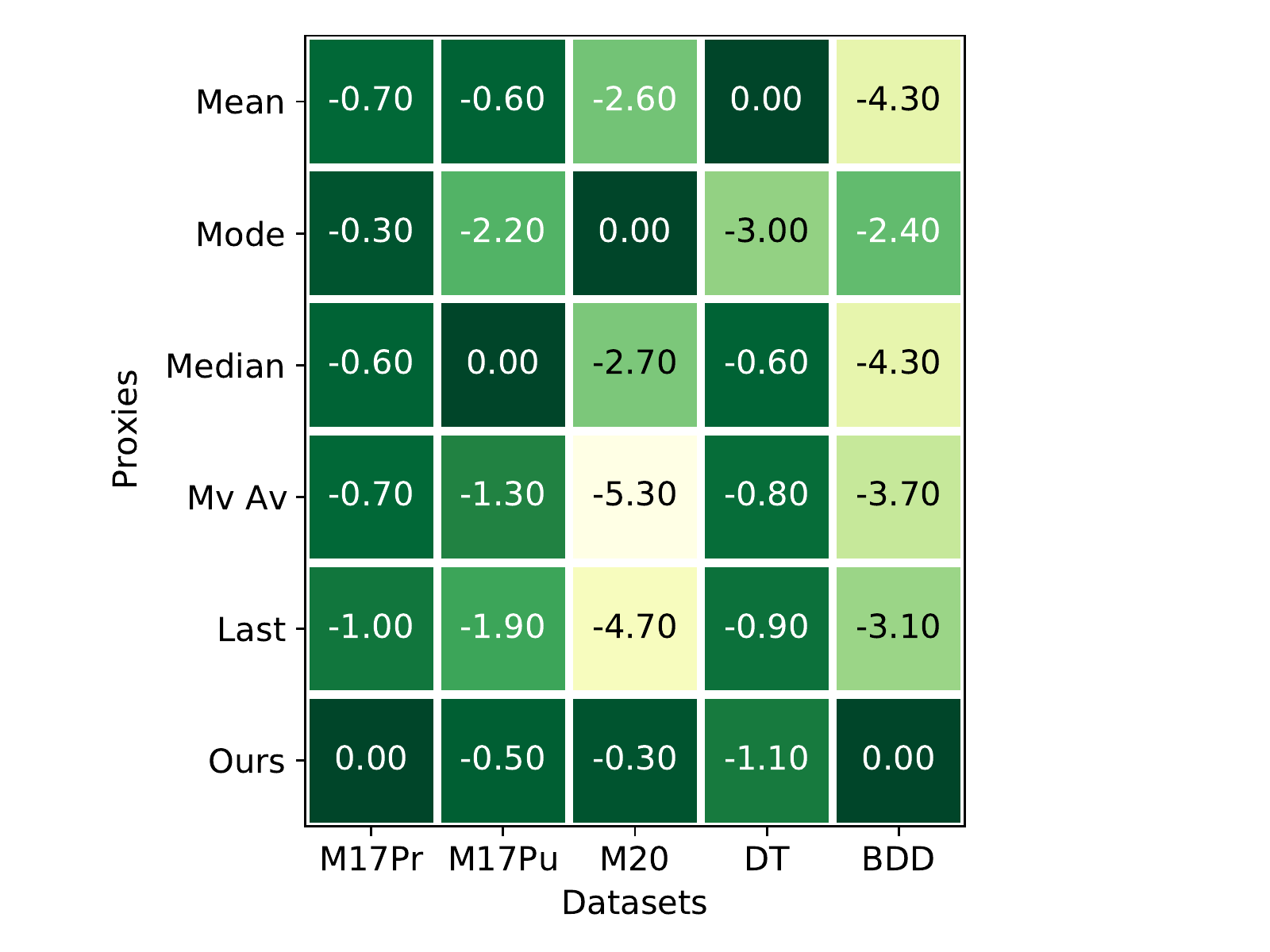}
        \vspace{-0.6cm}
        \caption{IDF1}
    \end{subfigure}
    \vspace{-0.3cm}
\caption{Drop in Performance for Different Proxies on Different Datasets. M17Pr = MOT17 private detections, M17Pu = MOT17 public detections, M20 = MOT20 public detections, DT = DanceTrack, BDD = BDD100k. Mean = Mean Feature, Mode = Mode Features, Median = Median Features, Mv Av = Moving Average of Features, Last = Last Features, Ours = Our Proxy Distance.}
    \label{fig:proxies}
\vspace{-0.5cm}
\end{figure*}

We now explore different proxies for the distance computation between new detections and inactive tracks. We start from the feature vectors generated using our reID network and normalize them before further processing. As introduced in the main paper, we utilize the mean of the distances of a new detection to all detections of an inactive track. This proxy distance between new detection $i$ and inactive track $k$ is given by:

\vspace{-0.2cm}
\begin{equation}
\begin{split}
    \Tilde{d}(i, k) & = \frac{1}{N_k} \sum_{n=1}^{N_k} d(f_i,f_k^n) \\
    & = \frac{1}{N_k} \sum_{n=1}^{N_k} 1-\frac{f_i \cdot f_k^n}{||f_i||\cdot||f_k^n||} \\
    & = 1-\frac{1}{N_k} \sum_{n=1}^{N_k} f_i \cdot f_k^n
\end{split}
\end{equation}
where $N_k$ is the number of detections in the inactive track and $f_{k}^n$ is the feature vector corresponding to its $n$-th detection. We omit $||f_i||\cdot||f_k^n||$ as we normalize all feature vectors.

Another option is to first compute a proxy feature vector and then compute the distance between a new detection and the proxy feature vector. We investigate four proxy feature vector computations and compare them on the validation set of all four datasets.

\noindent\textbf{Mean Feature Vector.} The mean feature vector of all detections in the inactive track $k$ which is also used in Tracktor \cite{DBLP:conf/iccv/BergmannML19} is given by
\begin{equation}
    \Tilde{f}_k = \frac{1}{N_k} \sum_{n=0}^{N_k} f_{k}^n
\end{equation}
Computing the cosine distance of this mean feature vector leads to 
\begin{equation}
\begin{split}
    \Tilde{d}(i, k) & = 1- \frac{f_i \cdot \frac{1}{N_k} \sum_{n=1}^{N_k} f_k^n}{||f_i||\cdot||\frac{1}{N_k} \sum_{n=1}^{N_k} f_k^n||} \\
    & = 1- \frac{\sum_{n=1}^{N_k} f_i \cdot f_k^n}{||\sum_{n=1}^{N_k} f_k^n||}
\end{split}
\end{equation}
This differs from our proxy distance by the normalization constant $\frac{1}{||\sum_{n=1}^{N_k} f_k^n||}$.

\noindent\textbf{Mode Feature Vector.} Compared to the mean feature vector, the feature vector of inactive track $k$ is given by the value that appeared most in each feature dimension.

\noindent\textbf{Median Feature Vector.} Viewing $f_k^n$ as a random variable, in each dimension the median feature vector contains the value for which $50\%$ of the probability mass of feature values in this dimension lies on the right and left of it, \ie, it divides the probability mass into two equal masses. 

\noindent\textbf{Exponential Moving Average Feature Vector.} Utilizing the exponential moving average (EMA) as feature vector as done in JDE \cite{DBLP:conf/nips/RenHGS15} or FairMOT \cite{DBLP:journals/ijcv/ZhangWWZL21} means that at given a new detection, the feature vector is updated by:

\begin{equation}
    \Tilde{f}_k^t = \Tilde{f}_k^{t-1} * \alpha + f_k^t * (1-\alpha)
\end{equation}
where $\Tilde{f}_k^{t-1}$ is the EMA feature vector at the previous time step, $f_k^t$ is the feature vector of the new detection, and $\alpha=0.9$ is a weighting factor. The EMA feature vectors build on the underlying assumption that feature vectors should change only slightly and, therefore, smooths the feature vector development.

We show the performance drop on different datasets when using different proxies in Fig~\ref{fig:proxies}. Ours, \textit{i.e.}, the mean distance shows the most stable performance over the different datasets and we, therefore, decided to utilize this proxy.

\begin{figure*}[t]
    \centering
    \begin{subfigure}[b]{0.49\textwidth}
        \centering
        \includegraphics[width=\textwidth]{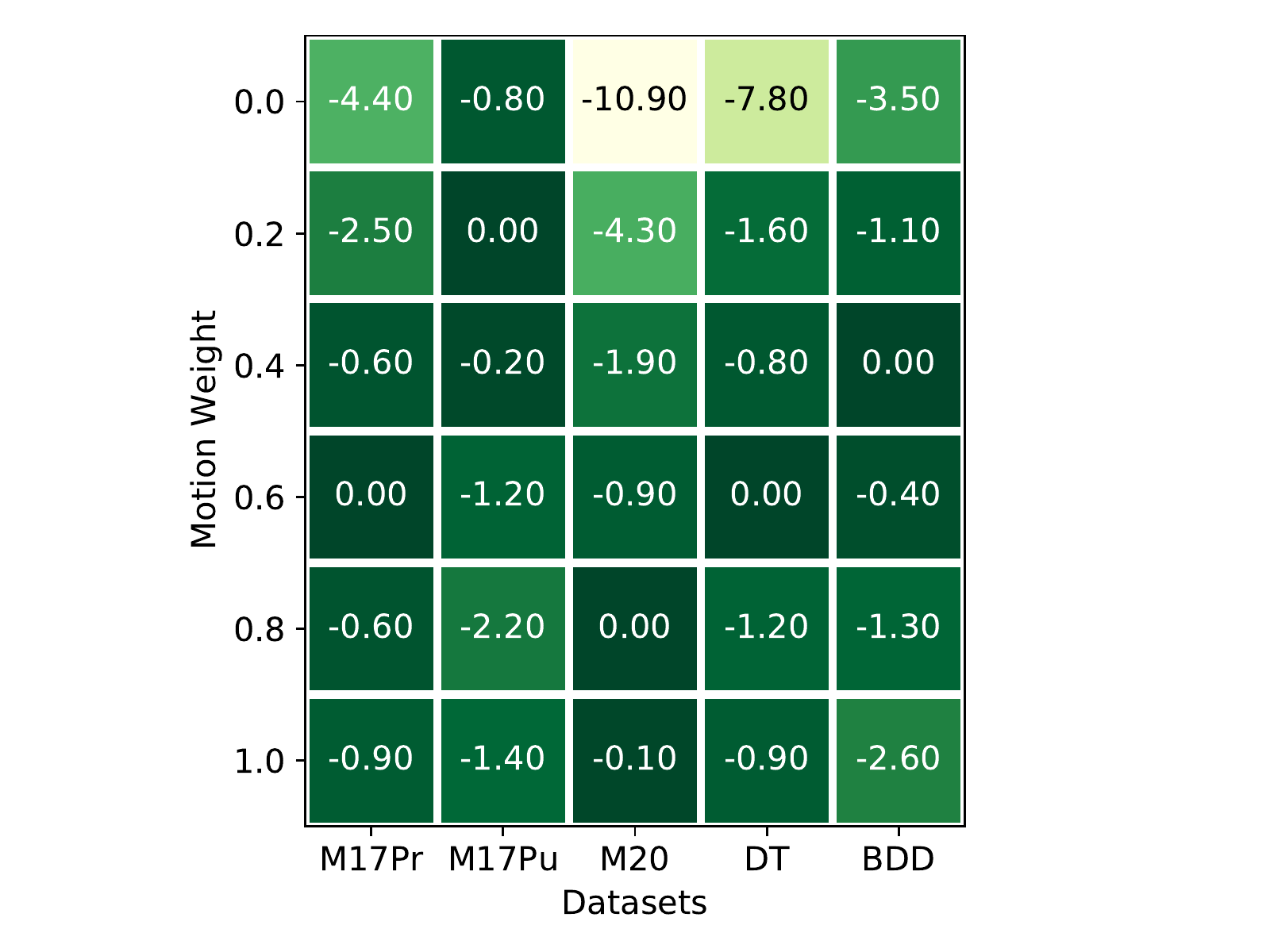}
        \vspace{-0.6cm}
        \caption{HOTA}
    \end{subfigure}
    \begin{subfigure}[b]{0.49\textwidth}   
        \centering 
        \includegraphics[width=\textwidth]{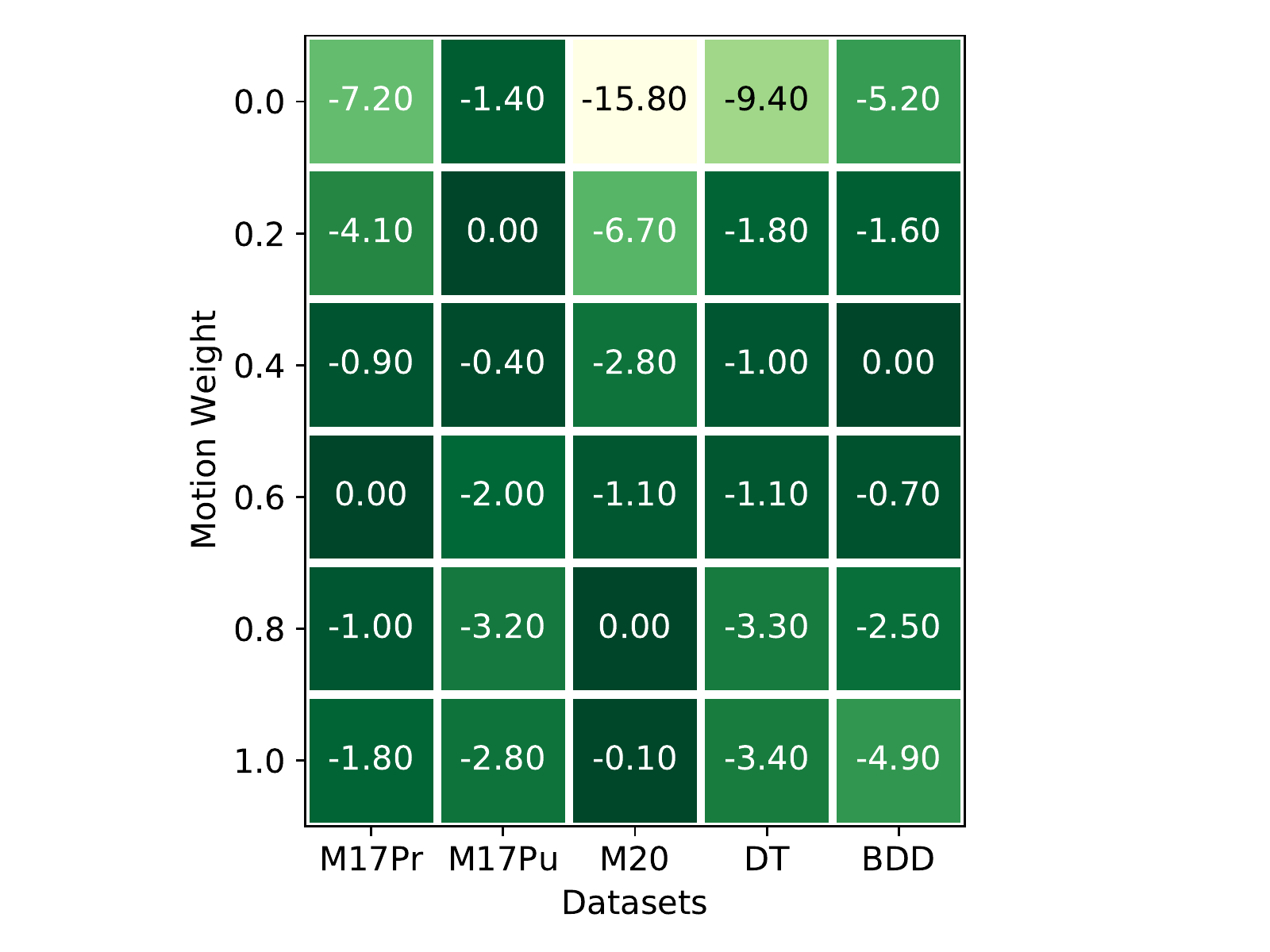}
        \vspace{-0.6cm}
        \caption{IDF1}
    \end{subfigure}
    \vspace{-0.2cm}
    \caption{Drop in Performance for Different Motion Weights on Different Datasets. M17Pr = MOT17 private detections, M17Pu = MOT17 public detections, M20 = MOT20 public detections, DT = DanceTrack, BDD = BDD100k.}
    \vspace{-0.3cm}
    \label{fig:weights}
\end{figure*}

\subsection{The Impact of Motion Weights}
\label{sec:weights}

In this subsection, we visualize the performance drop when utilizing different motion weights on different datasets (see Fig~\ref{fig:weights}). On MOT17 public detections, the best performance is achieved when using motion weight 0.4 while for private detections the best weight is 0.6. This is caused by the fact, that the appearance model gets less certain with increasing occlusion level and the private detections set contains more difficult, \textit{i.e.}, occluded detections. 
On MOT20 private detections a motion weight of 0.8 for private detections performs best as the occlusion level is generally much higher than on MOT17 dataset. 
On DanceTrack dataset, the best motion weight is 0.4. Since the motion and articulation on this dataset are generally more diverse and extreme, the performance of the motion model is less certain compared to the appearance model. 
BDD dataset solely contains sequences recorded using a moving camera. As we showed in the main paper, the performance of the motion distance decreases when moving cameras are used. This is due to the fact, that the observed motion gets less linear since the motion of the camera and the object add up. Consequently, a motion weight of 0.4 works best on BDD100k MOT dataset.
All those observations are in line with our analysis in the main paper as well as the more detailed analysis in this supplementary in Section~\ref{sec:detailed_rca}.

\subsection{The Impact of Different Numbers of Frames for Velocity Computation}
\label{sec:num_frames}

\begin{figure*}[t]
    \centering
    \begin{subfigure}[b]{0.49\textwidth}
        \centering
        \includegraphics[width=\textwidth]{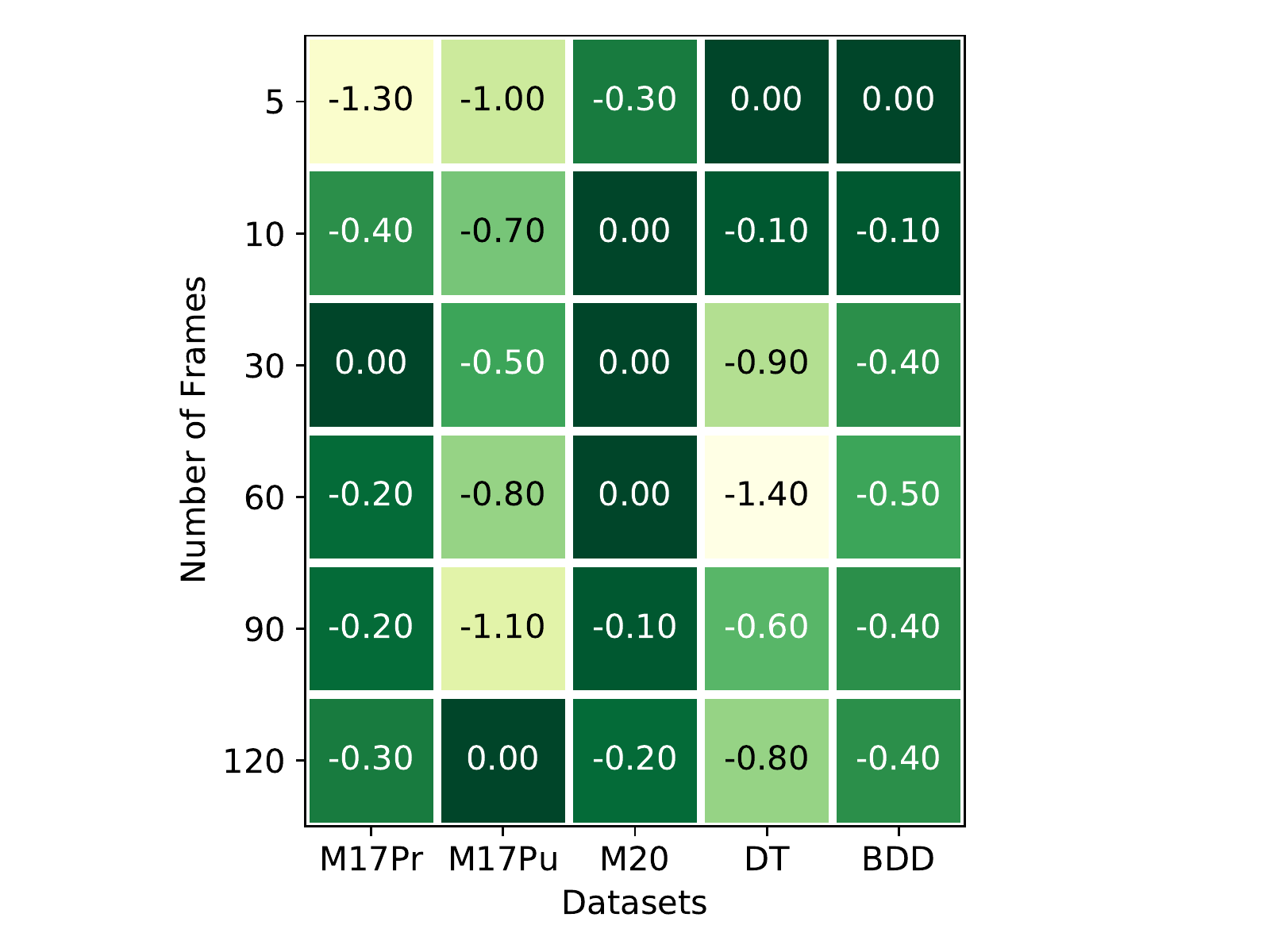}
        \vspace{-0.6cm}
        \caption{HOTA}
    \end{subfigure}
    \begin{subfigure}[b]{0.49\textwidth}   
        \centering 
        \includegraphics[width=\textwidth]{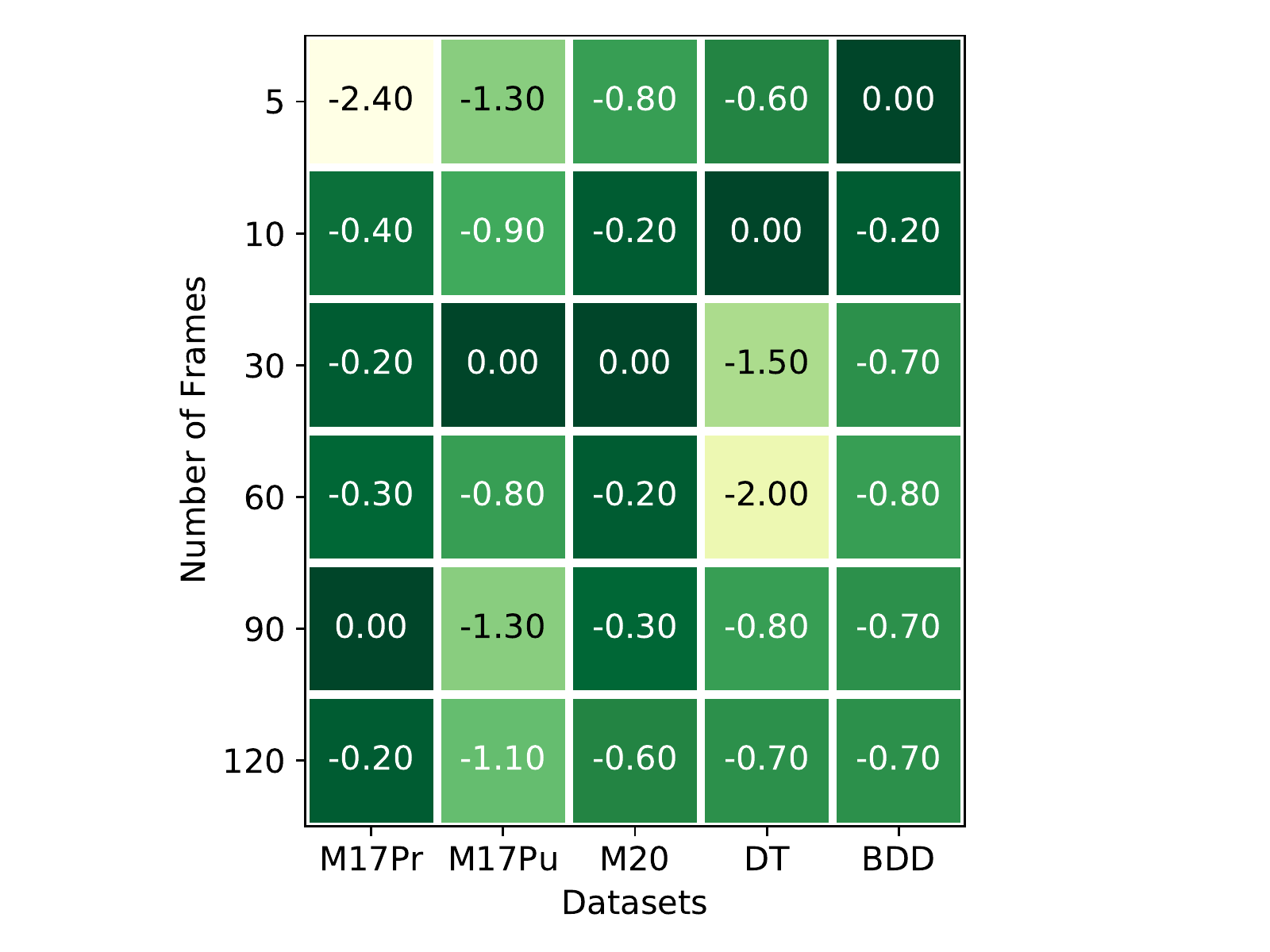}
        \vspace{-0.6cm}
        \caption{IDF1}
    \end{subfigure}
    \vspace{-0.2cm}
    \caption{Drop in Performance for Different Number of Frames for Velocity Computation on Different Datasets. M17Pr = MOT17 private detections, M17Pu = MOT17 public detections, M20 = MOT20 public detections, DT = DanceTrack, BDD = BDD100k.}
    \label{fig:num_frames}
    \vspace{-0.1cm}
\end{figure*}

The less linear the motion or the observed motion, the fewer frames approximate the future motion better. We visualize the impact of different numbers of frames in Fig~\ref{fig:num_frames}.
While on MOT17 private detections, the linear motion model performs well using the positions of the last 90 tracks (or less if a track contains less), on MOT20 using only the last 30 frames performs best since the scenes are highly crowded and, therefore, the motion is less linear.
On DanceTrack, the motion is more extreme and, therefore, using only the last 5 frames approximates the future motion best. 
Similarly, on BDD100k as the observed motion is more non-linear due to the combination of the camera motion and the object motion utilizing only the last 10 frames to approximate the motion performs best. The lower frame rate of BDD sequences compared to the frame rate of MOT17, MOT20 and DanceTrack even increases this effect, since more time passes within the same number of frames on BDD.
Overall, as already stated in the main paper, short-term future motion can be approximated fairly well utilizing a linear motion model. Depending on the characteristics of the motion, a different number of frames approximates the future motion best and, therefore, leads to the best tracking results.

\subsection{How to use Different Thresholds $\tau_{i}$}
\label{sec:threshs}
As stated in Subsection 3.2. in the main paper, we utilize different thresholds for active and inactive tracks. While commonly only one threshold is used, we empirically find that it is beneficial to allow different ones. Therefore, we apply the thresholds \textit{after} the bipartite matching to filter the detection-trajectory pairs ($i, j$). We visualize our matching in Algorithm~\ref{alg:matching}. $n$ represents the number of active track, $\tau_{act}$ and $\tau_{inact}$ the threshold for active and inactive tracks and $d$ the cost matrix.

\section{Different Inactive Patience Values}
\label{sec:memory}
\begin{figure*}[htp!]
    \centering
    \begin{subfigure}[b]{0.29\textwidth}
        \centering
        \includegraphics[height=3.5cm]{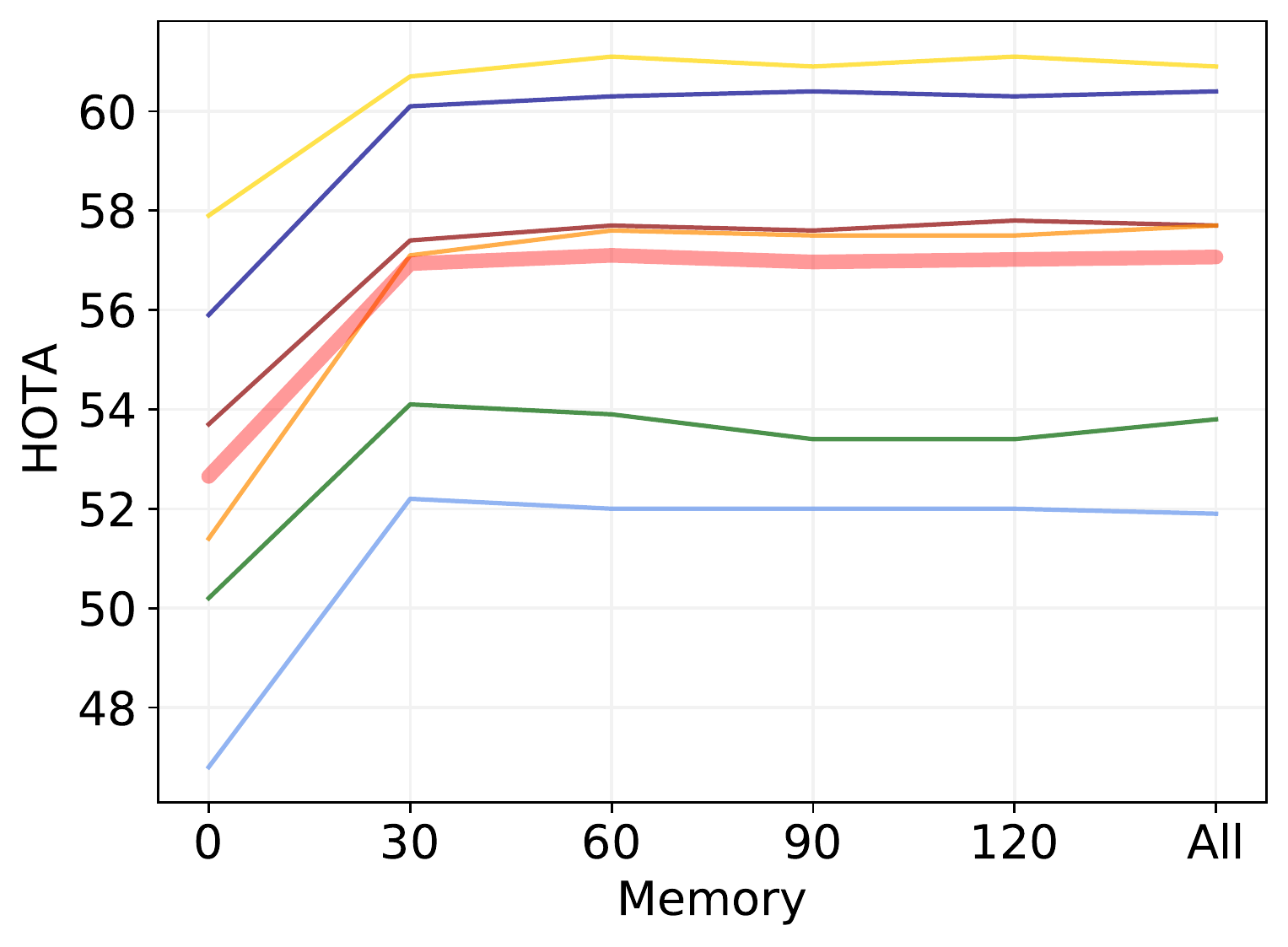}
        \caption{HOTA}
    \end{subfigure}
    \begin{subfigure}[b]{0.29\textwidth}   
        \centering 
        \includegraphics[height=3.5cm]{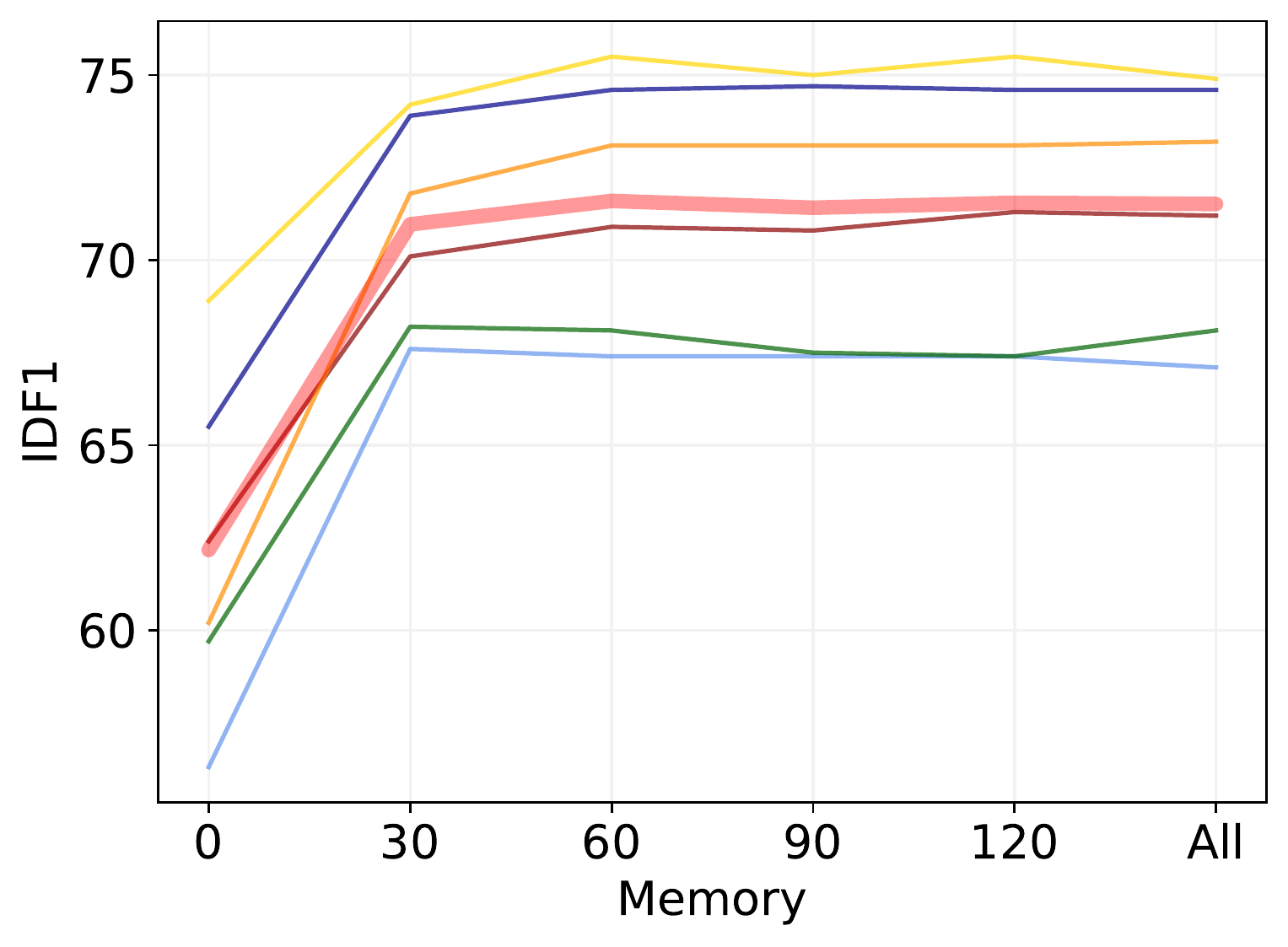}
        \caption{IDF1}
    \end{subfigure}
    \begin{subfigure}[b]{0.39\textwidth}   
        \centering 
        \includegraphics[height=3.5cm]{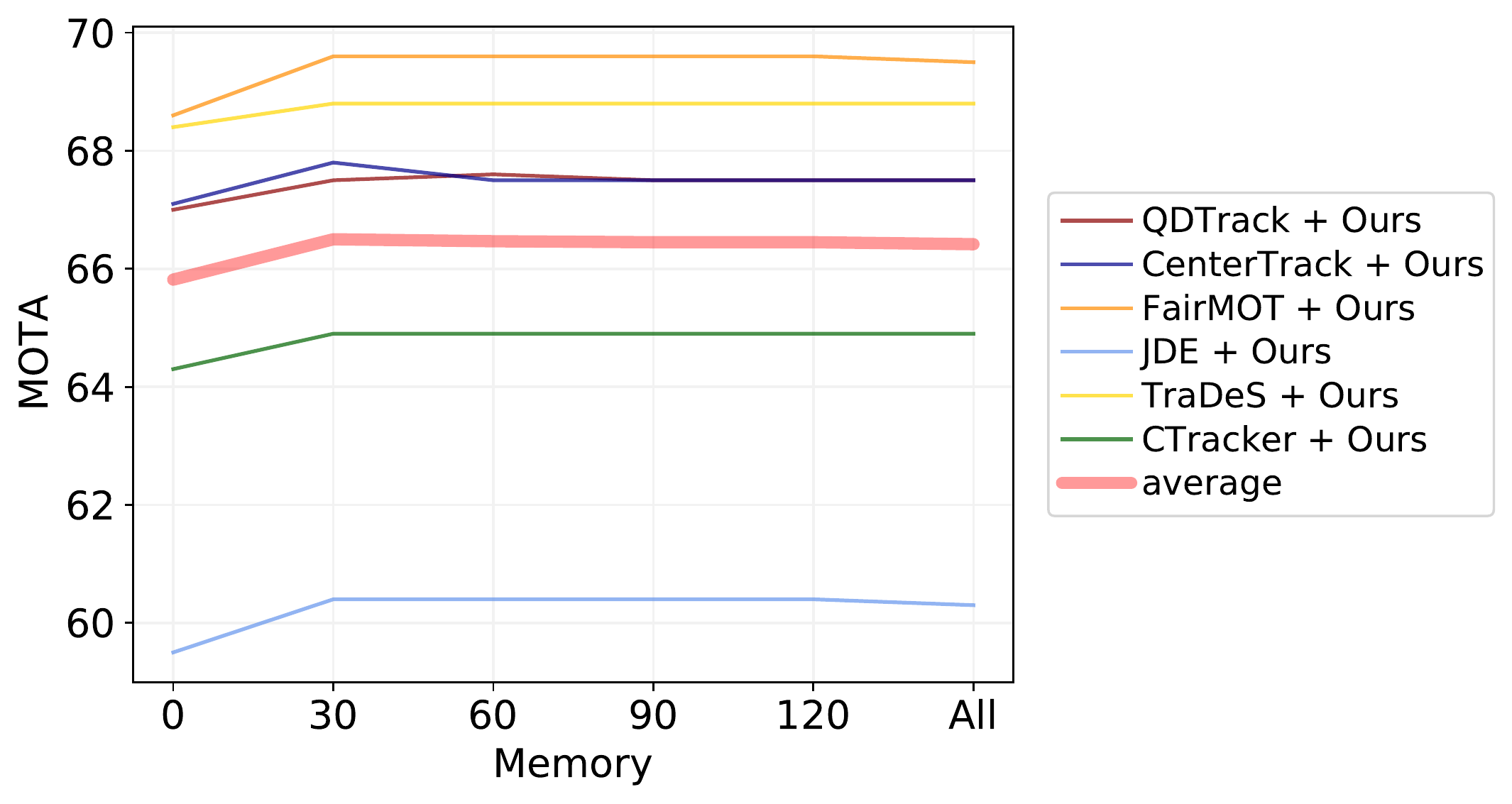}
        \caption{MOTA}
    \end{subfigure}
     \vspace{-0.1cm}
    \caption{Performance on MOT17 public validateion set with respect to different inactive patience values.}
    \label{fig:mem}
   \vspace{-0.5cm}
\end{figure*}

Similar to other approaches~\cite{DBLP:conf/iccv/BergmannML19,DBLP:conf/cvpr/StadlerB21,DBLP:journals/ijcv/ZhangWWZL21,DBLP:conf/nips/RenHGS15,DBLP:conf/ijcai/LiuC0Y20} that only keep inactive tracks for a fixed number of frames, called inactive patience, we keep them for 50 frames for all datasets. To show that this choice is reasonable, we visualize HOTA, IDF1, and MOTA on MOT17 validation set for different inactive patience values in Fig~\ref{fig:mem}. We use the same setting as in sections 4.4 and 4.5 in the main paper, \textit{i.e.}, we use the bounding boxes of MOT17 validation set of several private trackers. The performance drops heavily for inactive patience 0 and then only slightly changes up to using all frames of a sequence after 30 frames.
\vspace{-0.3cm}

\RestyleAlgo{ruled}
\begin{algorithm}
\caption{Assignment with different thresholds}\label{alg:matching}
\KwData{$n, \tau_{act}, \tau_{inact}, \text{cost matrix } d\in\mathcal{R}^{|T|\times|D|}$\;}
\KwResult{approved rows, approved cols\;}
$\text{approved rows} = \emptyset$, $\text{approved cols} = \emptyset$\;
$\text{matched rows}, \text{matched cols} = \text{Bipartite}(c)$\;
\For{$r, c \text{ in } \text{zip}(\text{matched rows}, \text{matched cols})$}{
     \uIf{$r < n \text{ and } d_{r, c} < \tau_{act}$}{
        $\text{approved rows} = \text{approved rows}  + r$\;
        $\text{approved cols} = \text{approved cols}  + c$\;}
     \uElseIf{$r >= n \text{ and } d_{r, c} < \tau_{inact}$}{
        $\text{approved rows} = \text{approved rows}  + r$\;
        $\text{approved cols} = \text{approved cols}  + c$\;}
     \Else{
        \text{Discard Match.}\;}
     }
\end{algorithm}

\vspace{-0.7cm}
\section{Computation of Distance Histograms}
\label{sec:distance_dist}
In the main paper, we visualize distributions of distances from active and inactive tracks to detections of the same and different classes in Fig 2. In Fig 2(a) we utilize the embeddings of the last detection of any existing track to compute the distance to the embeddings of new detections, in Fig 2(b) we utilize the distance computation as introduced in Sec 3.2, and in 2(c) we visualize the motion distance. Since different distance metrics could be applied for feature vector distance and motion distance, we define both in Sec 4.1.
To generate those distributions, we first match given detections to ground truth identities following the same matching as used for the computation of the MOTA metric \cite{DBLP:journals/pami/KasturiGSMGBBKZ09}. 
If a detection of the same ID occurred in the last frame, we compute the distance to it and add it to the distances between active tracks and detections of the same ID. 
Similarly, if there was a detection present but not in the last frame, we compute the distance and add it to the distances from inactive tracks to detections of the same ID. 
Afterwards, we compute the distances to all other IDs that occurred in the last frame as well as to all other IDs that occurred prior to the last frame and add them to the distances from detections to active and inactive tracks of a different ID, respectively.
We can add inactive patience and proxy computation methods to this basic framework. Despite we only show the distributions for MOT17 validation set, this method can be used for any dataset for which ground truth detections are available.
\vspace{-0.2cm}
\section{On Similar Approaches}
\label{sec:similar}
In this section, we discuss the differences between state-of-the-art trackers that share some of their components with GHOST.
ByteTrack \cite{DBLP:journals/corr/abs-2110-06864} uses a Kalman Filter as motion model while we use a more simple linear motion model. More importantly, the authors treat active and inactive \textbf{tracks} the same but distinguish between high and low confidence \textbf{detections}, \textit{i.e.}, they differentiate on a detection level while we differentiate on track level. However, as we showed in the main paper, active and inactive tracks show significant differences and treating them the same way does not leverage the full potential of the underlying cues. Also, their assignment strategy leads to a multi-level association process while GHOST only requires a single association step.
Similarly, in \cite{DBLP:conf/cvpr/BaeY14} the authors treat \textbf{high and low confidence tracks} differently, \textit{i.e.}, high confidence tracks are assigned locally to new detections and low confidence tracks are globally assigned with other tracks and detections. The inactivity of a track is only one factor of confidence. Note that this again involves multiple bipartite matchings while we assign active and inactive tracks \textbf{at the same time} which only requires one.
The authors of DeepSORT \cite{DBLP:conf/icip/WojkeBP17} utilize a Kalman filter as well as an appearance model. However, the parameter that weights appearance and motion is set to $\lambda=0$, \textit{i.e.}, only appearance is considered. However, as we show in our analysis motion can compensate for failure cases of appearance, especially in low visibility regimes making our approach more robust.
Moreover, the authors propose a cascaded matching strategy that requires more than one bipartite matching per frame while we, again, only require one.
With respect to domain adaptation, HCC \cite{DBLP:conf/accv/MaTBG18} trains on tracking sequences and uses sophisticated test-time mining to fine-tune. We rely on a simpler scheme and do not need any of the above.

Despite all of the above-mentioned approaches showing similarities to our approach, they still differ with respect to significant design choices positioning our approach as a complementary work with respect to them. Furthermore, our approach leverages the motion and appearance cues in a simple yet highly effective and general way without multi-level association procedures. 

\section{On the Generality of ByteTrack \cite{DBLP:journals/corr/abs-2110-06864}}
\label{sec:generality}
Recently, ByteTrack \cite{DBLP:journals/corr/abs-2110-06864}, which also follows the tracking-by-detection paradigm, also reported results on MOT17 and MOT20 as well as on the highly different datasets DanceTrack and BDD100k MOT. In this section, we compare GHOST to ByteTrack with respect to generality. Despite not being mentioned in the paper, the authors add tricks to the tracking procedure which are different for each dataset.
%
%
\textbf{First}, they multiply their IoU cost matrix, which they obtain by using a Kalman Filter, by the detection confidence when applying their tracker to MOT17, but not when applying it to other datasets. 
\textbf{Second}, the authors apply interpolation on MOT17 and MOT20 datasets which turns their approach into an offline approach. 
\textbf{Third}, on DanceTrack and BDD100k, they allow all bounding boxes to be used, while they filter out bounding boxes if $\frac{w}{h} > 1.6$ on MOT17 and MOT20, where $w$ and $h$ are bounding box width and height, respectively. 
\textbf{Fourth}, ByteTrack uses a reID model, namely UniTrack \cite{DBLP:conf/nips/WangZLWTB21}, on BDD100k dataset whilst they do not use any reID model on the other datasets. 
\textbf{Fifth}, they adapt the tracking thresholds \textit{per sequence} on MOT17 and MOT20 during training \textit{and testing}. On BDD and DanceTrack the tracking thresholds are applied per dataset.
\textbf{Sixth}, as commonly done the authors adapt other model parameters, \textit{e.g.}, the matching thresholds, confidence threshold for detections as well as the confidence threshold for new tracks.

We believe these are small but significant changes that put into question the generality of ByteTrack. 
In contrast, we keep our tracking pipeline the same over different datasets but solely change our model parameters for each dataset as a \textbf{whole}. To be specific, we adapt the thresholds $\tau_i$, the detection confidence thresholds to filter plain detections and start new tracks, the motion weights, as well as the number of frames used in the linear motion model. 
This makes our approach more general and easier to apply to new datasets. 

\section{Latency}
\label{sec:latency}
For a fair comparison with other methods, we evaluated GHOST, Tracktor \cite{DBLP:conf/iccv/BergmannML19}, and FairMOT \cite{DBLP:journals/ijcv/ZhangWWZL21} on the public detections on the MOT17 validation set utilizing the same GPU, namely a Quadra P6000. Note that we utilize CenterTrack pre-processed detections here.
With 10FPS GHOST is on the same magnitude of speed as current SOTA trackers. While Tracktor \cite{DBLP:conf/iccv/BergmannML19} runs at 2FPS, FairMOT \cite{DBLP:journals/ijcv/ZhangWWZL21} runs at 17FPS as it was optimized for real-time. 
When evaluating the private detection setting, our method's latency increases slightly due to the increase of bounding boxes to process to 6FPS.
The average latency per frame and per model part is given by 10ms for the computation of the reID features, 30ms for the reID distance computation, 0.1 ms for updating the velocity per track, 0.0025ms for the motion step per track, 0.4ms for the motion distance computation, 0.35ms for the biparite matching, and 0.1ms for updating all tracks.

\section{Visualizations}
\label{sec:vis}
In this section, we visualize associations on CenterTrack re-fined public bounding boxes that our model is able to correctly associate while CenterTrack is not. Correct and wrong associations are determined in the same way as done for the computation of the RCA Section~\ref{sec:detailed_rca}. This means we determine wrong associations by first matching all detection bounding boxes to the ground truth IDs. A wrong association is given if the prior detection of the same ground truth ID as a current detection was assigned to a different tracker ID than the current detection. In Figures \textcolor{red}{10}-\textcolor{red}{17}, we visualize the prior detection on the left side and the current detection on the right side. All examples were associated wrongly by CenterTrack and correctly by GHOST.
We give the time distance between the prior and the current frame in the caption as well as the visibility level of the re-appearing pedestrian. 
By our combination of appearance and motion, we are able to correctly associate pedestrians after long occlusions and low visibility in highly varying sequences.

\begin{figure*}[t]
    \centering
    \begin{subfigure}[b]{0.35\textwidth}   
        \centering 
        \includegraphics[width=\textwidth]{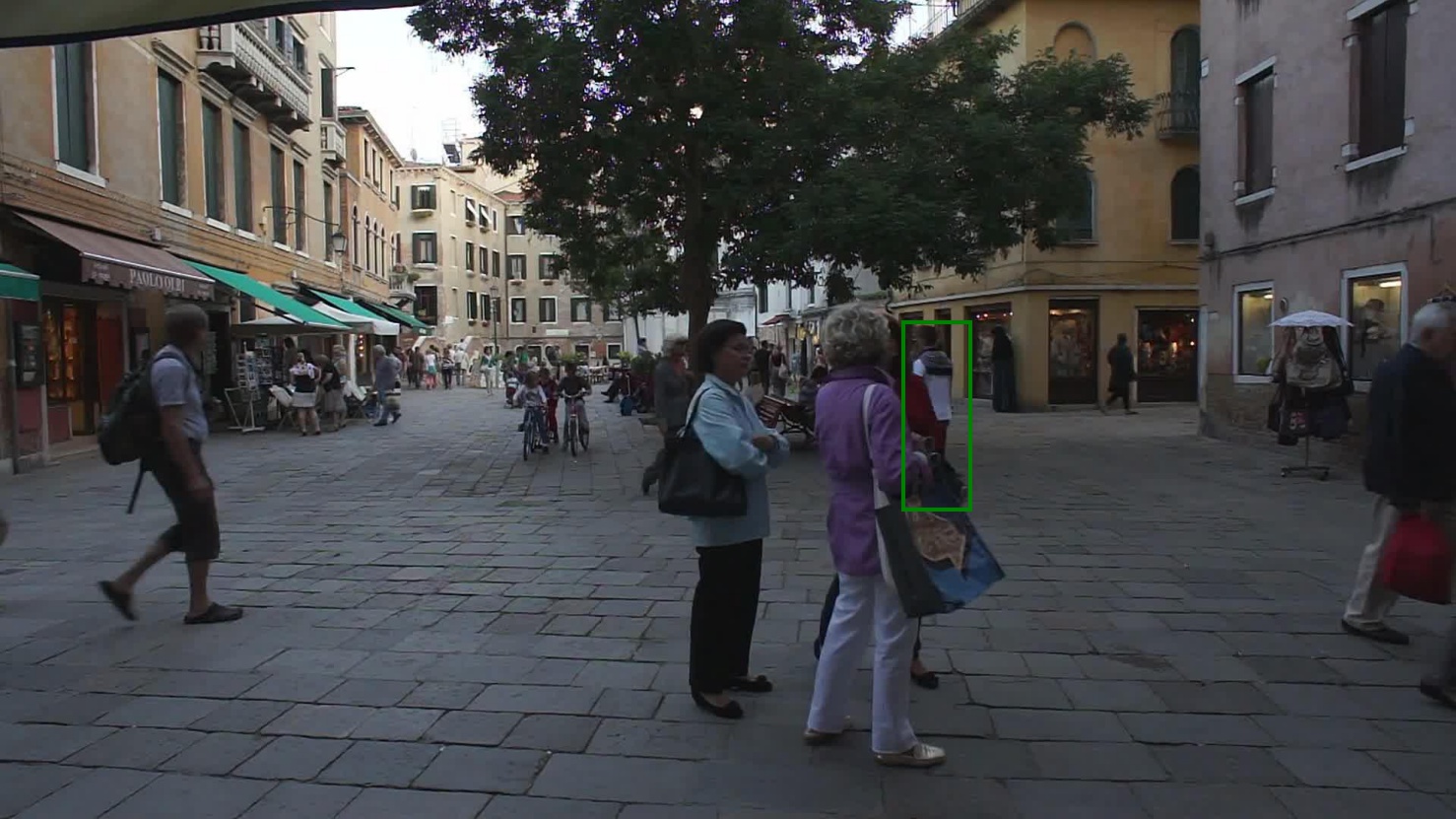}
        
    \end{subfigure}
    \begin{subfigure}[b]{0.35\textwidth}   
        \centering 
        \includegraphics[width=\textwidth]{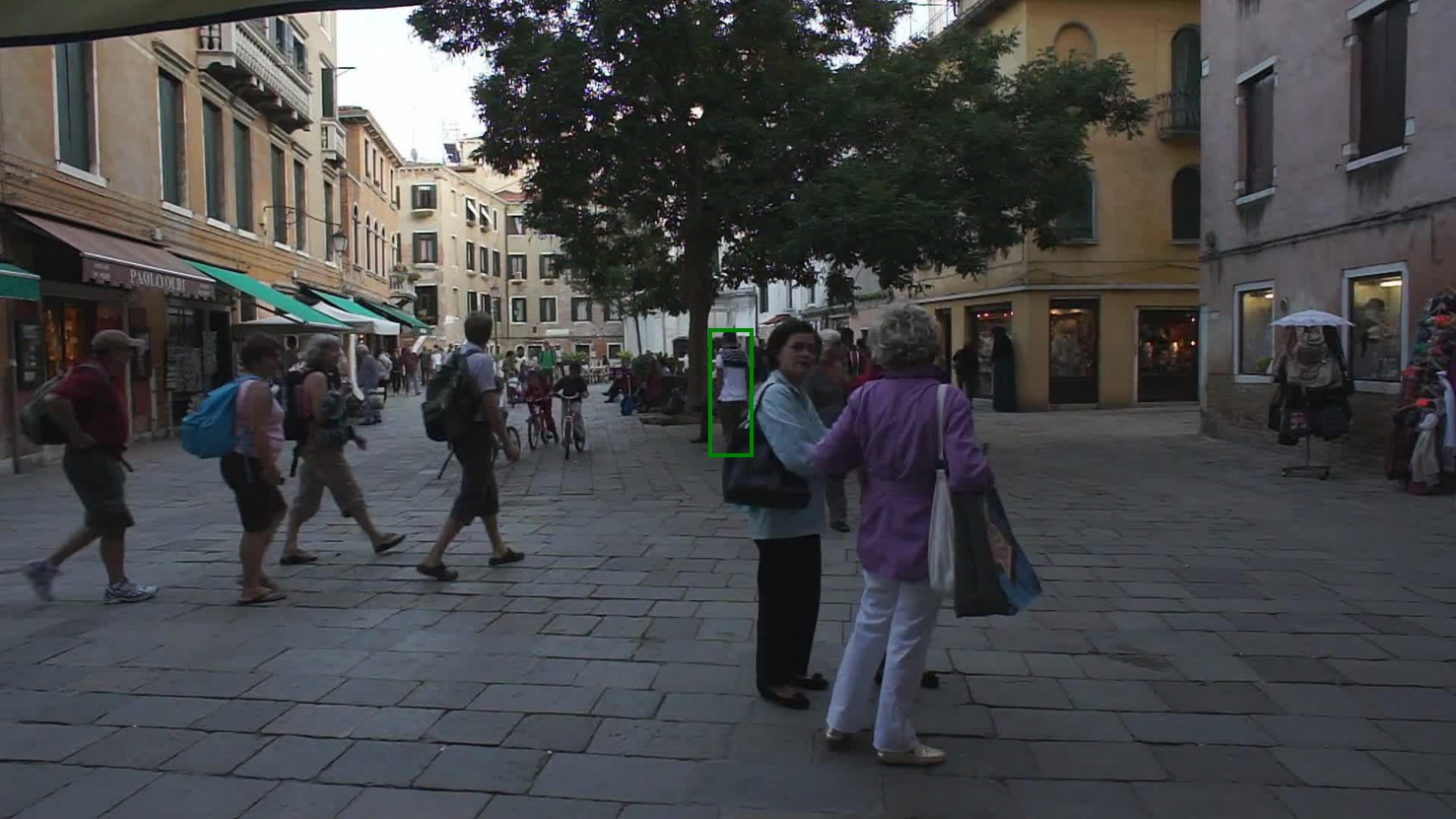}
    
    \end{subfigure}
    \caption{Occlusion time: $2s$, visibility of re-appearing pedestrian: $0.6$.}
    \label{fig:mot17_2_1}
\end{figure*}

\begin{figure*}
    \centering
    \begin{subfigure}[b]{0.35\textwidth}   
        \centering 
        \includegraphics[width=\textwidth]{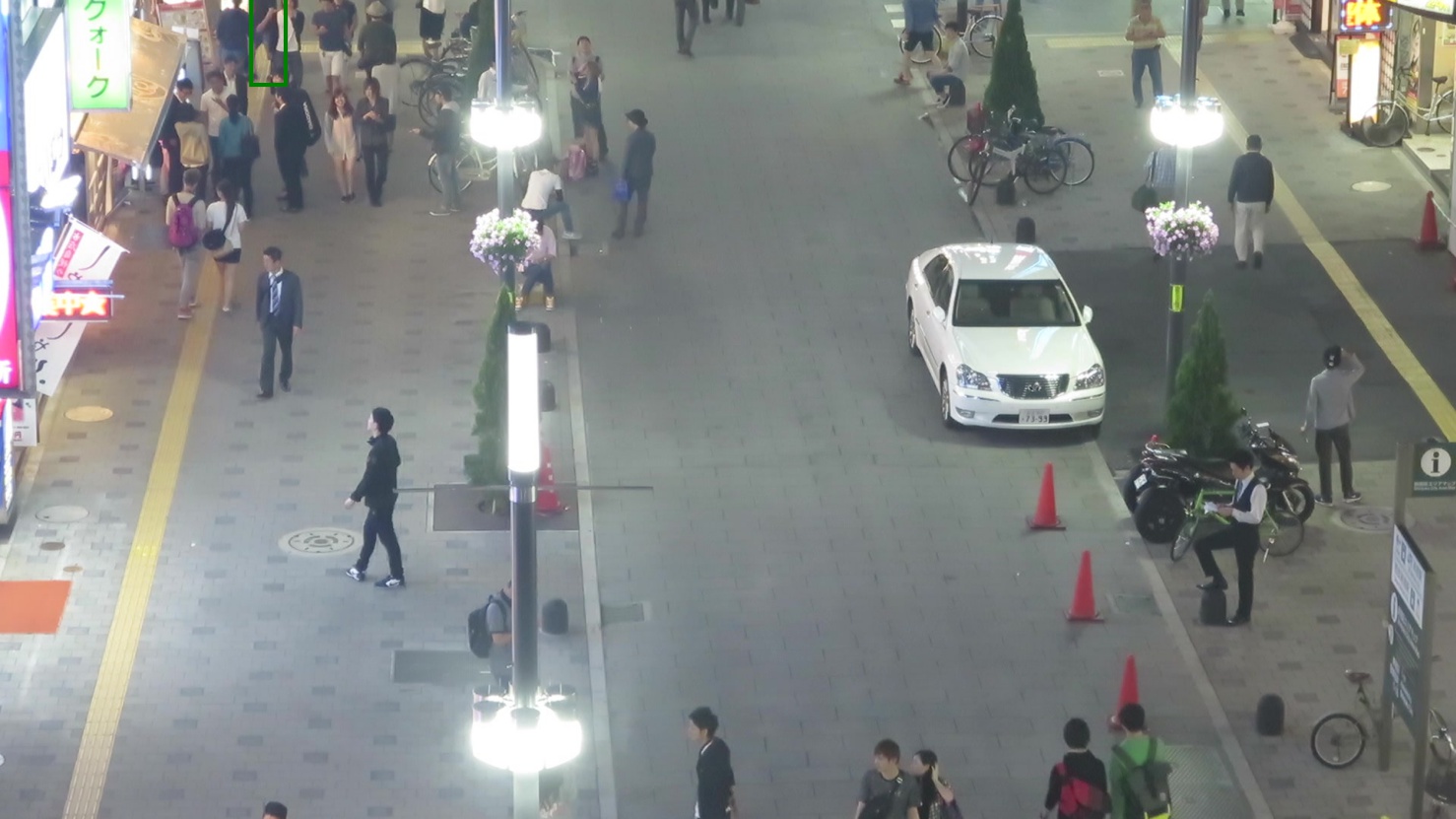}

    \end{subfigure}
    \begin{subfigure}[b]{0.35\textwidth}   
        \centering 
        \includegraphics[width=\textwidth]{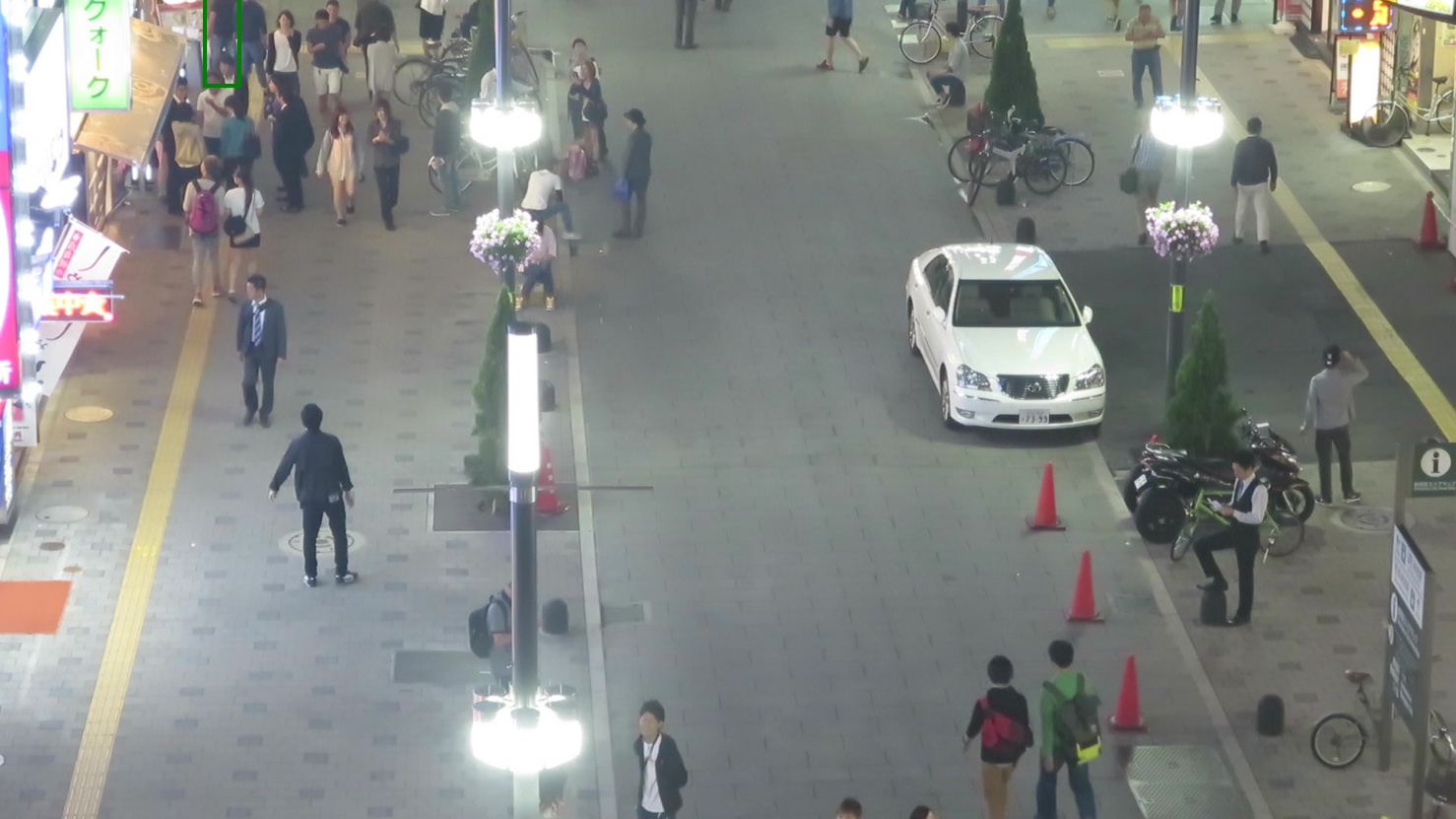}

    \end{subfigure}
    \caption{Occlusion time: 1.1s, visibility of re-appearing pedestrian: 0.3.}
    \label{fig:mot17_4_1}
\end{figure*}
\begin{figure*}[t]
    \centering
    \begin{subfigure}[b]{0.35\textwidth}   
        \centering 
        \includegraphics[width=\textwidth]{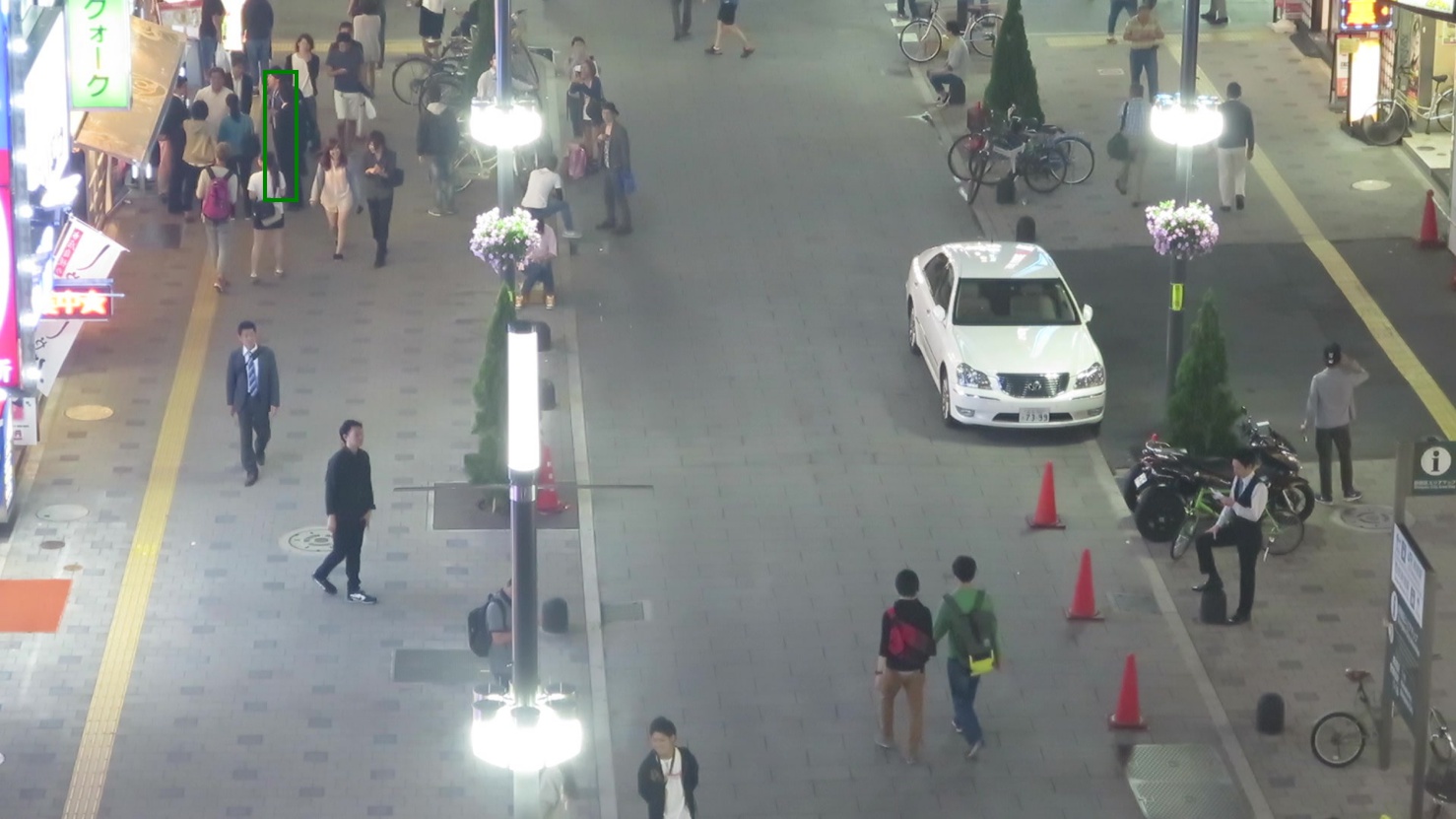}

    \end{subfigure}
    \begin{subfigure}[b]{0.35\textwidth}   
        \centering 
        \includegraphics[width=\textwidth]{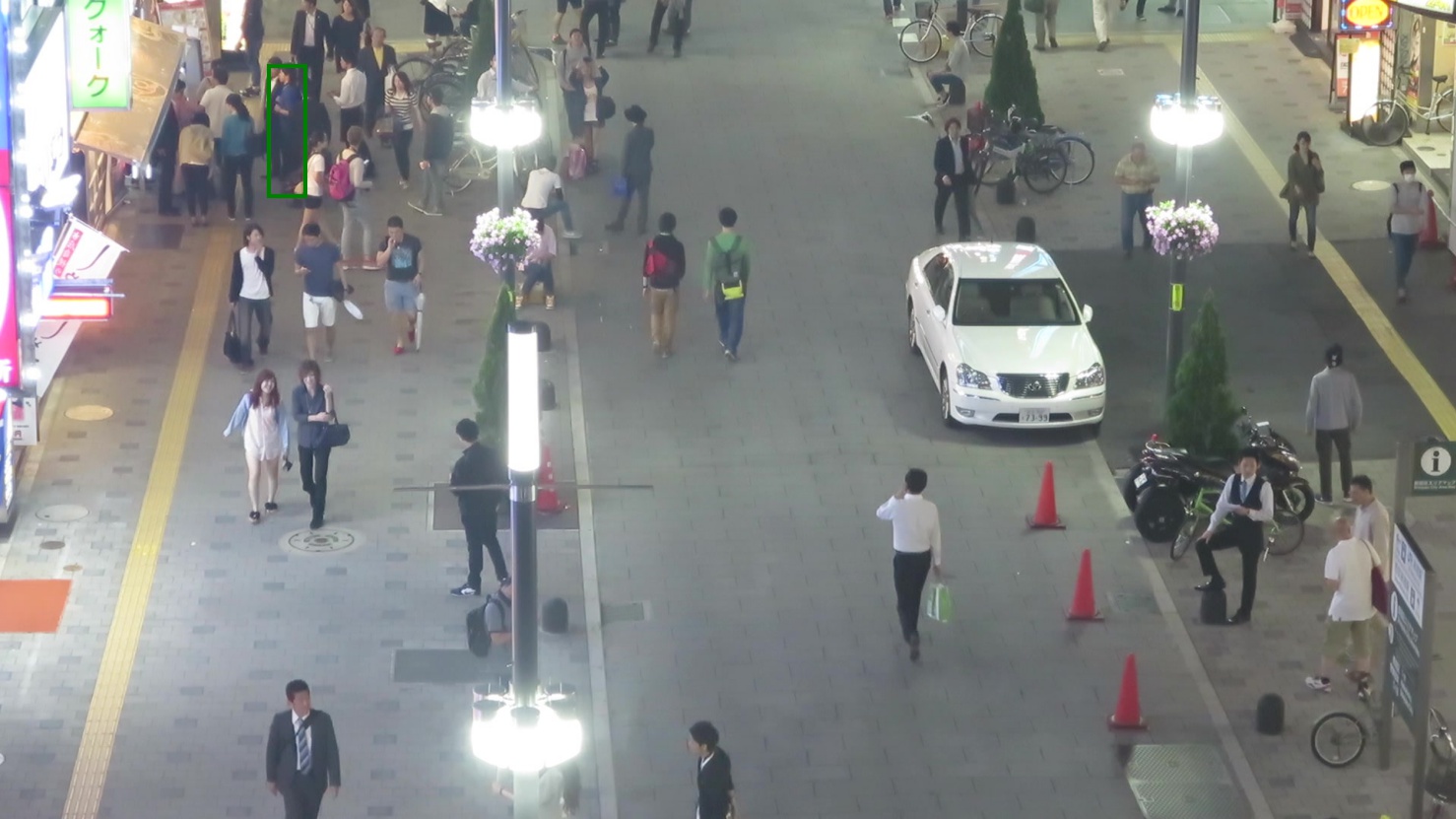}

    \end{subfigure}
    \caption{Occlusion time: 9.4s, visibility of re-appearing pedestrian: 0.4.}
    \label{fig:mot17_4_2}
\end{figure*}

\begin{figure*}[t]
    \centering
    \begin{subfigure}[b]{0.35\textwidth}   
        \centering 
        \includegraphics[width=\textwidth]{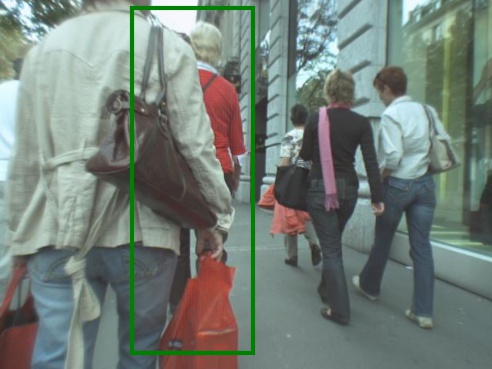}

    \end{subfigure}
    \begin{subfigure}[b]{0.35\textwidth}   
        \centering 
        \includegraphics[width=\textwidth]{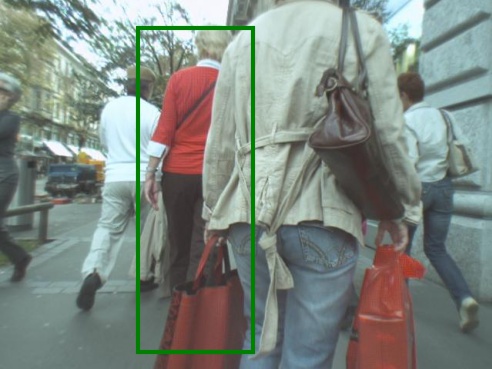}

    \end{subfigure}
    \caption{Occlusion time: 1.1s, visibility of re-appearing pedestrian: 0.5.}
    \label{fig:mot17_5_1}
\end{figure*}

\newpage

\begin{figure*}[t]
    \centering
    \begin{subfigure}[b]{0.35\textwidth}   
        \centering 
        \includegraphics[width=\textwidth]{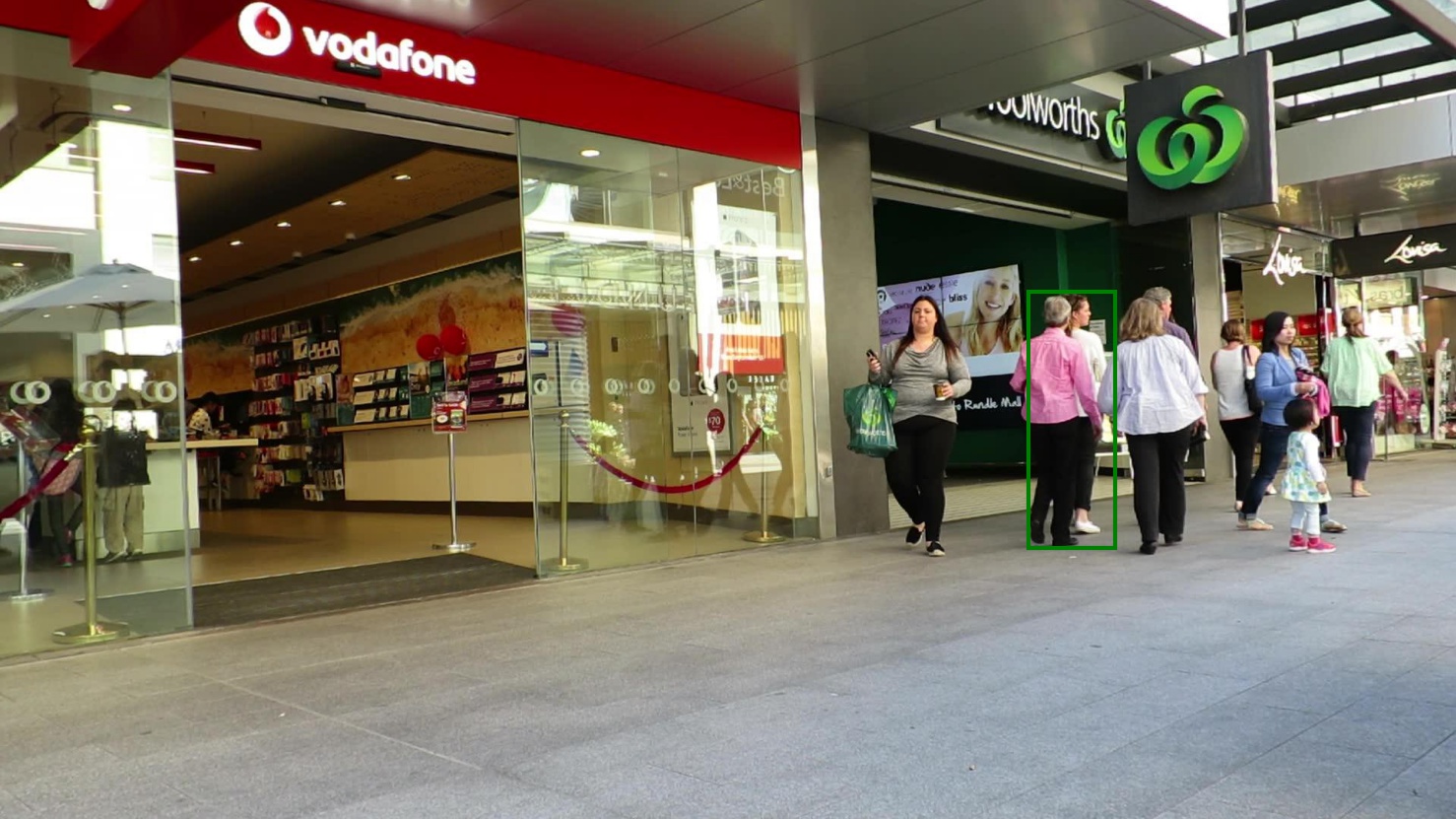}

    \end{subfigure}
    \begin{subfigure}[b]{0.35\textwidth}   
        \centering 
        \includegraphics[width=\textwidth]{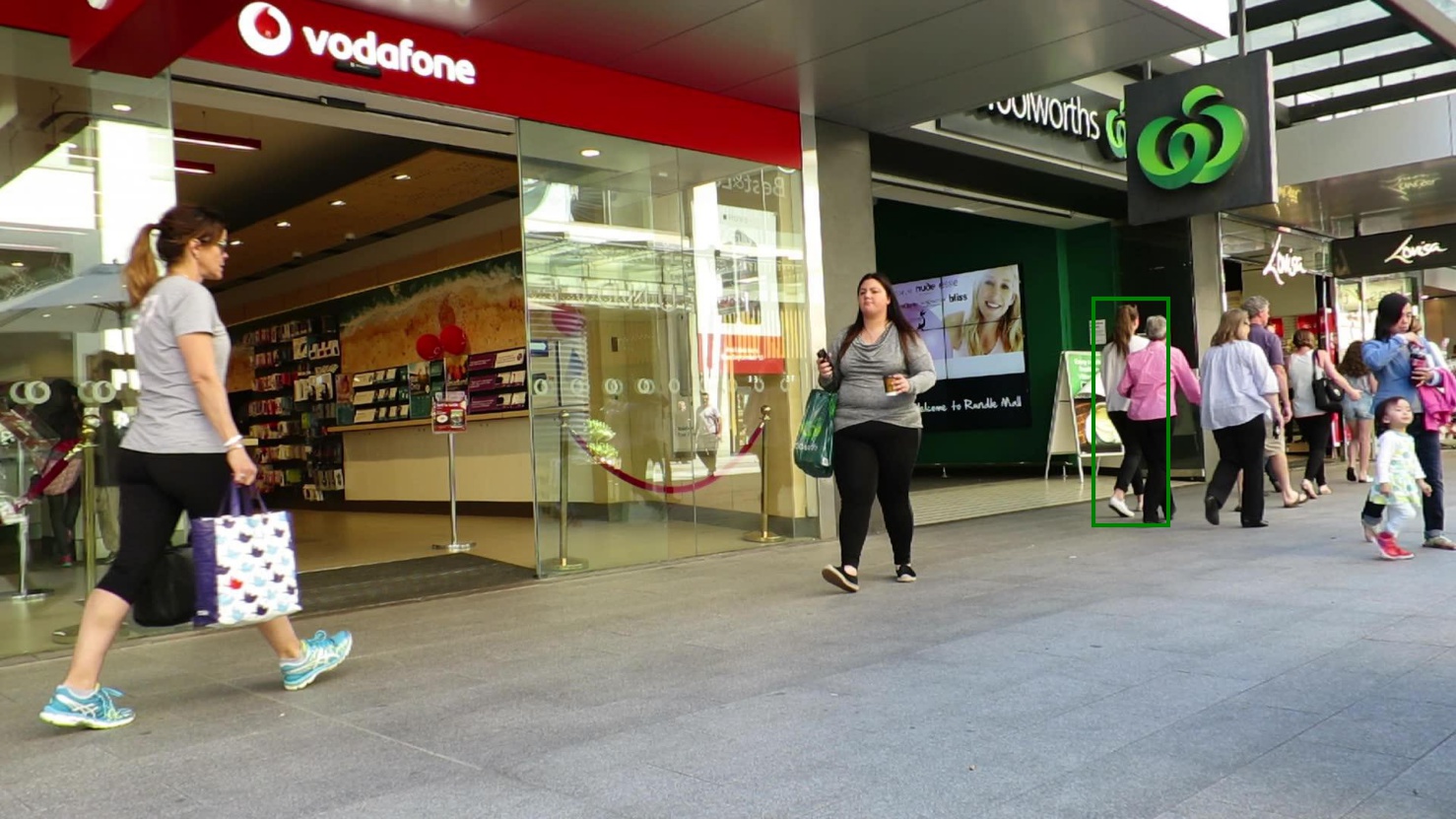}

    \end{subfigure}
    \caption{Occlusion time: 1.1s, visibility of re-appearing pedestrian: 0.2.}
    \label{fig:mot17_9_1}
\end{figure*}


\begin{figure*}[t]
    \centering
    \begin{subfigure}[b]{0.35\textwidth}   
        \centering 
        \includegraphics[width=\textwidth]{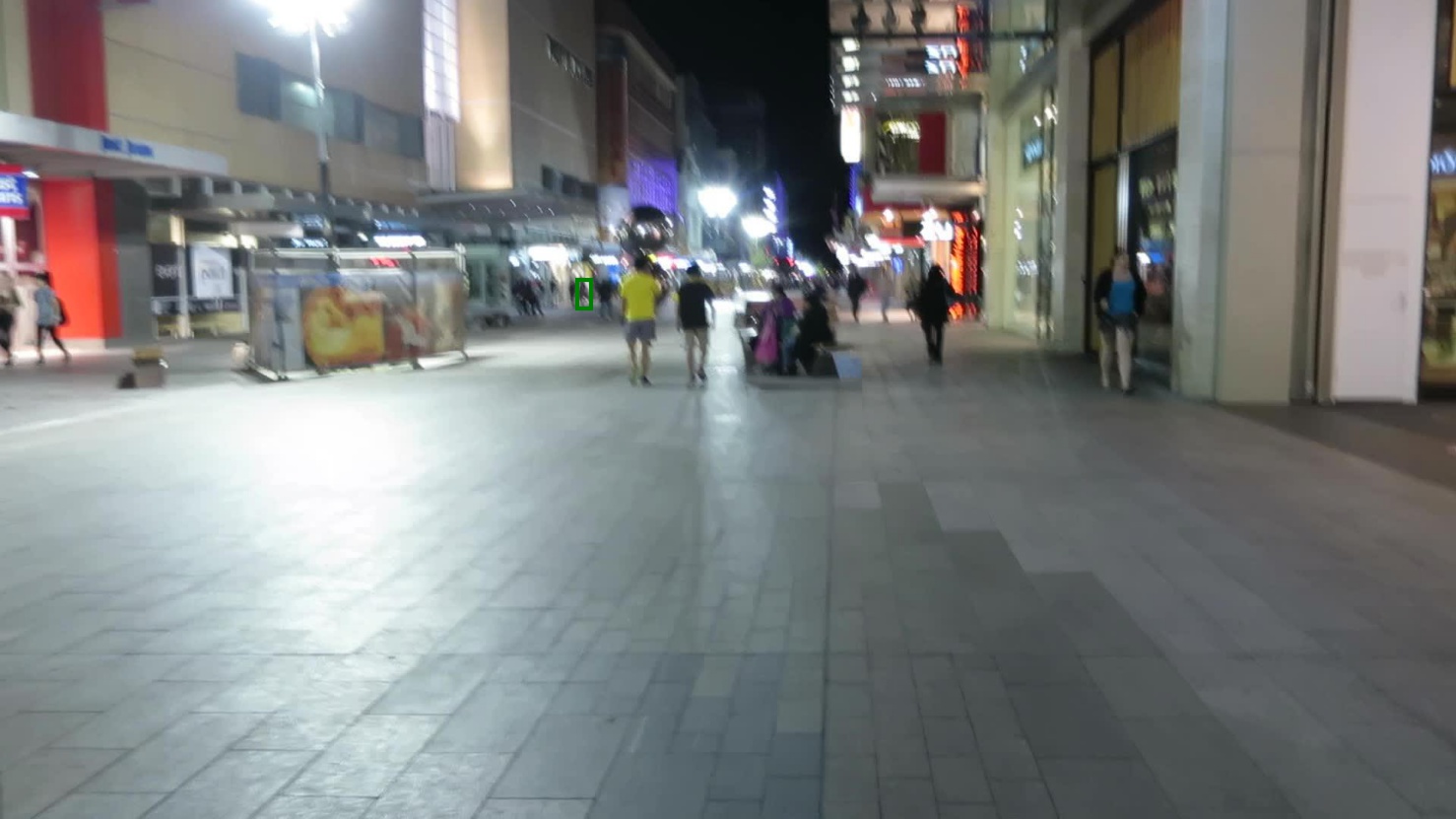}

    \end{subfigure}
    \begin{subfigure}[b]{0.35\textwidth}   
        \centering 
        \includegraphics[width=\textwidth]{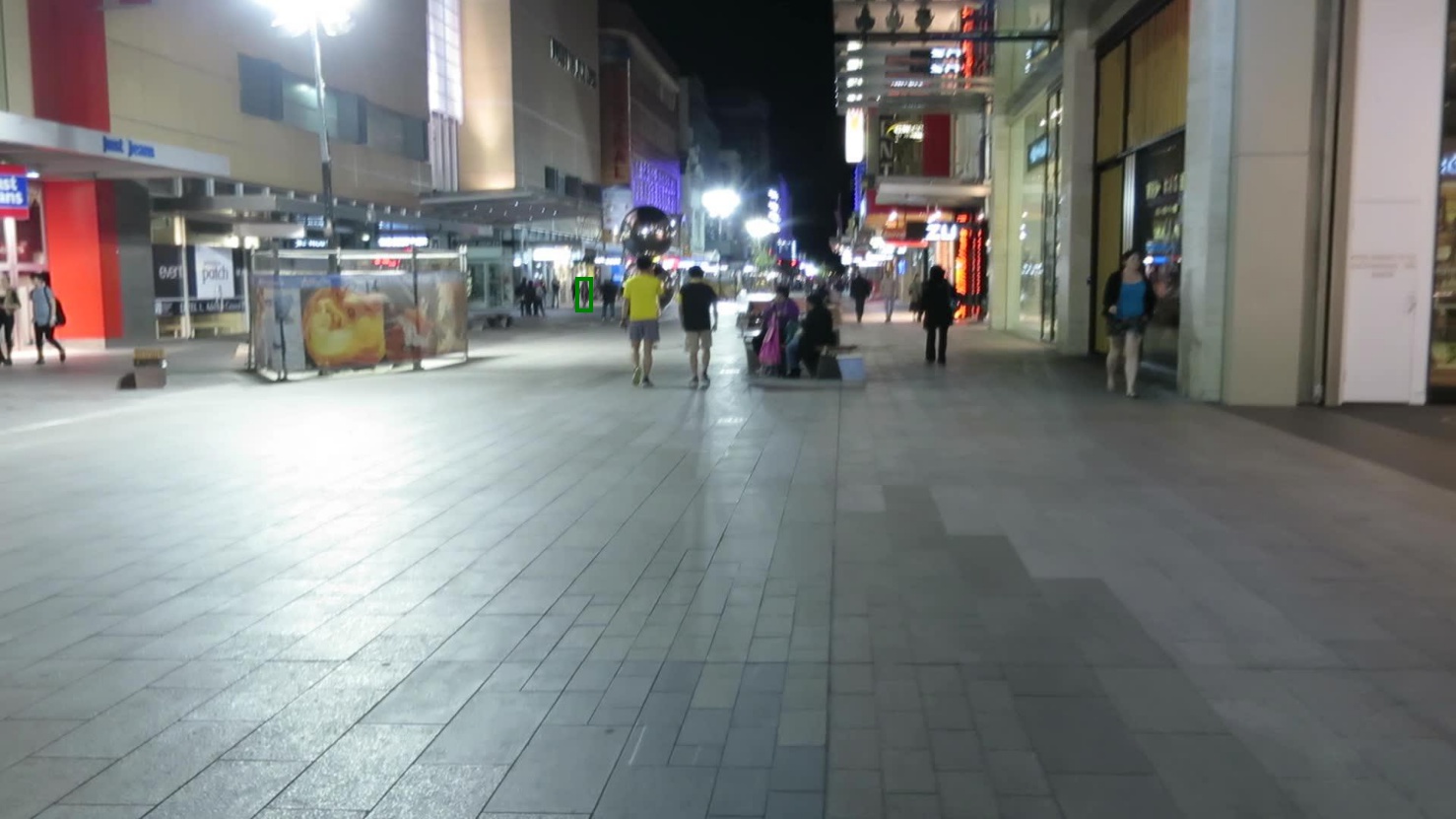}

    \end{subfigure}
    \caption{Occlusion time: 0.1s, visibility of re-appearing pedestrian: 0.6.}
    \label{fig:mot17_10_1}
\end{figure*}


\begin{figure*}[t!]
    \centering
    \begin{subfigure}[b]{0.35\textwidth}   
        \centering 
        \includegraphics[width=\textwidth]{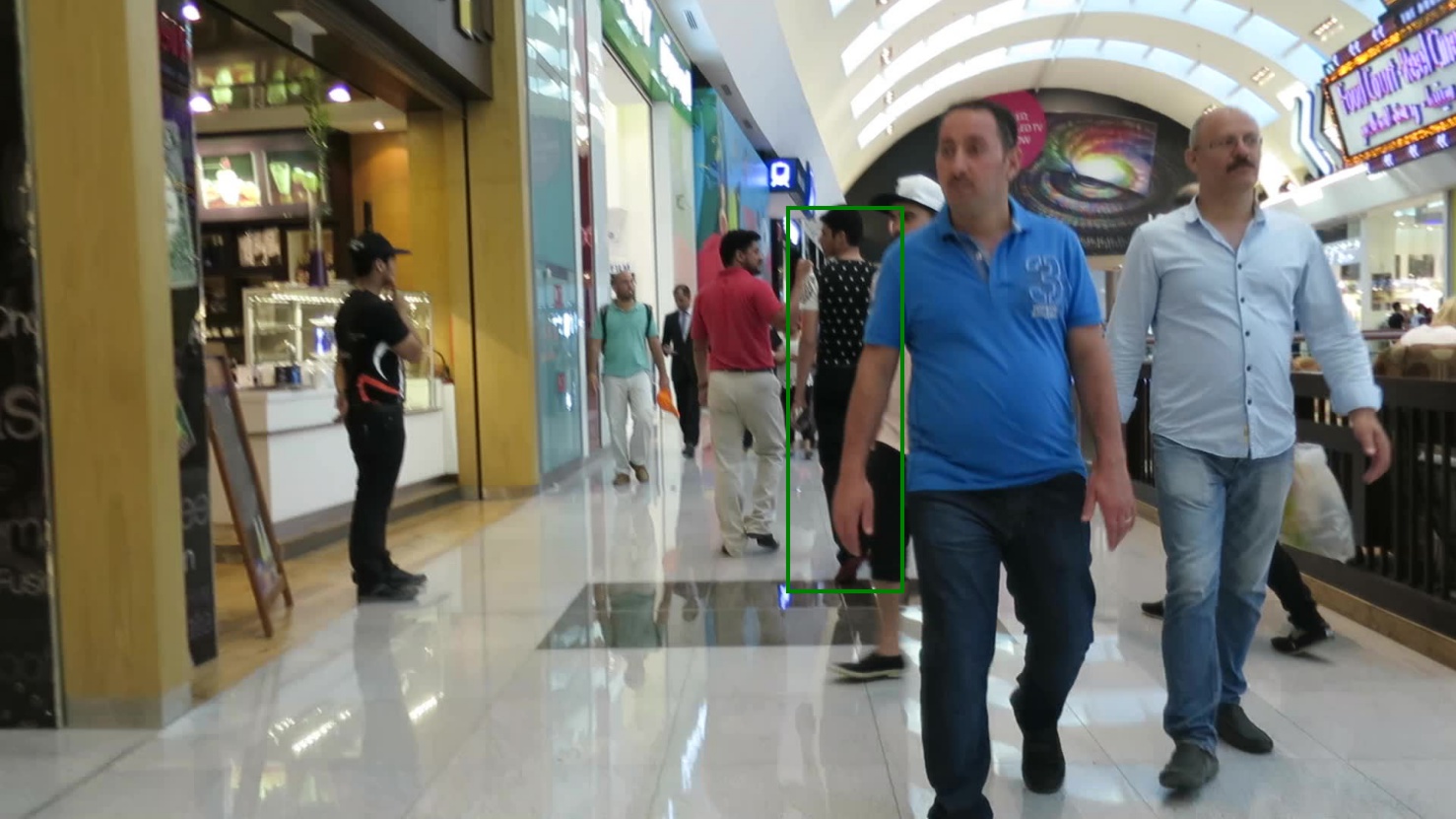}

    \end{subfigure}
    \begin{subfigure}[b]{0.35\textwidth}   
        \centering 
        \includegraphics[width=\textwidth]{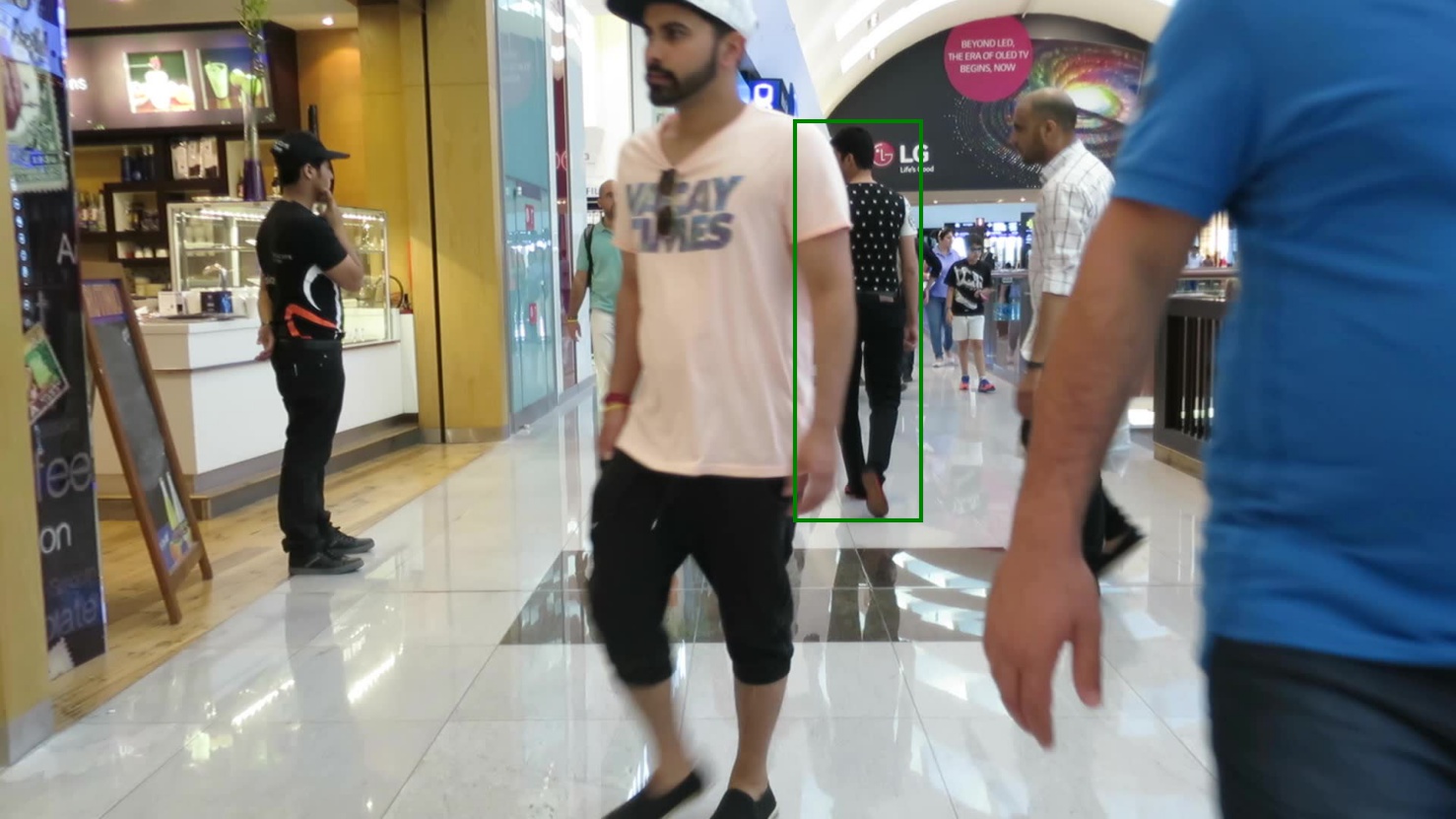}

    \end{subfigure}
    \caption{Occlusion time: 0.9s, visibility of re-appearing pedestrian: 0.5.}
    \label{fig:mot17_11_1}
\end{figure*}

\begin{figure*}[t!]
    \centering
    \begin{subfigure}[b]{0.35\textwidth}   
        \centering 
        \includegraphics[width=\textwidth]{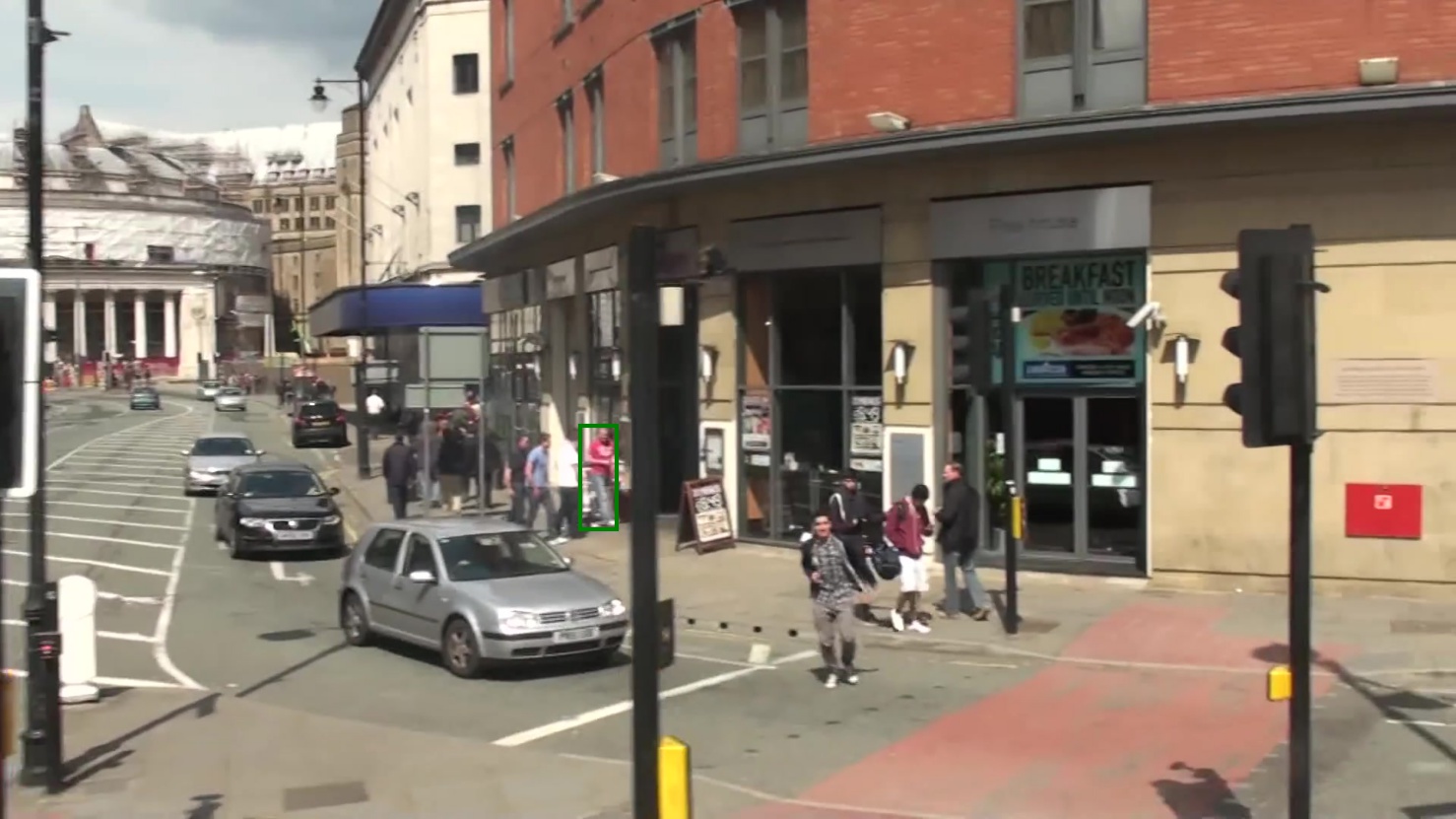}

    \end{subfigure}
    \begin{subfigure}[b]{0.35\textwidth}   
        \centering 
        \includegraphics[width=\textwidth]{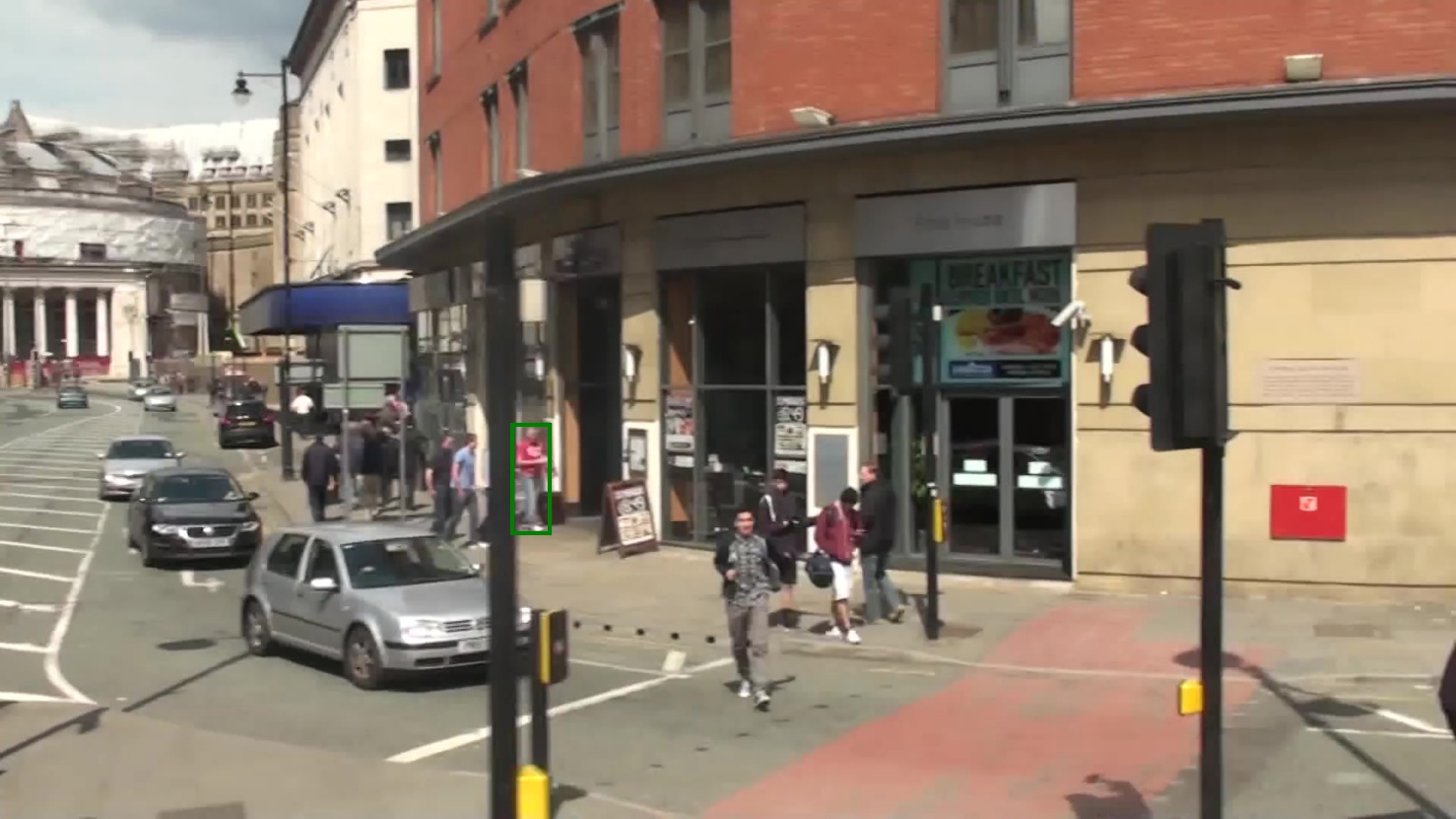}

    \end{subfigure}
    \caption{Occlusion time: 0.2s, visibility of re-appearing pedestrian: 0.8.}
    \label{fig:mot17_13_1}
\end{figure*}

\clearpage

%% file: tables_supp/bdd.tex
\begin{table*}[t!]
\centering
\resizebox{0.9\textwidth}{!}{
\begin{tabular}{@{}l|ccc|ccc|ccc|c@{}}
\toprule

  &  \multicolumn{3}{c|}{ByteTrack \cite{DBLP:journals/corr/abs-2110-06864}}   &  \multicolumn{3}{c|}{QDTrack \textbf{}}  &  \multicolumn{3}{c|}{GHOST}  & \\
 \hline
 &  \textbf{HOTA} $\uparrow$  &  \textbf{IDF1} $\uparrow$ & \textbf{MOTA} $\uparrow$  &  \textbf{HOTA} $\uparrow$ &  \textbf{IDF1} $\uparrow$  &  \textbf{MOTA} $\uparrow$  & \textbf{HOTA} $\uparrow$  &  \textbf{IDF1} $\uparrow$  &  \textbf{MOTA} $\uparrow$ & \# GT det\\
\hline\hline

pedestrian & 48.2 & 58.8 & 56.0 & 46.9 & 59.8 & 49.1 & 48.9 & 59.9 & 54.8 & 56865 \\
rider & 42.9 & 56.3 & 45.1 & 38.0 & 51.7 & 35.2 & 44.7 & 60.6 & 47.0 & 2527 \\
car & 64.5 & 72.8 & 73.5 & 64.5 & 74.9 & 69.5 & 64.5 & 73.5 & 72.9 & 339521 \\
bus & 60.5 & 70.6 & 56.2 & 52.7 & 62.3 & 40.7 & 59.9 & 70.0 & 56.0 & 9035 \\
truck & 53.3 & 61.1 & 47.9 & 48.9 & 58.1 & 39.2 & 54.0 & 63.4 & 48.2 & 27280 \\
train & 0 & 0 & 0 & 0 & 0 & 0 & 0 & 0 & -0.6 & 307 \\
motorcycle & 47.8 & 59.7 & 39.6 & 43.5 & 56.1 & 28.4 & 48.3 & 62.5 & 40.0 & 898 \\
bycicle & 46.0 & 57.6 & 43.4 & 38.9 & 49.2 & 28.6 & 45.0 & 55.1 & 41.1 & 4123 \\
\hline
class average & 45.4 & 54.6 & 45.2 & 41.7 & 51.5 & 36.3 & 45.7 & 55.6 & 44.9 & 440556 \\
detection average & 61.6 & 70.2 & 68.7 & 60.9 & 71.4 & 63.7 & 61.7 & 70.9 & 68.1 & 440556 \\

\bottomrule

\end{tabular}}
\vspace{-0.2cm}
\caption{Class-Wise BDD100k Validation Set Performance.}
\label{tab:bdd}
\end{table*}